%% file: main.tex
\documentclass[runningheads]{llncs}

\usepackage{eccv}

\usepackage{eccvabbrv}

\usepackage{graphicx}
\usepackage{booktabs}

\usepackage[accsupp]{axessibility}  

\usepackage{hyperref}

\usepackage{orcidlink}

\usepackage{url}
\usepackage{graphicx} 
\usepackage{multirow}
\usepackage{algorithm}
\usepackage{algorithmic}
\usepackage{amsmath, amssymb}
\usepackage{booktabs}
\usepackage{subcaption}
\usepackage{caption}  
\usepackage{pgfplots}
\usepackage[table]{xcolor}
\usepackage{wrapfig}
\usepgfplotslibrary{statistics, groupplots} 
\usepgfplotslibrary{colorbrewer} 
\usepgfplotslibrary{fillbetween} 
\usepgfplotslibrary{colormaps} 
\pgfplotsset{compat=1.18}

\definecolor{mypurple}{RGB}{70,22,147} 
\definecolor{mygreen}{RGB}{0,204,0}

\newcommand{\name}{Controlled Random Zigzag Sampling}
\newcommand{\abb}{$\mathtt{Ctrl}$-$\mathtt{Z}$ Sampling}
\newcommand{\ab}{$\mathtt{Ctrl}$-$\mathtt{Z}$}
\newcommand{\rev}[1]{#1}

\begin{document}

\title{Ctrl-Z Sampling: \rev{Scaling} Diffusion Sampling with Controlled Random Zigzag Explorations} 

\titlerunning{Ctrl-Z Sampling}

\author{Shunqi Mao \orcidlink{0009-0007-0326-9004} \and Wei Guo\orcidlink{0009-0005-0920-2949} \and Chaoyi Zhang\orcidlink{0000-0001-8492-9711} \and Jieting Long\orcidlink{0009-0003-0753-5243} \and Ke Xie\orcidlink{0009-0004-0137-6269} \and Weidong Cai\orcidlink{0000-0003-3706-8896}}

\authorrunning{S.~Mao et al.}

\institute{School of Computer Science, The University of Sydney, Australia \\
\email{\{shunqi.mao, wei.guo, chaoyi.zhang, jieting.long, kxie0655, tom.cai\}@sydney.edu.au} 
}

\maketitle

\begin{abstract}
Diffusion models generate conditional samples by progressively denoising Gaussian noise, yet the denoising trajectory can stall at visually plausible but low-quality outcomes with conditional misalignment or structural artifacts. 
We interpret this behavior as local optima in a surrogate quality landscape: Once early denoising commits to a suboptimal global structure, later steps mainly sharpen details and seldom correct the underlying mistake.
While existing inference-time approaches explore alternative diffusion states via re-noising with fixed strength or direction, they exhibit limited capacity to escape steep quality plateaus. We propose \name{} (\abb{}), a scalable sampling strategy that detects plateaus in quality landscape via a surrogate score, and allocates exploration only when a plateau is detected. Upon detection, \abb{} rolls back to noisier states, samples a set of alternative continuations, and updates the trajectory when a candidate improves the score, otherwise escalating the exploration depth to escape the current plateau.
The proposed method is model-agnostic and broadly compatible with existing diffusion frameworks.
Experiments show that \abb{} consistently improves generation quality over other inference-time scaling samplers across different NFE budgets, offering a scalable compute-quality trade-off.
Code available at: \url{https://github.com/ShunqiM/Ctrl-Z-Sampling}.
\end{abstract}

\begin{figure}
    \centering
    \includegraphics[width=\linewidth]{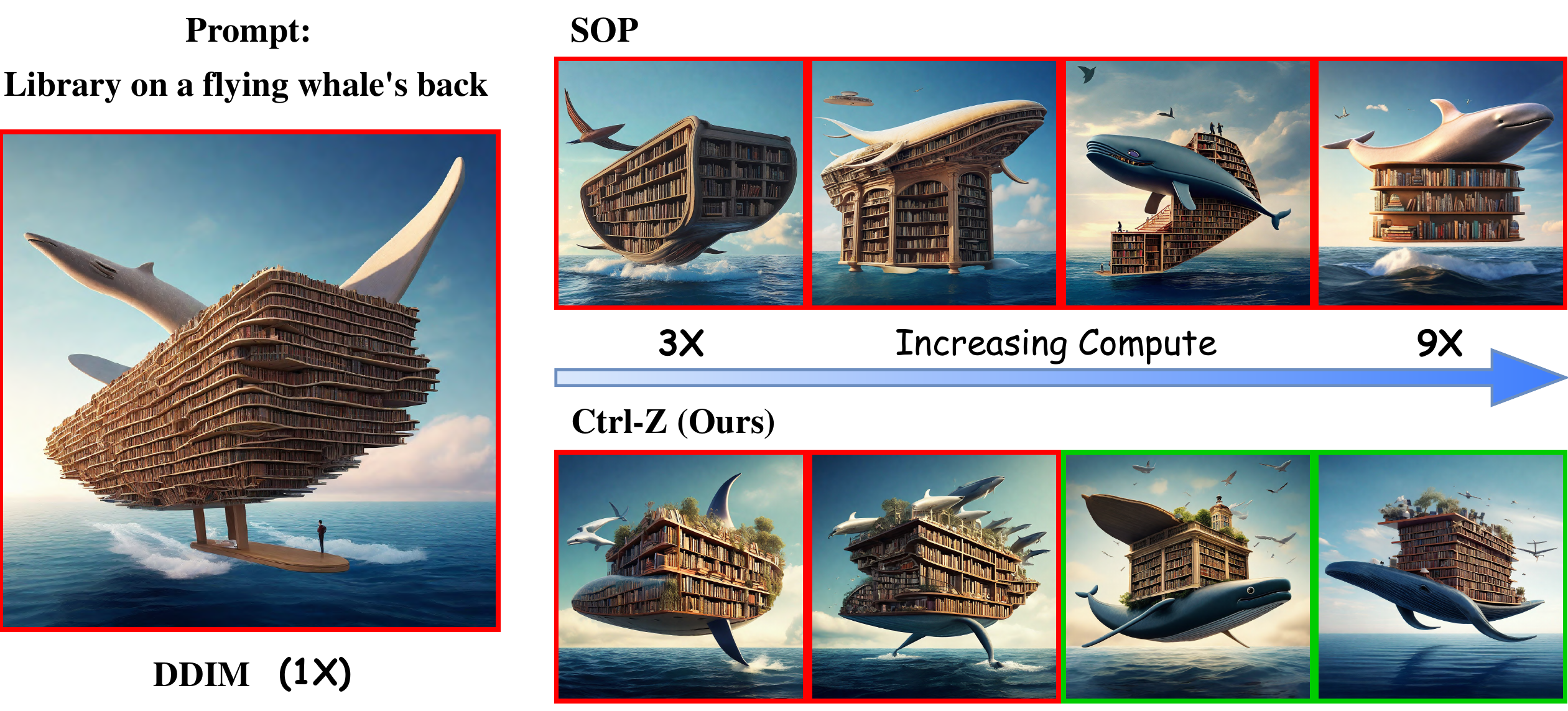}
    \caption{We introduce \abb{}, a diffusion sampling strategy that scales with compute to improve generation quality. Under comparable budgets, it outperforms inference-scaling baselines such as Search-over-Path (SOP). For the prompt \textit{``Library on a flying whale’s back''}, SOP produces misaligned compositions (\textit{red}), while \ab{} generate prompt-consistent (\textit{green}) output as compute increase from $3\times$ to $9\times$ NFEs.}
    \label{fig:teaser}
\end{figure}

\section{Introduction}
\label{sec:intro}

Diffusion models are a powerful class of generative models, achieving state-of-the-art performance across a wide range of generative tasks, including image \cite{ldm, podell2023sdxl}, video \cite{ho2022video, blattmann2023stable}, text \cite{li2022diffusion}, audio \cite{kong2021diffwave, guo2026gotta}, 3D object \cite{poole2023dreamfusion}, along with a growing number of domains \cite{chen2024diffusion, xu2022geodiff, chamberlain2021grand, janner2022diffuser}. These models gradually transform Gaussian noise into data samples through an iterative denoising process \cite{sohl2015deep, ddpm, ddim}. Modern variants are typically designed to accept conditional inputs, enabling guided generation based on labels, texts, or other structured signals \cite{controlnet, bansal2023universal, wang2025lavin}.

Despite their strong generative performance, diffusion models often exhibit semantic misalignment or global inconsistency in conditional generation. 
\rev{Taking image generation as an example, the denoising process is an iterative refinement procedure on the data manifold. In practice, this refinement can stagnate in suboptimal regions where the current sample already looks locally plausible but remains semantically flawed (\eg, missing objects, incorrect relations, or anatomically implausible details). We refer to this sampling stagnation or collapse as converging to a \emph{local optima} in a conceptual sample quality space, where quality reflects how faithful, structurally correct, and well aligned the image is with the text condition.}
This notion of optimality is defined by a quality objective rather than the diffusion model likelihood.
Such failures are commonly triggered by misaligned initial noise, which misguides the trajectory toward outputs that are visually compelling but misaligned with the intended condition~\cite{noise1}.

Some existing works mitigate these failures by reinforcing conditional information during sampling.
Classifier-free guidance (CFG)~\cite{cfg} amplifies the conditioning signal, while other approaches alternate conditional denoising with unconditional inversion steps~\cite{resample, zigzag} to trade off plausibility for improved conditioning.
These strategies can strengthen conditional influence, yet their intervention is often loosely controlled and may be insufficient when progress stalls on broad plateaus.
Recent test-time scaling methods allocate additional inference compute via score-guided exploration, for example applying small mutations to the current state or re-noising for a fixed number of steps before re-denoising~\cite{diffusion_tts, traj_search, evo_search, classical}.
However, exploration is typically shallow, so escaping steep or broad local optima often requires substantially increasing the exploration budget.
Moreover, exploration is commonly applied at every step or at preset intervals, which can spend compute broadly while still failing to trigger sufficiently non-local moves when stagnation occurs.
As a result, these methods can be impractical for local inference settings.

In this paper, we propose \name{} (\abb{}), a diffusion sampling strategy that alleviate above issues and delivers consistent quality gains across different NFE budgets (as shown in \Cref{fig:teaser}).
Specifically, \abb{} escapes local optima through randomized backward exploration into adaptively noisier timesteps.
\abb{} detects potential local maxima using a stagnation criterion on the preference scores from a reward model, which we use as a surrogate for conceptual sample quality.
Upon detection, it injects noise and reverts to an earlier, noisier timestep to explore beyond the current quality plateau.
The reward model evaluates each candidate state, and \abb{} accepts only those that yield more promising future trajectories.
When no improvement is found, \abb{} retreats to deeper noise levels to increase the chance of escape.
Overall, \abb{} induces a controlled zigzag trajectory that alternates between forward refinement and backward exploration, improving both condition alignment and visual quality.

\abb{} is broadly applicable across diffusion backbones. Experiments on text-to-image benchmarks show that \abb{} consistently improves generation quality across multiple metrics under low-budget NFEs, providing a controllable quality-compute trade-off via its exploration strength. This suggests \abb{} as a practical test-time scaling alternative for single-device, per-sample inference, where large candidate pools or extreme NFE budgets are often infeasible.
To summarize, our key contributions are: 
\begin{itemize}
   
    \item We interpret conditional diffusion sampling as a hill-climbing-like process in a surrogate sample-quality space, and empirically show that existing strategies can stall on broad plateaus that exhibit conditional misalignment or global inconsistency due to insufficient exploration depth.
    \item We propose \name{} (\abb{}), a reward-guided diffusion sampler with controlled explorations for adaptive deeper exploration, enabling escape from local optima during generation.
    \item We conduct extensive experiments demonstrating that \abb{} substantially improves text-to-image generation quality. 
    Results show that, unlike search methods relying on numerous shallow trials across many directions, \abb{} achieves better exploration efficiency by making fewer, progressively deeper steps in the generation landscape.
\end{itemize}

\section{Related Works}

\subsection{Diffusion Models Inference Techniques}

Diffusion models generate data by gradually transforming Gaussian noise into structured samples through iterative denoising \cite{sohl2015deep, ddpm, ddim}, with both U-Net-based \cite{ldm, podell2023sdxl, imagen} and Transformer-based architectures \cite{dit, li2025fractal} achieving strong results across modalities. 
However, training or finetuning remains costly, and constrained by computational demands \cite{diffusion_dpo, fan2023dpok}.
To address these challenges, various inference-time techniques have been proposed to enhance conditional generation \cite{cg}. Classifier-free guidance (CFG) \cite{cfg, chung2024cfg++} contrasts conditional and unconditional predictions to improve alignment, though at the risk of fidelity loss. Others guide generation by modifying attention maps during sampling \cite{StructureDiffusion, chefer2023attend, li2023divide, hong2024smoothed}, but these can struggle with abstract or global conditions.
While pruning techniques for diffusion models facilitate efficient inference by reallocating saved compute to additional denoising steps for enhanced generation quality \cite{prune1, prune2, prune3}, they do not improve per-step denoising quality independently.
More recent methods exploit the sensitivity of generation to noise initialization, leveraging information leakage from noisy priors \cite{leak1, leak2} and developing strategies to optimize or select noise vectors \cite{noise1, noise2, zhou2024golden, guo2024initno, noise3, dno}, or inject low-frequency condition cues \cite{guttenberg2023diffusion}. 
\rev{However, guided optimization of noise latents can be inefficient and may also lead to degenerate solutions by collapsing the denoising process.}
In contrast, \abb{} improves conditional alignment without retraining or costly optimization by adaptively enhancing exploration strength, enabling escape from broad local maxima and improving generation quality at inference.

\subsection{Diffusion Sampling Strategies}

DDPM \cite{ddpm} introduces diffusion sampling as a stochastic reverse process, which DDIM \cite{ddim} later reformulates as a deterministic ODE for faster inference. Building on this, recent methods such as DPM-Solver \cite{lu2022dpm}, UniPC \cite{zhao2023unipc}, and AYS \cite{ays} improve both efficiency and quality through high-order ODE solvers and adaptive step sizes, while some other work incorporates directional guidance \cite{watson2022learning, ghosh2023implicit, fair, yin2026accelaes}.
Later works improve sampling quality via latent-space inversion or zigzag explorations \cite{resample, xu2023restart, zigzag}.
In parallel, guided search by reward models has been explored via particle-based sampling \cite{kim2025test, li2025dynamic} and beam search \cite{beam1}. 
More recent efforts extend these toward test-time scaling in diffusion models \cite{diffusion_tts, traj_search, fk_steering, evo_search, classical}, aiming to effectively improve sampling performance with more function evaluations.
However, these approaches primarily focus on enlarging the candidate pool through shallow, local perturbations, often neglecting the role of exploration depth and thus remaining vulnerable to local optima. 
\rev{In contrast, concurrent works}{\cite{classical, dts}} \rev{adopt DFS-style search strategies that roll back diffusion states to significantly earlier, noisier timesteps, increasing exploration depth but requiring costly re-denoising from these deeper noise levels.}
In contrast, \abb{} mitigates this by applying controlled exploration only upon detecting stagnation, and adaptively increases inversion depth until improvement is found, avoiding unnecessary computation elsewhere.
Additional relevant sampling and scaling methods are discussed in the Supp. \Cref{sec::extend_related_work}.

\section{Preliminaries}
\label{sec:2}

\subsection{Diffusion Process}
Diffusion models are generative frameworks that synthesize data by progressively denoising a sample drawn from a Gaussian prior. Let $\mathcal{N}$ represent the standard normal distribution and $\mathcal{D}$ denote the target data distribution. Starting from a noise sample $x_T \sim \mathcal{N}$, a sample $x_0 \in \mathcal{D}$ is generated through a sequence of $T$ denoising steps, conditioned on $x_T$ and auxiliary prompt $c$: 
$
x_0 = \Phi(x_T, c),
$
where $\Phi$ denotes the sampling procedure, which may be stochastic or deterministic.

\subsection{Deterministic Sampling with DDIM}
We adopt the DDIM sampling framework~\cite{ddim}, which enables efficient and deterministic generation by removing stochasticity from the denoising process. 
Given a predefined noise schedule $\{\beta_t\}_{t=1}^T$, let $\alpha_t = \prod_{i=1}^{t} (1 - \beta_i)$ denote the cumulative product.
The generation process is decomposed into $T$ successive mappings:
\begin{equation}
x_0 = \Phi^{1} \circ \Phi^{2} \circ \cdots \circ \Phi^{T}(x_T, c),
\end{equation}
where each mapping $\Phi^t$ transforms $x_t$ into $x_{t-1}$ using the predicted noise $\epsilon_\theta^t(x_t, c)$:
\begin{equation}
\begin{aligned}
x_{t-1} = \Phi^t(x_t, c) &=
 \sqrt{\alpha_{t-1}} \cdot \frac{x_t - \sqrt{1 - \alpha_t} \cdot \epsilon_\theta^t(x_t, c)}{\sqrt{\alpha_t}}  + \sqrt{1 - \alpha_{t-1}} \cdot \epsilon_\theta^t(x_t, c).
\end{aligned}
\label{eq:ddim-step}
\end{equation}

Note the $\sigma$ term in DDIM is zeroed and omitted for determinism. The unscaled first term in \Cref{eq:ddim-step} is a Tweedie-style posterior-mean estimation \cite{Efron2011TweediesFA} of the clean sample corresponding to $x_t$, which we denote as:
\begin{equation}
\hat{x}_0^{\,t-1}(x_t, c) = \frac{x_t - \sqrt{1 - \alpha_t} \cdot \epsilon_\theta^t(x_t, c)}{\sqrt{\alpha_t}}.
\label{eq:x0-estimate}
\end{equation}
This intermediate estimate serves as a useful proxy for the final output and is used in our method to assess semantic alignment via a reward model. Although $\Phi^t$ formally returns only $x_{t-1}$, we compute and reuse $\hat{x}_0^{\,t-1}$ jointly for efficiency.

\subsection{Latent Inversion}
DDIM defines a deterministic denoising trajectory but lacks an explicit inversion \rev{re-noising} mechanism. To reintroduce noise in a controlled manner, we define an inversion operator
$\Psi(x_t, \Delta)$, which simulates a forward transition from $x_t$ to a noisier latent state (or pixel-space state, for pixel diffusion models) $x_{t+\Delta}$ using:
\begin{equation}
\label{eq:ddim-inversion}
x_{t+\Delta} = \Psi(x_t, \Delta; \epsilon) = \sqrt{\frac{\alpha_{t+\Delta}}{\alpha_t}} \cdot x_t + \sqrt{1 - \frac{\alpha_{t+\Delta}}{\alpha_t}} \cdot \epsilon,
\end{equation}
where $\alpha_t = \prod_{i=1}^{t} (1 - \beta_i)$ is the cumulative noise schedule inherited from the original diffusion process, and $\epsilon  \sim \mathcal{N}(0, \mathbf{I})$ a random noise.

This operator enables transitions into higher-noise manifold of the latent space while retaining denoising progress. It provides a mechanism for test-time exploration that perturbs the generation trajectory without completely discarding previously accumulated semantic structure.

\begin{figure*}[!t]
  \centering
    \includegraphics[width=\linewidth]{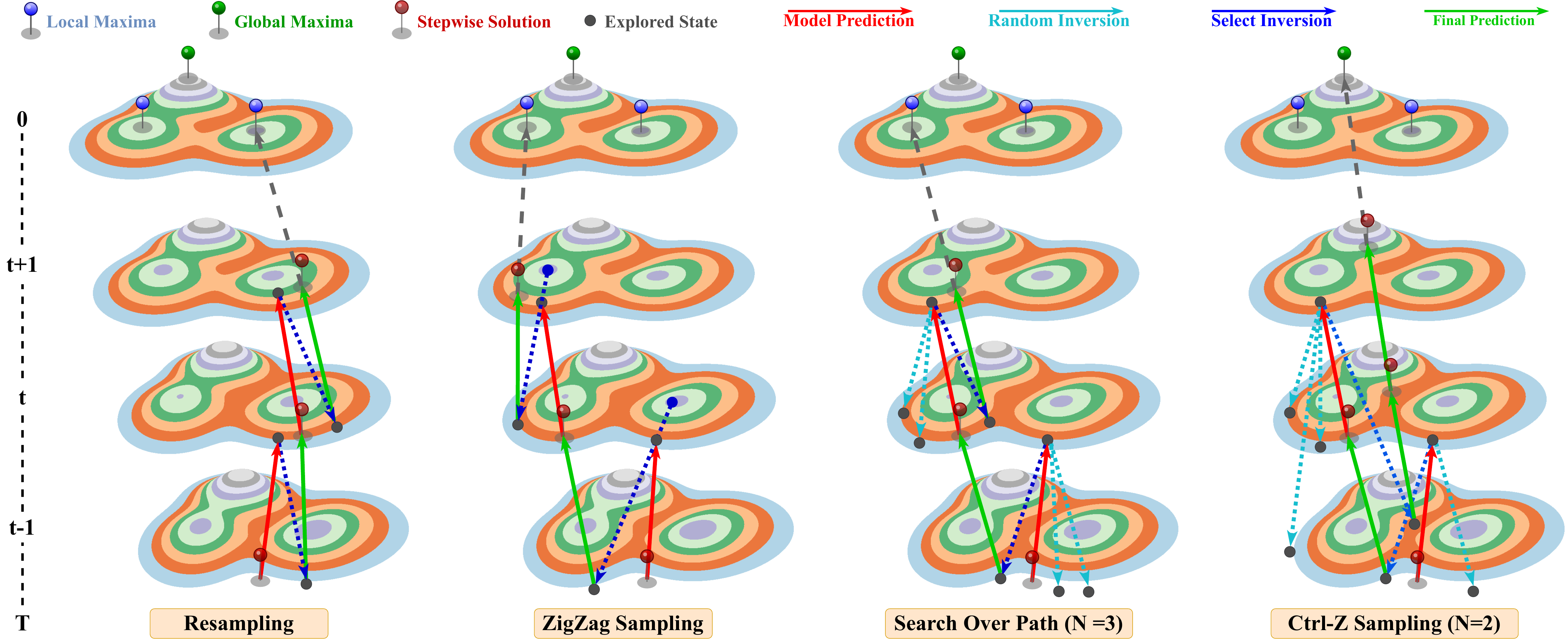}

    \caption{Illustration of different sampling strategies during the diffusion process.
The diffusion trajectory is depicted as ascending a rugged, {conceptual quality} landscape for sample quality and condition alignment, measured by a surrogate score.
Each strategy begins with a standard denoising update (\textit{\textcolor{red}{\textbf{red}}}), followed by either a single or multiple inversion explorations.
The selected/executed inversions are marked in \textit{\textcolor{blue}{\textbf{blue}}}, while discarded ones are in \textit{\textcolor{cyan}{\textbf{cyan}}}.
The accepted inversion is followed by a forward conditional denoising step (\textit{\textcolor{mygreen}{\textbf{green}}}).
Subsequent steps are omitted and shown in \textit{\textcolor{gray}{\textbf{gray}}}.
Unlike prior methods, the proposed \abb{} adaptively increases exploration strength through larger inversion steps when local perturbations fail to reveal improved trajectories, enabling escape from broader local maxima. 
$N$ denotes the number of inversion candidates per exploration stage.
}
  \label{fig::main}
\end{figure*}

\section{Methods}
In this section, we first develop the intuition for \abb{} through a hill-climbing-like view in surrogate quality space in \Cref{sec:hill_climb}, and then present its technical details; the full procedure is summarized in \Cref{algo:main}.

\subsection{Denoising Trajectories in Surrogate Quality Space}\label{sec:hill_climb}

As illustrated in \Cref{fig::main}, we analyze deterministic sampling trajectories through a surrogate quality objective.
Conceptually, we posit a quality functional $Q(x,c)$ and approximate it with an off-the-shelf scorer $R(x,c)$.
At step $t$, sampling yields a clean prediction $\hat{x}_0^{(t)}$ and a score $r_t = R(\hat{x}_0^{(t)},c)$, forming a trajectory $\{r_t\}_t$ over timesteps.
In practice, the score trajectory often plateaus: the sample remains visually plausible under the base model, yet fails to improve under $R$.
We refer to such plateaus as \textit{local optima in surrogate quality space}, meaning that within a bounded exploration budget the nearby continuations explored from the current state do not yield sufficient score gains.
From this perspective, standard deterministic sampling behaves like a greedy local search that can become trapped on broad plateaus (as visualized in the trajectory analysis in Supp. \Cref{fig:treepath}).

\begin{algorithm}[t]
    \caption{\abb{}}
    \scriptsize

\begin{algorithmic}[1]
\STATE {\bfseries Input:} Denoising operator $\Phi^t$, clean \rev{sample} estimate $\hat{x}_0^{\,t-1}(x_t, c)$, inversion operator $\Psi(x_t, \Delta; \epsilon)$, condition $c$, total steps $T$, reward model $R$, exploration window $\lambda$, accept threshold $\delta$, max inversion depth $d_{\max}$, max number of candidates $N$.
\STATE {\bfseries Output:} Final image estimate ${x}_0$
\STATE Sample Gaussian noise $x_T$
\STATE Initialize reward score $r_{\text{prev}} \leftarrow -\infty$
\FOR{$t = T$ \TO $1$}
    \STATE {{\color{red}  $x_{t-1} \leftarrow \Phi^t(x_t, c)$; \quad $\hat{x}_0^{\,t-1} \leftarrow \hat{x}_0^{\,t-1}(x_t, c)$} \hfill \# Equation~(\ref{eq:ddim-step},~\ref{eq:x0-estimate})}%
    \IF{$t > T - \lambda$}
        \STATE $r \leftarrow R([c], \hat{x}_0^{\,t-1})$
        \IF{$r \geq r_{\text{prev}} + \delta$}
            \STATE $r_{\text{prev}} \leftarrow r$
        \ELSE
            \STATE $\text{inversion\_step} \leftarrow 1$, $\text{best\_score} \leftarrow r$, $\text{best\_state} \leftarrow x_{t-1}$
            \WHILE{$\text{inversion\_step} \leq d_{\max}$}
                \STATE $\Delta \leftarrow \min(\text{inversion\_step}, T-t)$
                \FOR{$i = 1$ \TO $N$}
                    \STATE Sample $\epsilon \sim \mathcal{N}(0, \mathbf{I})$
                    \STATE {\color{cyan} $\tilde{x}_{t+\Delta} \leftarrow \Psi(x_t, \Delta; \epsilon)$ \hfill \# Inversion step}
                    \FOR{$k = t+\Delta$ \TO $t$}
                        \STATE $\tilde{x}_{k-1} \leftarrow \Phi^k(\tilde{x}_k, c)$
                    \ENDFOR
                    \STATE Estimate $\hat{x}_0^{\,t-1} \leftarrow \hat{x}_0^{\,t-1}(\tilde{x}_t, c)$
                    \STATE $r_{\text{cand}} \leftarrow R(c, \hat{x}_0^{\,t-1})$
                    \IF{$r_{\text{cand}} > \text{best\_score}$}
                        \STATE {$\text{best\_score} \leftarrow r_{\text{cand}},\quad \text{best\_state} \leftarrow \tilde{x}_{t-1}$}
                    \ENDIF
                \ENDFOR
                \IF{$\text{best\_score} \geq r_{\text{prev}} + \delta$}
                    \STATE \textbf{break} from search
                \ENDIF
                
                \STATE $\text{inversion\_step} \leftarrow \text{inversion\_step} + 1$
            \ENDWHILE
            \STATE {\color{mygreen}  $x_{t-1} \leftarrow \text{best\_state}$, \quad $r_{\text{prev}} \leftarrow \text{best\_score}$}
        \ENDIF
    \ENDIF
    \STATE $x_t \leftarrow x_{t-1}$
\ENDFOR
\RETURN $x_0$
\end{algorithmic}

    \label{algo:main}
\end{algorithm}

In the illustrated example, Resampling \cite{resample} applies a random, shallow random inversion to perturb the state and probe nearby alternatives, which may be insufficient when the plateau is broad.
Z-Sampling \cite{zigzag} performs inversion biased by an unconditional trajectory and then re-applies conditional denoising, repeatedly re-injecting the prompt signal.
SOP \cite{diffusion_tts} extends Resampling by scoring multiple candidates, but uses a fixed exploration depth, which limits its ability to escape wider plateaus.
In contrast, \abb{} adaptively adjusts the strength of latent perturbation based on reward feedback. When shallow exploration fails, \abb{} incrementally expands its search radius, enabling escape from both narrow and broad local maxima while preserving output quality.


\subsection{\name{}}

To enable conditional diffusion models to escape local optimal, we propose a reward-guided sampling strategy that incorporates controlled backward exploration. Our approach detects potential suboptimal states using a reward model, then dynamically perturbs the latent trajectory \rev{(\ie, the sequence of intermediate latent states ${x_t}$ produced by DDIM sampling)} through guided inversion steps of increasing strength. This process forms a zigzag trajectory in the latent space, alternating between conditional refinement and noise-space exploration.

\subsection{Local Maxima Detection}

At each selected step $t$, the clean \rev{sample} estimate $\hat{x}_0^{\,{t-1}}$ is computed from the current latent state $x_t$ using \Cref{eq:x0-estimate}, and its quality is evaluated by a reward model $R(c, \hat{x}_0^{\,{t-1}})$  that assigns a scalar score to the candidate state. A local maximum is detected when the current reward fails to improve over the most recently accepted score by at least $\delta$:
\begin{equation}
R(c, \hat{x}_0^{\,{t-1}}) < r_{\text{prev}} + \delta,
\label{eq::reward}
\end{equation}
where $\delta$ denotes the acceptance threshold. By default, $\delta = 0$ enforces non-decreasing reward over time.
$\delta$ governs the minimal improvement required for a candidate to be accepted. A positive $\delta$ suppresses trivial gains and promotes robustness in candidate selection, while $\delta \le 0$ allows more permissive updates, including those with marginal or slightly lower reward values, which may carry a higher risk of degraded outputs.
This criterion provides a simple yet effective signal for identifying local maxima during sampling. However, relying on a fixed offset from the prior best can rigidly penalize high-reward plateaus, especially when $\delta$ is large. Designing more flexible criteria such as relative thresholds or global reward schedulers could mitigate such cases and improve adaptability. We leave such extensions to future work.

\subsection{Controlled Latent Inversion}
To escape from a detected local maximum, we apply the inversion operator $\Psi(x_t, \Delta; \epsilon)$ (defined in \Cref{eq:ddim-inversion}) to perturb the current latent $x_t$ toward a higher-noise state $x_{t+\Delta}$. The step size $\Delta \in \mathbb{N}$ corresponds to discrete timestep indices in the DDIM schedule, where smaller values induce mild perturbations and larger values enable broader exploration. This controlled noise injection serves as a test-time search mechanism that reintroduces uncertainty while preserving structural information acquired during prior denoising, thereby enabling the model to explore nearby alternative trajectories effectively.

\subsection{Candidate Selection and Adaptive Inversion Step}
After applying the inversion, the model resumes forward denoising from $x_{t+\Delta}$ to $x_{t-1}$ using standard conditional DDIM steps $\Phi^k(\cdot, c)$ for $k = t+\Delta, \dots, t$. The resulting candidate $\hat{x}_0^{\,t-1}$ is then evaluated by the reward model. At each iteration, $N$ candidates are generated using distinct noise vectors $\epsilon$, and the one with the highest reward is selected. If its score exceeds the most recently accepted reward by at least a threshold $\delta$, the update is accepted.

Otherwise, the inversion step size $\Delta$ is incrementally increased (\eg, from 1 to $d_{\max}$), enabling progressively stronger perturbations to search for better trajectories. This defines a greedy yet adaptive hill-climbing strategy: the search proceeds until a satisfactory candidate is found or the maximum inversion depth is reached. If no improvement beyond the threshold is identified, the best candidate observed during the search is retained, promoting stability while often yielding results comparable to or better than the default forward step.

The inversion depth can be unbounded with $d_{max} = \infty$, and the candidate budget $N$ per step guarantees termination. The candidate evaluation process is inherently parallelizable since each candidate trajectory is conditionally independent until selection, thus can be efficiently distributed across computational resources, supporting scalable inference with minimal overhead.

Empirically, at early timesteps (\eg, $t = 600/1000$), the DDIM update already assigns substantial weight (above 0.02) to the clean estimate $\hat{x}_0^{\,{t-1}}$ via \Cref{eq:ddim-step}, meaning $x_{t-1}$ is strongly influenced by $\hat{x}_0^{\,{t-1}}$. As a result, much of the image’s low-frequency structure is established early, making later corrections less effective. 
Therefore, \abb{} restricts exploration to the first $\lambda$ steps from the start of sampling (the high-noise regime, $t \in \{T, \dots, T-\lambda+1\}$), where trajectory perturbations can still meaningfully influence global structure.

\section{Experiments and Results}

\begin{table}[t!]
\centering
\small
\caption{Quantitative results of sampling methods on Pick-a-Pic and DrawBench, evaluated using four human-aligned metrics. Results are reported for both Stable Diffusion 2.1 and Hunyuan-DiT, along with average number of function evaluations (NFEs). \rev{$\dagger$ and $\ddagger$ indicate $\mathtt{Ctrl}$-$\mathtt{Z}$ variants with different exploration parameters, demonstrating performance under varying NFEs.}}
\label{tab:sampling-results}
\begin{tabular}{lcccc|cccc|c}
\toprule
 & \multicolumn{4}{c}{Pick-a-Pic$\uparrow$} & \multicolumn{4}{c}{DrawBench$\uparrow$} & \multirow{2}{*}{NFEs $\downarrow$} \\
\cmidrule(lr){2-5} \cmidrule(lr){6-9}
\textbf{Method} & HPS v2  & AES & PickScore & IR  & HPS v2  & AES  & PickScore  & IR & \\
\midrule
& \multicolumn{8}{c}{\textbf{Stable Diffusion 2.1}} & \\
DDIM & 25.34 & 5.649 & 20.67 & 0.194 & 24.90 & 5.410 & 21.39 & 0.046 & 1.00 \\
Resampling & 26.00 & 5.648 & 20.78 & 0.322 & 25.49 & 5.385 & 21.52 & 0.205 & 2.00 \\
Z-Sampling & 26.29 & 5.653 & 20.71 & 0.346 & 25.95 & 5.465 & 21.47 & 0.296 & 3.00 \\
SOP-1 & 26.34 &	5.684 &	20.85 &	0.735 &25.78&	5.426&	21.64&	0.637 & 3.00\\
SOP-4 & 27.23 & 5.700 & \textbf{21.12} & 1.113 & 26.72 & 5.469 & \textbf{21.89} & {1.008} & 9.00 \\
\rowcolor{gray!15} $\mathtt{Ctrl}$-$\mathtt{Z}^\dagger$&  26.44&	5.686&	20.88&	0.720 & 26.15&	5.476&	21.68&	0.650 & 2.77\\
\rowcolor{gray!15}  $\mathtt{Ctrl}$-$\mathtt{Z}^\ddagger$ & \textbf{27.34} & \textbf{5.705} & {21.02} & \textbf{1.138} & \textbf{26.73} & \textbf{5.501} & {21.84} & \textbf{1.025} & 7.72 \\
\midrule
& \multicolumn{8}{c}{\textbf{Hunyuan-DiT}} &\\
DDIM & 30.22 & 6.324 & 22.15 & 1.071 & 28.90 & 5.930 & 22.47 & 0.827 & 1.00 \\
Resampling & 30.20 & 6.338 & 22.14 & 1.112 & 28.70 & 5.912 & 22.49 & 0.904 & 2.00 \\
Z-Sampling & 30.10 & \textbf{6.398} & 22.18 & 1.111 & 28.60 & \textbf{5.997} & 22.49 & 0.899 & 3.00 \\
 SOP-1 & 30.56&	6.333&	22.19&	1.232 & 28.77&	5.902&	22.50	&1.102 & 3.00\\
SOP-4 & {30.81} & 6.303 & 22.22 & \textbf{1.444} & 29.49 & 	5.942	 & 22.58 & 	1.246 & 9.00 \\
\rowcolor{gray!15} $\mathtt{Ctrl}$-$\mathtt{Z}^\dagger$& 30.69&	6.310&	22.24&	1.214 & 29.27&	5.906	&22.51	&1.067	& 2.85 \\
\rowcolor{gray!15}  $\mathtt{Ctrl}$-$\mathtt{Z}^\ddagger$ & \textbf{30.91} & {6.325} & \textbf{22.30} & {1.441} & 
\textbf{29.79} & 5.908 & \textbf{22.62} & \textbf{1.319} & {8.79} 
\\

\bottomrule
\end{tabular}

\vspace{7mm}

\centering
\scriptsize
\caption{Quantitative results of different methods on T2I-CompBench.
}
\label{tab:compbench-results}
\begin{tabular}{lccccc|ccccc}
\toprule
 & \multicolumn{5}{c|}{\textbf{Stable Diffusion 2.1} $\uparrow$} & \multicolumn{5}{c}{\textbf{Hunyuan-DiT} $\uparrow$} \\
\cmidrule(lr){2-6} \cmidrule(lr){7-11}
\textbf{Method} & Color & Shape& Texture & Spatial & Numeracy& Color & Shape & Texture & Spatial  & Numeracy  \\
\midrule
DDIM        & 46.27 & 41.01 & 46.06 & 13.80 & 46.44 & 66.46 & 44.32 & 60.06 & 24.18 & 53.99 \\
Resampling  & 48.67 & 40.99 & 47.46 & 14.21 & 46.76 & 68.86 & 48.58 & 60.86 & 25.29 & 54.06 \\
Z-Sampling  & 50.38 & 42.19 & 48.84 & 13.65 & 47.77 & 68.80 & 48.51 & 60.46 & 22.89 & 54.27 \\
SOP-1 & 53.85	&45.00	&51.53&18.25	&50.81& 71.01&	49.11&	63.33&	21.00&	54.51 \\
SOP-4         & 59.64 & 48.91 & 60.56 & 16.97 & 52.85 & 73.74 & 53.74 & 64.85 & 21.78 & \textbf{57.58} \\
\rowcolor{gray!15} \texttt{Ctrl}-\texttt{Z}$^\dagger$ &58.65&	47.10&	57.75&	18.55&	51.83 & 71.10&	51.59&	63.76&	21.13&	56.35 \\
\rowcolor{gray!15} 
\texttt{Ctrl}-\texttt{Z}$^\ddagger$ 
            & \textbf{61.26} & \textbf{53.97} & \textbf{62.24} & \textbf{19.29} & \textbf{53.73} 
            & \textbf{73.77} & \textbf{54.99} & \textbf{66.45} & \textbf{25.43} & {56.69} \\
\bottomrule
\end{tabular}
\end{table}

\subsection{Experiment Settings}

We evaluate our method on three representative text-to-image benchmarks: \textbf{Pick-a-Pic}~\cite{pick}, \textbf{DrawBench}~\cite{imagen}, and \textbf{T2I-CompBench}~\cite{huang2023t2icompbench} covering real-world diversity, compositional complexity, and large-scale open-world scenarios. 
Evaluations are conducted using four metrics: \textbf{HPSv2}~\cite{wu2023human}, \textbf{PickScore}~\cite{pick}, \textbf{ImageReward (IR)}~\cite{ir}, and \textbf{Aesthetic Score (AES)}~\cite{aes_laion}. All experiments are conducted on SD-2.1-base \cite{StructureDiffusion} and Hy-DiT \cite{hunyuandit}. 
We use $T=50$ inference steps with CFG (scale 5.5), an exploration window $\lambda=40$, $N=4$ candidates, $d_{\max}=3$, $\delta=0$, and ImageReward as the reward model.
We compare \mbox{\textbf{\abb{}}} against standard \textbf{DDIM}, \textbf{Resampling}~\cite{resample}, \textbf{Z Sampling}~\cite{zigzag}, and \textbf{SOP}~\cite{diffusion_tts} under settings with comparable number of function evaluations (NFEs), which is reported as the average number of denoiser forward passes per denoising step. 
Unless specified otherwise, \ab{} uses the above default setting; in the main results we denote this default as $\mathtt{Ctrl}$-$\mathtt{Z}^\ddagger$ and the low-NFE variant as $\mathtt{Ctrl}$-$\mathtt{Z}^\dagger$, where $\lambda=30$, $N=2$, and $d_{\max}=3$.
More details are available in the Supplementary \Cref{sec:setting}.

\subsection{Quantitative Evaluations}

The quantitative results in \Cref{tab:sampling-results,tab:compbench-results} show consistent improvements over baselines brought by \abb{}. In particular, our method achieves significant gains on ImageReward across both Pick-a-Pic and DrawBench. 
\rev{Compared to SOP, which also relies on ImageReward, $\mathtt{Ctrl}$-$\mathtt{Z}$ Sampling achieves higher HPSv2 and PickScore, indicating stronger improvements in human-aligned image quality. The smaller gain on ImageReward itself is likely due to SOP’s always-on search strategy, which repeatedly optimizes the surrogate ImageReward score at each step and thus over-optimizes the reward model without consistently improving overall quality.}
\Cref{tab:compbench-results} shows that \abb{} performs well on CompBench, whose prompts emphasize object relationships and visual attribute binding. \abb{} adaptively increases exploration strength to revisit and escape early-stage plateaus that lock in coarse suboptimal layout. In contrast, prior methods rely on fixed perturbations and struggle to revise suboptimal trajectories once coarse structures are formed. We leave some further analysis on the results in Supplementary \Cref{sec::further}.

Importantly, $\mathtt{Ctrl}$-$\mathtt{Z}$ provides a simple test-time scaling scheme with a controllable compute--quality trade-off.
By tuning exploration depth and width, it can operate in a low-cost regime at roughly $3\times$ NFEs that exceeds other baselines across most metrics, while increasing the budget to about $7$--$9\times$ NFEs yields further improvements.
Compared to SOP at similar budgets, $\mathtt{Ctrl}$-$\mathtt{Z}$ achieves comparable overall performance with fewer NFEs, and its adaptive depth escalation enables deeper exploration when shallow candidates fail.
At comparable NFE budgets, $\mathtt{Ctrl}$-$\mathtt{Z}$ benefits from adaptive inversion depth, enabling deeper trajectory revisions than fixed-depth exploration and resulting in consistently stronger performance across both small and large compute regimes.

Results on both SD2.1 and Hy-DiT confirm the effectiveness of \abb{} across U-Net and Transformer–based diffusion models, showcasing its compatibility with major frameworks. While our experiments focus primarily on latent diffusion, the hill-climbing intuition naturally extends to pixel-space models such as EDM \cite{Karras2022edm}, where denoising similarly seeks \rev{regions of less noise, higher quality, and better alignment with the input conditions}. We expect \abb{} to remain effective in such settings, leaving direct validation to future work due to computational constraints.

\subsection{Ablations}
The following sections include ablation experiments on the effectiveness of various design and hyper-parameters used in \abb{}. 
Additional ablation experiments regarding the max inversion steps, accept threshold, the exploration initiation criteria, and the effect of exploration window can be found at the Supplementary \Cref{sec::more_abla}.

\subsubsection{Scaling Effect of Exploration Depth and Width.}
\begin{figure*}[t]
    \centering

    \begin{subfigure}[t]{0.25\textwidth}
        \includegraphics[width=\linewidth]{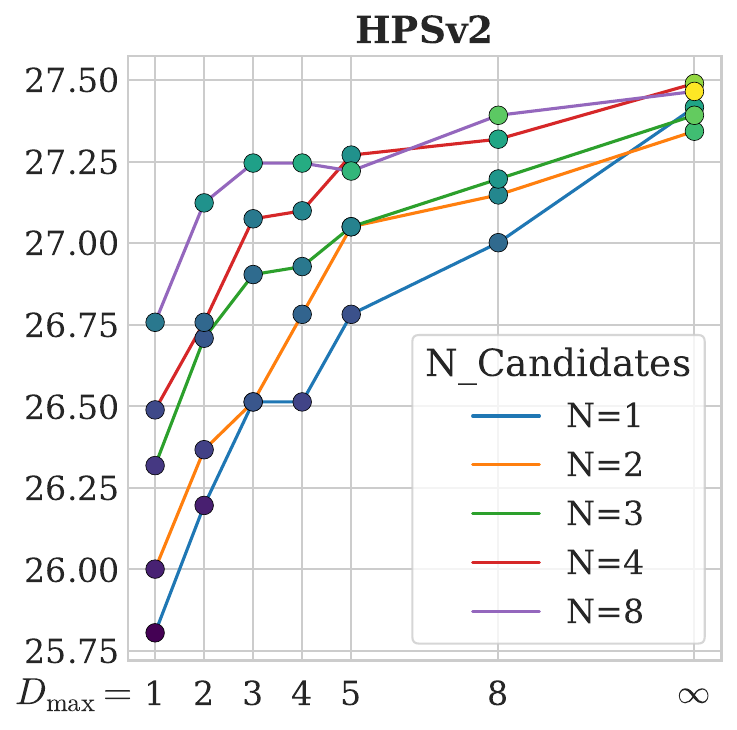}
    \end{subfigure}%
    \begin{subfigure}[t]{0.25\textwidth}
        \includegraphics[width=\linewidth]{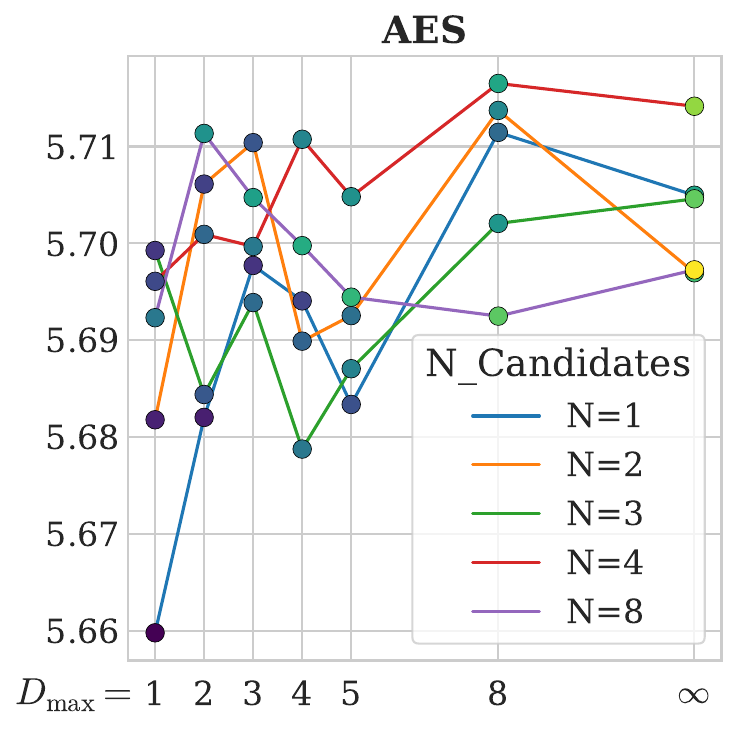}
    \end{subfigure}%
    \begin{subfigure}[t]{0.25\textwidth}
        \includegraphics[width=\linewidth]{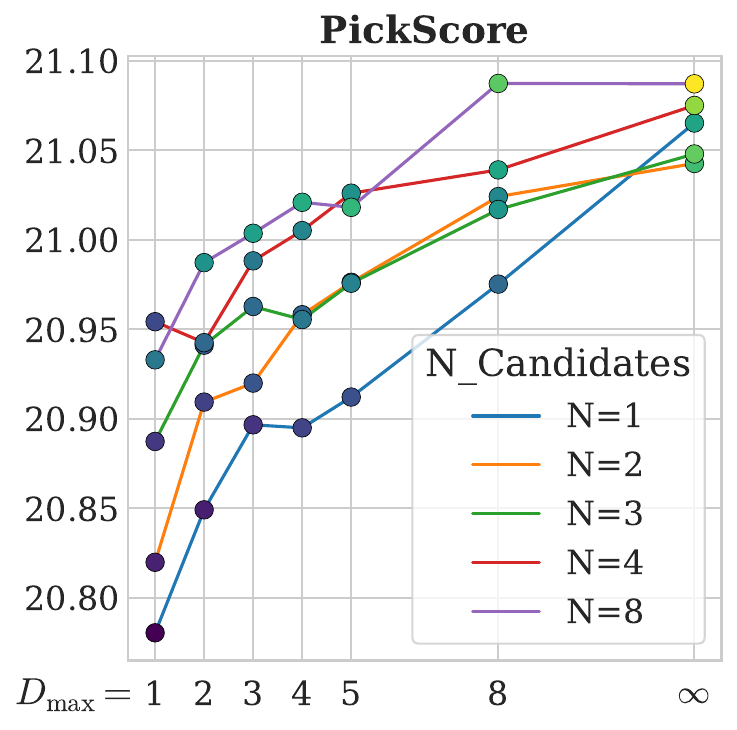}
    \end{subfigure}%
    \begin{subfigure}[t]{0.25\textwidth}
        \includegraphics[width=\linewidth]{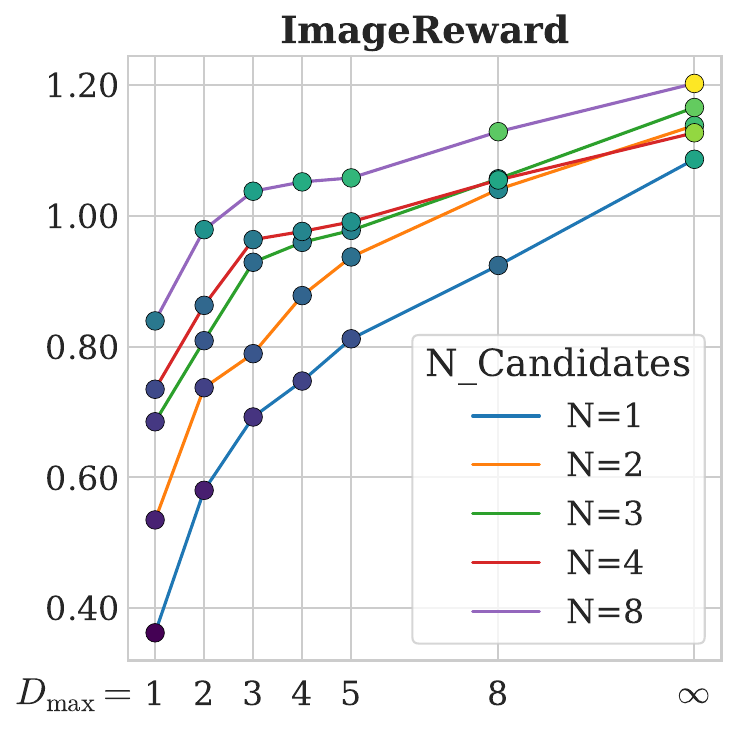}
    \end{subfigure}

    \caption{Effect of exploration depth ($d_{\max}$) and candidate width ($N$). 
    Each marker aggregates results over the same prompts and seeds for a fixed $(d_{\max},N)$.
    Lighter colors indicate higher realized average NFEs.
    \abb{} benefits from both deeper and wider exploration, with increased depth sometimes yielding stronger gains than additional candidates under similar NFEs.
    Comparisons are best viewed zoomed in.}
    \label{fig:metrics_across_depths}
    
\end{figure*}

We investigate how the maximum exploration depth $d_{\max}$ and the number of candidates $N$ per inversion step affect generation quality across evaluation metrics (We fix $\lambda$ as 30 here for efficiency concerns). 
As shown in \Cref{fig:metrics_across_depths}, increasing $d_{\max}$ improves performance across most metrics, which is consistent with enabling revisions to suboptimal early trajectory decisions, while increasing $N$ raises the chance of identifying a higher-scoring alternative at each triggered step.
Overall, depth and width complement each other in improving alignment and perceived quality.

We observe that deeper exploration can reduce reliance on large candidate sets.
When $d_{\max}$ is small, increasing $N$ tends to help by providing more local alternatives per step.
When $d_{\max}$ is large, further increasing $N$ yields less consistent gains, suggesting diminishing returns from widening the candidate set once the method can already escape via deeper inversions.
Under comparable NFEs, deeper-but-narrower settings often outperform wider-but-shallow settings, as indicated by points with similar colors in \Cref{fig:metrics_across_depths}.
This supports the importance of adaptive depth escalation for compute-efficient scalable inference.

As exploration budgets increase, improvements are consistent across most metrics.
AES is a partial exception, since it measures aesthetic preference that is not directly optimized by ImageReward-based selection, and therefore may correlate less with the exploration objective that measures the conditional alignment and structure correctness.
While larger-scale settings are beyond the scope of this study due to resource constraints, the observed trends suggest that \abb{} can scale effectively by trading off depth and width.

\subsubsection{Choice of Reward Model.}
We evaluate the impact of different reward models using all employed metrics and also CLIPScore \cite{clipscore}, with results presented in \Cref{tab:ablas}.
Compared to the DDIM baseline reported in \Cref{tab:sampling-results}, \abb{} improves most metrics under a wide range of surrogate reward choices, indicating that its effectiveness is not tied to a specific scorer. Crucially, for each surrogate reward, the remaining evaluation metrics are held out from the exploration loop. The predominantly positive changes on these non-optimized measures across both benchmarks indicate that Ctrl-Z Sampling improves generation quality beyond simply maximizing the selected scorer.

In particular, using AES as the reward model increases aesthetic scores, but has limited impact on alignment-sensitive metrics since it does not evaluate condition faithfulness.
Among CLIP-derived scorers, PickScore tends to outperform CLIPScore when used for exploration, while overall PickScore and ImageReward yield stronger results than other tested objectives.

We adopt ImageReward in the main experiments due to its broad applicability, efficiency, and reduced benchmark coupling, since PickScore is trained on data closely related to Pick-a-Pic.
Notably, the choice of $R$ also affects realized NFEs by changing the trigger and acceptance dynamics, yielding different compute--performance profiles.
While~\cite{diffusion_tts} suggests that combining multiple reward models can further improve performance, we leave this to future work.

\begin{table*}[t!]
\caption{Ablation results on exploration guided by different reward models and exploration initiation criteria. PS and IR denote PickScore and ImageReward, respectively. Best results are \textbf{bolded}; second-best are \underline{underlined}.}
\label{tab:ablas}
\centering
\small
\begin{tabular}{lcccc|cccc|r}
\toprule
 & \multicolumn{4}{c}{Pick-a-Pic$\uparrow$} & \multicolumn{4}{c}{DrawBench$\uparrow$} & \multirow{2}{*}{NFEs $\downarrow$} \\
\cmidrule(lr){2-5} \cmidrule(lr){6-9}
\textbf{Method} & HPS v2  & AES  & PS  & IR & HPS v2 & AES  & PS & IR  & \\
\midrule
\multicolumn{9}{l}{\textbf{Reward Model for Controlled Exploration}} \\
CLIPScore &
  26.59 & 5.614 & 20.97 & 0.573 &
  26.29 & 5.333 & 21.70 & 0.483 & 11.75 \\
HPS v2 &
  \underline{27.39}&	5.723&	\underline{21.05}&	0.503 &
  \textbf{27.14} & \underline{5.503} & 21.80 & 0.478&3.44\\
AES &
  26.29&	\textbf{6.147}&	20.87&	0.306 &
  25.54&	\textbf{5.902}&	21.62&	0.231 & 11.17\\
PickScore &
  \textbf{27.51}&	\underline{5.763}&	\textbf{21.69}&	\underline{0.589} &
  \underline{27.12}&	5.497&	\textbf{22.38}&	\underline{0.573} & 7.01\\
\rowcolor{gray!15}ImageReward &
  27.34&	5.705&	21.02&	\textbf{1.138}&
   {26.73} & {5.501} & \underline{21.84} & \textbf{1.025} & 7.72 \\
\midrule 
\multicolumn{9}{l}{\textbf{Exploration Initiation Criteria}}
 \\
{Always ($p=1.0$) } & \textbf{27.86}& 	\textbf{5.737}	& \textbf{21.10}& 	\textbf{1.317} &
\textbf{27.17}&	5.482	&\textbf{21.91}	&\textbf{1.174} & 16.72
  \\
{Random ($p=0.5$)} &
  27.00&	\underline{5.713}&	\underline{21.03}	&1.062&
  26.71	&{5.469}	&21.83&	0.906 &  7.81
  \\
 \rowcolor{gray!15} Reward-Based &
   \underline{27.34}&	5.705&	21.02&	1.138&
   \underline{26.73} & \underline{5.501} & \underline{21.84} & \underline{1.025} & 7.72 \\
\bottomrule
\end{tabular}

\end{table*}

\subsubsection{When to Explore.}
This section evaluates different strategies for triggering zigzag exploration. We compare the proposed reward-based initiation mechanism, which activates exploration upon detecting a reward plateau, against two baselines: \textit{Always}, which performs exploration at every denoising step, and \textit{Random}, which triggers exploration at each step with probability $p = 0.5$.

As shown in \Cref{tab:ablas}, the Always strategy yields the highest generation quality but incurs substantial computational overhead, averaging 16.72× NFEs. 
In contrast, the reward-based strategy attains competitive performance at much lower cost ($7.72\times$ NFEs), offering a stronger quality--compute trade-off.
The Random strategy provides modest quality gains with a similar computational budget (7.81× NFEs), 
though the probability can be flexibly tuned to interpolate performance between never and always exploring.
Overall, the proposed strategy delivers improved sampling quality relative to cost and remains more efficient than the Always strategy. Notably, all three strategies are effective within the \abb{} framework and can be selected based on the desired trade-off between performance and inference cost.

\subsection{Sample Diversity Analysis}
Reward-guided inference-time scaling can potentially improve precision by concentrating probability mass on a smaller set of high-scoring outputs, at the cost of reduced sample diversity. We therefore evaluate distributional quality and diversity on a COCO subset with 2k images using Stable Diffusion 2.1. Table~\ref{tab:div} reports FID, precision, recall, and LPIPS under the default DDIM sampler and two Ctrl-Z configurations with increasing inference budgets.

\begin{table}[t]
\centering
\caption{Sampling diversity analysis on COCO-2k with Stable Diffusion 2.1. Ctrl-Z improves distributional quality and precision while preserving diversity close to the DDIM baseline.}
\label{tab:div}
\begin{tabular*}{\linewidth}{@{\extracolsep{\fill}}lccccc@{}}
\toprule
\textbf{Method} & FID$\downarrow$ & Precision$\uparrow$ & Recall$\uparrow$ & LPIPS$\uparrow$ & NFEs$\downarrow$ \\
\midrule
DDIM & 40.22 & 71.20 & 55.85 & 75.00 & 1.00 \\
$\mathtt{Ctrl}$-$\mathtt{Z}^{\dagger}$ & 40.10 & 73.95 & 54.80 & 75.71 & 2.71 \\
$\mathtt{Ctrl}$-$\mathtt{Z}^{\ddagger}$ & 39.98 & 75.10 & 55.20 & 75.54 & 7.54 \\
\bottomrule
\end{tabular*}
\end{table}

As the inference budget increases, Ctrl-Z improves both distributional quality and precision. Importantly, these gains are not accompanied by systematic diversity degradation: LPIPS remains slightly above the DDIM baseline, while recall stays close to the baseline across both Ctrl-Z configurations. This suggests that the precision improvements primarily arise from better visual quality and prompt alignment, rather than from a monotonic precision--recall trade-off caused by mode concentration.

This behavior is consistent with the design of \ab{}. Unlike methods that draw a large pool of independently completed samples and apply final best-of-(N) reranking, \ab{} performs bounded trajectory-level selection only when the reward trajectory reaches a plateau. Its exploration therefore revises locally suboptimal denoising paths rather than globally filtering outputs toward a narrow set of reward-preferred modes. Within the evaluated compute range, the improvements in precision primarily reflect better visual quality and prompt alignment, while preserving intra-prompt diversity comparable to standard DDIM sampling.

\subsection{Qualitative Evaluations}
\begin{figure*}[t!]
  \centering

  \scriptsize

  \begin{minipage}[t]{0.05\textwidth}
      \rotatebox{90}{\parbox{2.1cm}{\centering a vase on the \\ \textbf{right}  of a cat}}
    \end{minipage}
\begin{subfigure}[t]{0.185\textwidth}
  \centering
  \includegraphics[width=\linewidth, trim=0pt 0pt 0pt 0pt, clip]{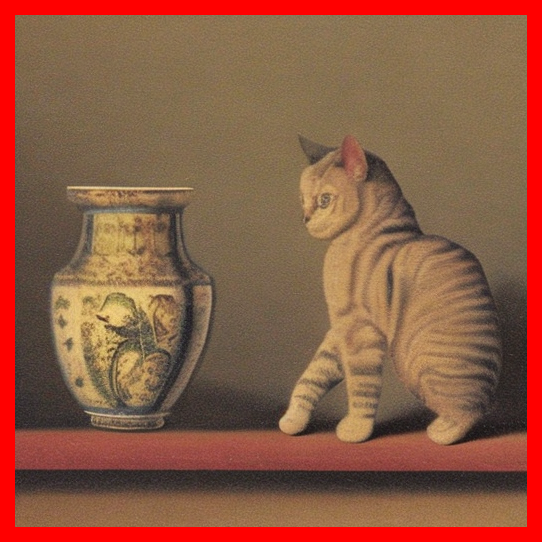}
\end{subfigure}\hspace{1pt}%
\begin{subfigure}[t]{0.185\textwidth}
  \centering
  \includegraphics[width=\linewidth, trim=0pt 0pt 0pt 0pt, clip]{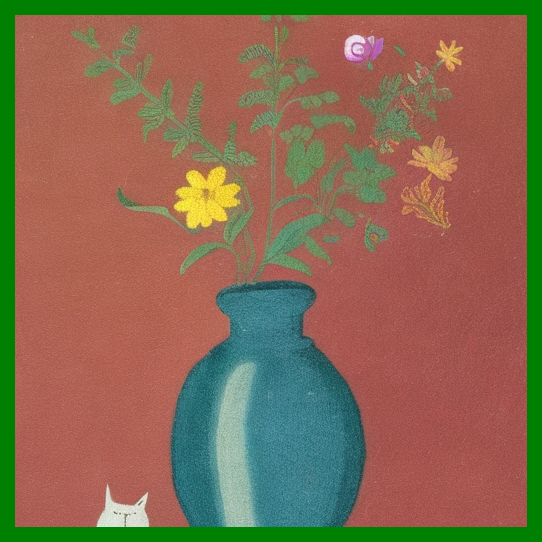}
\end{subfigure}\hspace{1pt}%
\begin{subfigure}[t]{0.185\textwidth}
  \centering
  \includegraphics[width=\linewidth, trim=0pt 0pt 0pt 0pt, clip]{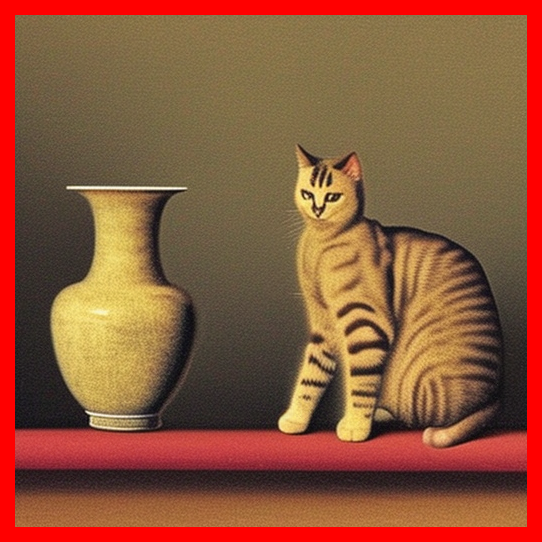}
\end{subfigure}\hspace{1pt}%
\begin{subfigure}[t]{0.185\textwidth}
  \centering
  \includegraphics[width=\linewidth, trim=0pt 0pt 0pt 0pt, clip]{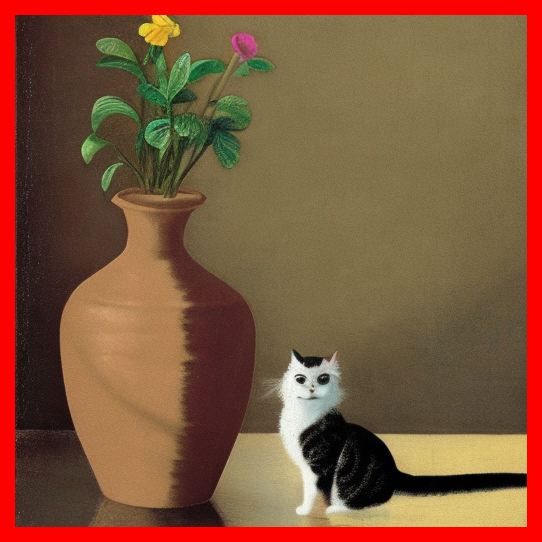}
\end{subfigure}\hspace{1pt}%
\begin{subfigure}[t]{0.185\textwidth}
  \centering
  \includegraphics[width=\linewidth, trim=0pt 0pt 0pt 0pt, clip]{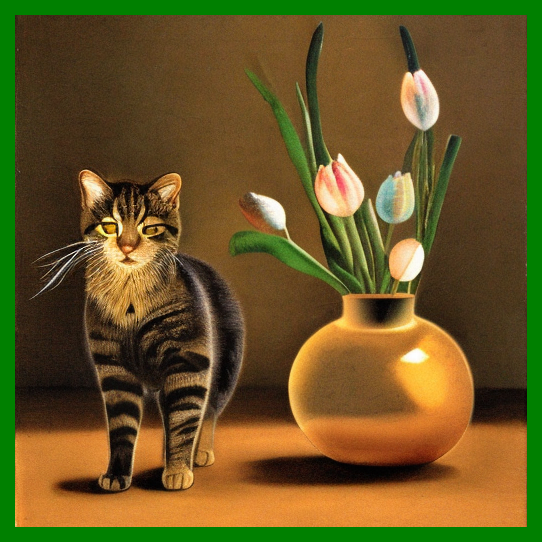}
\end{subfigure}

    \begin{minipage}[t]{0.05\textwidth}
      \rotatebox{90}{\parbox{2.1cm}{\centering \textbf{A blue bird} \& \\a brown bear}}
    \end{minipage}
  \begin{subfigure}[t]{0.185\textwidth}
    \centering
    {\includegraphics[width=\linewidth]{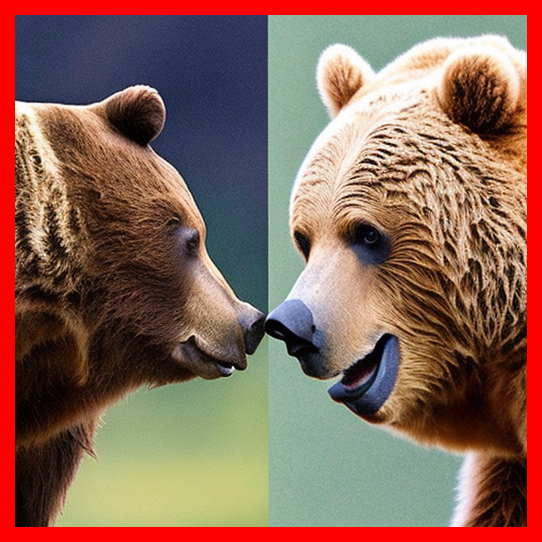}}
  \end{subfigure}\hspace{1pt}%
  \begin{subfigure}[t]{0.185\textwidth}
    \centering
    \includegraphics[width=\linewidth]{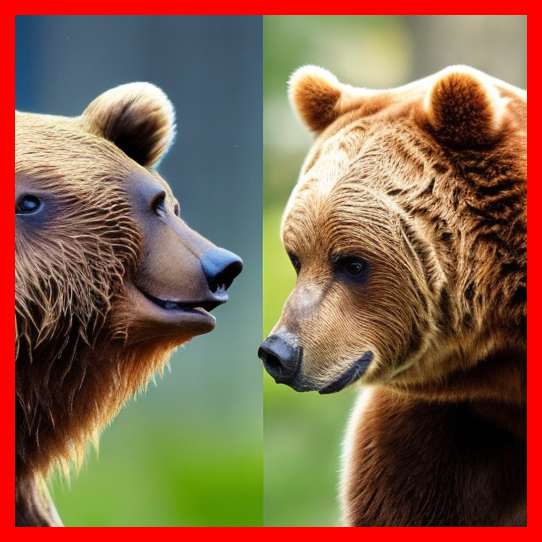}
  \end{subfigure}\hspace{1pt}%
  \begin{subfigure}[t]{0.185\textwidth}
    \centering
    \includegraphics[width=\linewidth]{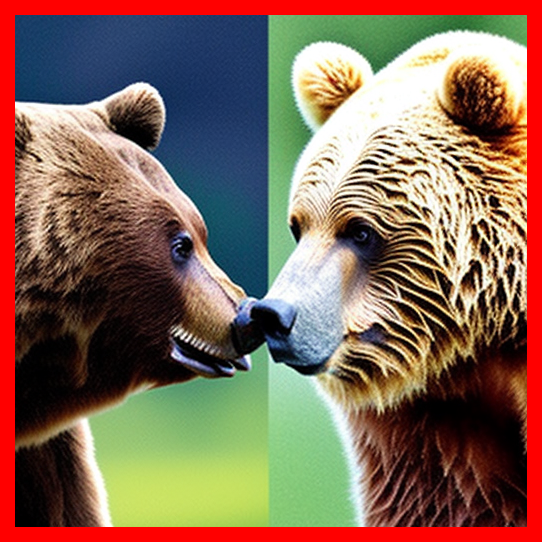}
  \end{subfigure}\hspace{1pt}%
  \begin{subfigure}[t]{0.185\textwidth}
    \centering
    \includegraphics[width=\linewidth]{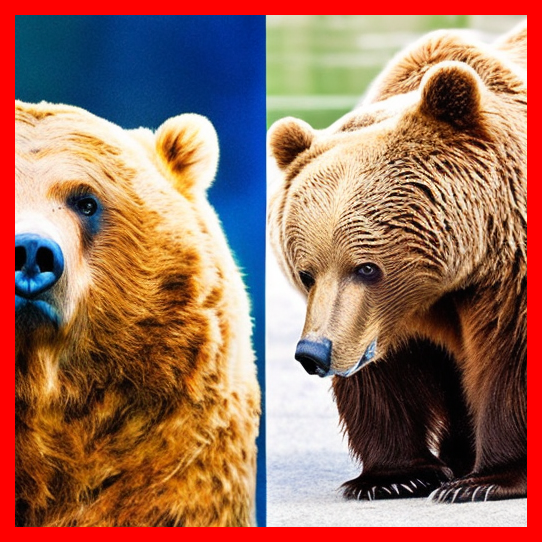}
  \end{subfigure}\hspace{1pt}%
  \begin{subfigure}[t]{0.185\textwidth}
    \centering
    \includegraphics[width=\linewidth]{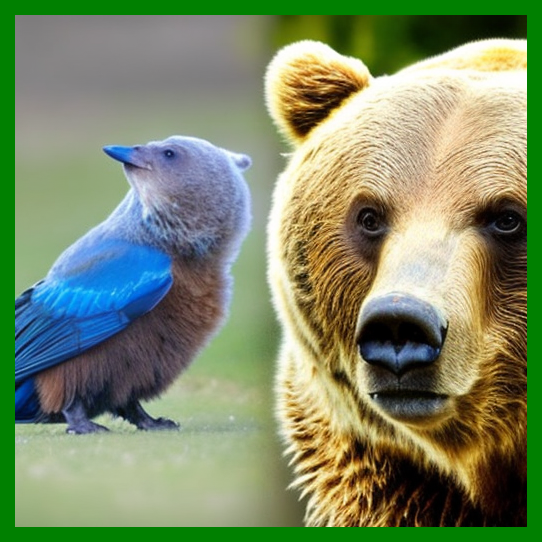}
  \end{subfigure}

  \begin{minipage}[t]{0.05\textwidth}
      \rotatebox{90}{\parbox{2.1cm}{\centering \textbf{two} pillows \\ and \textbf{one} lemon}}
    \end{minipage}
  \begin{subfigure}[t]{0.185\textwidth}
    \centering
    \includegraphics[width=\linewidth]{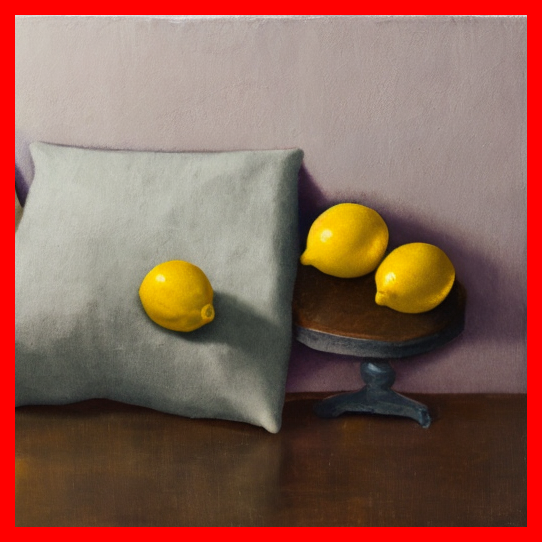}
  \end{subfigure}\hspace{1pt}%
  \begin{subfigure}[t]{0.185\textwidth}
    \centering
    \includegraphics[width=\linewidth]{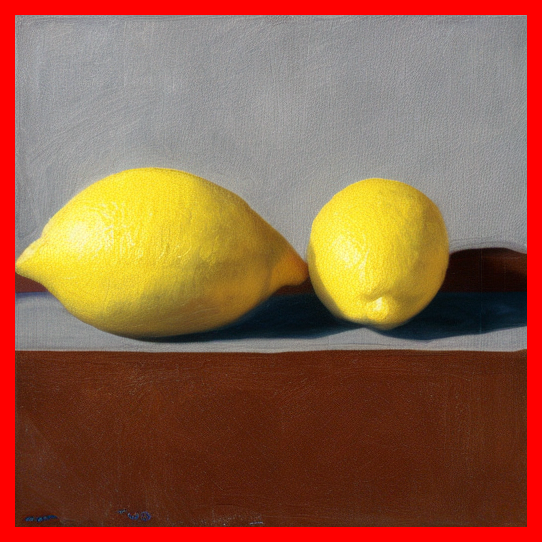}
  \end{subfigure}\hspace{1pt}%
  \begin{subfigure}[t]{0.185\textwidth}
    \centering
    \includegraphics[width=\linewidth]{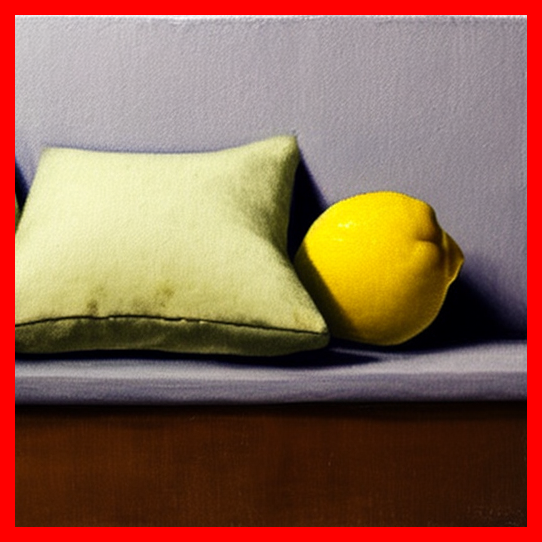}
  \end{subfigure}\hspace{1pt}%
  \begin{subfigure}[t]{0.185\textwidth}
    \centering
    \includegraphics[width=\linewidth]{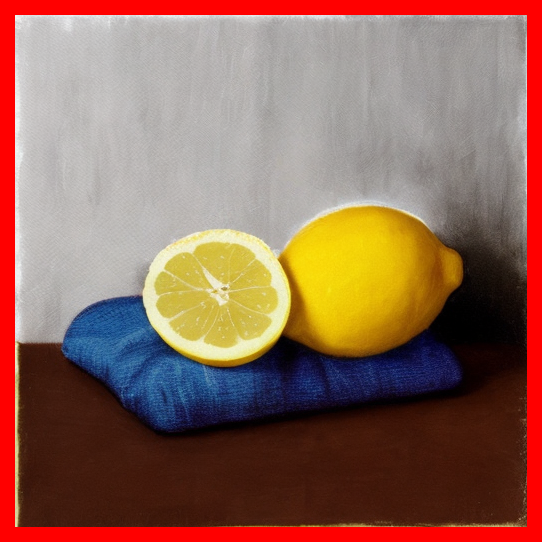}
  \end{subfigure}\hspace{1pt}%
  \begin{subfigure}[t]{0.185\textwidth}
    \centering
    \includegraphics[width=\linewidth]{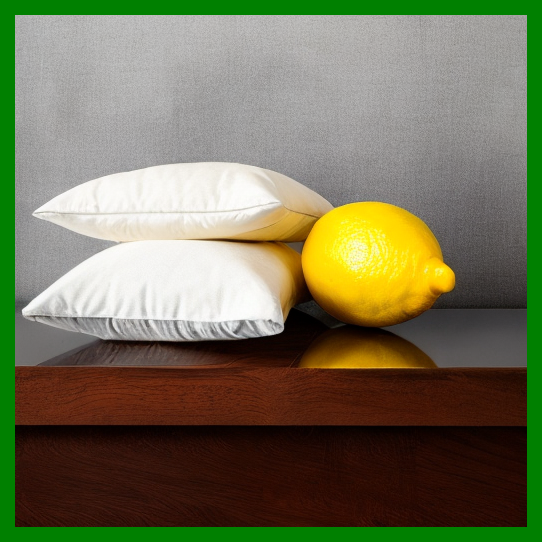}
  \end{subfigure}

  \begin{minipage}[t]{0.05\textwidth}
      \rotatebox{90}{\parbox{2.1cm}{\centering A dog \\ \textbf{driving} a bus}}
    \end{minipage}
  \begin{subfigure}[t]{0.185\textwidth}
    \centering
    \includegraphics[width=\linewidth]{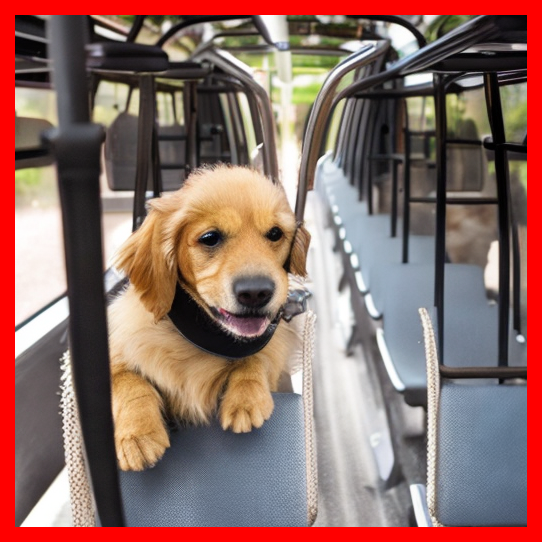}
    \textbf{DDIM} \par
  \end{subfigure}\hspace{1pt}%
  \begin{subfigure}[t]{0.185\textwidth}
    \centering
    \includegraphics[width=\linewidth]{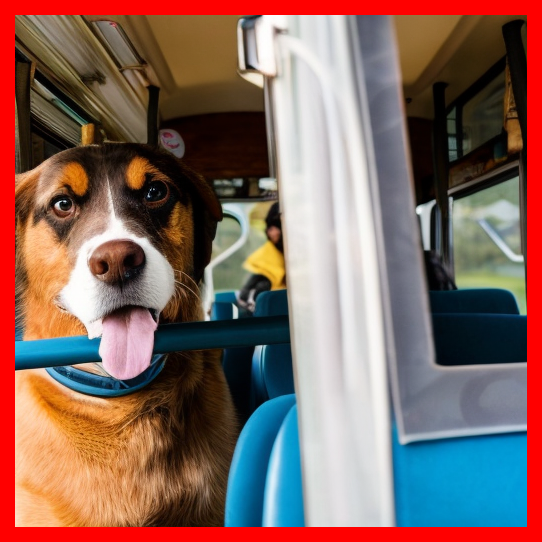}
    \textbf{Resampling} \par
  \end{subfigure}\hspace{1pt}%
  \begin{subfigure}[t]{0.185\textwidth}
    \centering
    \includegraphics[width=\linewidth]{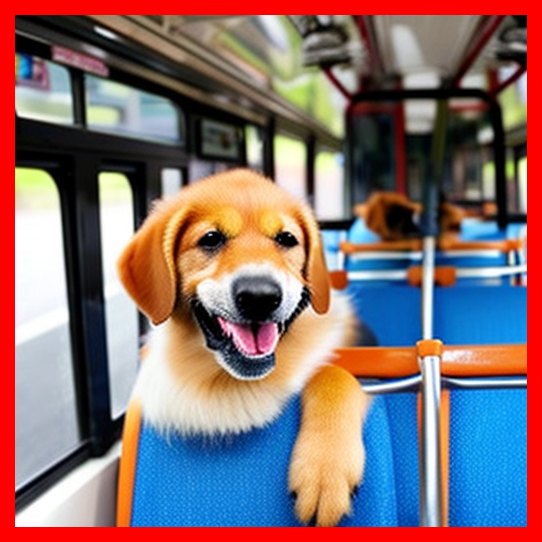}
    \textbf{Z-Sampling} \par
  \end{subfigure}\hspace{1pt}%
  \begin{subfigure}[t]{0.185\textwidth}
    \centering
    \includegraphics[width=\linewidth]{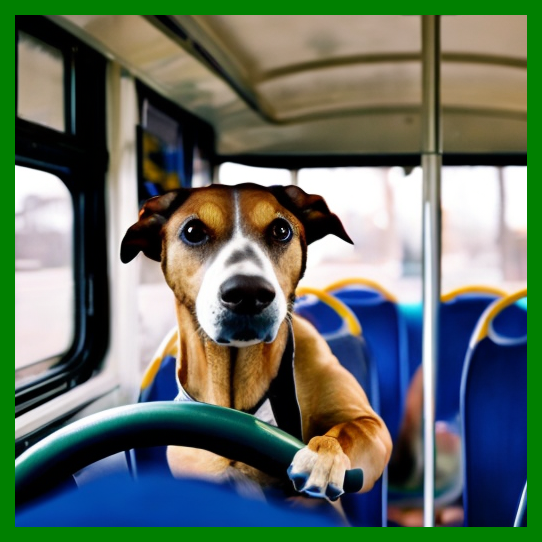}
    \textbf{SOP} \par
  \end{subfigure}\hspace{1pt}%
  \begin{subfigure}[t]{0.185\textwidth}
    \centering
    \includegraphics[width=\linewidth]{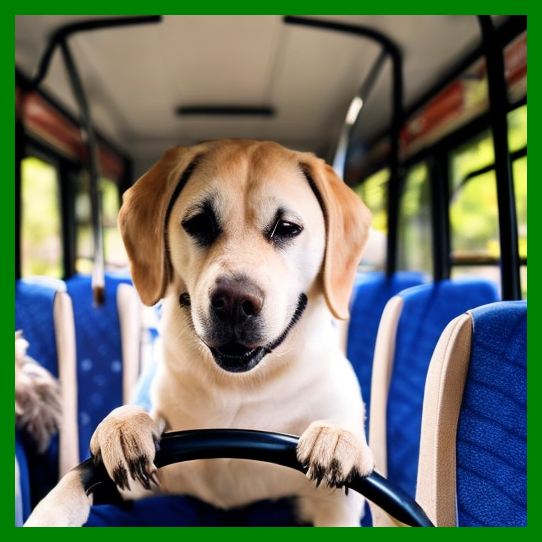}
    \textbf{$\mathtt{Ctrl}$-$\mathtt{Z}$ (Ours)} \par
  \end{subfigure}

  \caption{Qualitative comparison of different sampling methods with SD-2.1-base. 
  Generated images that align with the input condition (shown on the left) are highlighted with \textit{green bounding boxes}, while erroneous or suboptimal images are marked in \textit{red}. 
  }
  \label{fig:method_comparison}
\end{figure*}

\Cref{fig:method_comparison} presents qualitative comparisons between baseline methods and the proposed \abb{}. \abb{} produces outputs that are both visually coherent and semantically aligned with input prompts, as reflected in examples involving spatial relations, numeracy, color, and action understanding. In contrast, baseline methods often produce results that are either semantically misaligned with the prompt or lack visual coherence.
While SOP can improve image quality by incurring a larger candidate pool, we observe failures in challenging cases. The fixed exploration strength makes SOP vulnerable to low-frequency errors in the latent space, leading to local reward gains from object presence. Consequently, the trajectory often converges to visually plausible but semantically imprecise states (\eg, “cat and vase” or “blue and bear”). In contrast, \abb{} adaptively increases exploration strength, allowing the model to escape such traps and generate more aligned content with distinct low-frequency components.
These findings underscore the effectiveness of \abb{} in escaping suboptimal generations through adaptive exploration.
Further qualitative examples and analysis can be found at Supplementary \Cref{sec:sup_qualitative}.

\section{Conclusion}
We present \abb{}, an inference-time scaling sampling strategy for diffusion models that improves text-to-image generation by escaping plateaus and local optima in the surrogate quality landscape. \abb{} detects stagnation and performs controlled inversion with adaptively increased depth using a surrogate quality score. Experiments on text-to-image benchmarks show consistent gains in conditioning alignment and visual fidelity in different NFE regimes, offering a practical compute--quality trade-off without modifying the base denoiser. Future work includes validating \abb{} in broader, higher-budget inference-time scaling settings and developing reward-model-agnostic schedules that jointly adapt exploration initiation and strength.

%
%
\bibliographystyle{splncs04}
\bibliography{main}

\input{supp.tex}

\end{document}

%% file: supp.tex
\clearpage
\appendix
\begin{center}
\Large\bfseries Ctrl-Z Sampling: Scaling Diffusion Sampling with Controlled Random Zigzag Explorations \\
\vspace{2mm}
\large Appendix
\vspace{1em}
\end{center}

\section{Extended Related Works}\label{sec::extend_related_work}

\subsection{Efficient Diffusion Sampling}
Aside from improving generation quality, a complementary line of research focuses on accelerating diffusion sampling. One major approach to this problem is model distillation, where a teacher model is distilled into a faster student \cite{salimans2022progressive, katzir2024noisefree, consistency_models}. Alternatively, advanced sampling through higher-order ODE integrations and scheduling schemes have been proposed to achieve high-quality outputs with only a few diffusion steps \cite{Karras2022edm, lu2022dpm, lu2025dpm, zhang2023fast, tong2025learning}.
Some works also explore parallel sampling \cite{shih2023parallel, chen2024accelerating}, which improves efficiency by reformulating denoising steps to run concurrently, trading space for time.
Inspired by speculative decoding in LLMs, \cite{bortoli2025accelerated} further introduces drafting strategies that efficiently generate multiple candidate trajectories and verify them in parallel. Overall, these methods primarily trade off image quality and memory against inference speed, whereas our method aligns more closely with inference scaling, leveraging additional computations during inference only to improve quality. For this reason, we do not include diffusion accelerators in our comparisons.

\subsection{Inference Scaling}
Rather than allocating additional computation during pretraining \cite{brown2020language, kaplan2020scaling, hoffmann2022training}, recent advances in LLMs have highlighted the potential of inference scaling (or test-time scaling (TTS)) as a paradigm to enhance performance during deployment \cite{zhang2025and}. Foundational techniques like chain-of-thought prompting encourage step-by-step reasoning to elicit better outputs \cite{wei2022chain}, while self-consistency methods generate multiple responses in parallel and aggregate them via voting \cite{wang2023self}. Sequential approaches, such as self-refinement, iteratively revise drafts based on self-feedback \cite{madaan2023self, mao2025through}, and hybrid strategies combine exploration and exploitation, as in ReAct \cite{yao2023react}, ToT \cite{yao2023tree}, and GoT \cite{besta2024graph}. More advanced internal scaling trains models to autonomously extend reasoning \cite{openai2024learning, deepseek2025deepseek}, often via bootstrapping \cite{zelikman2022star} or post-training enhancements \cite{muennighoff2025s1} with algorithms like proximal policy optimization \cite{schulman2017proximal}.

\subsection{Reward-Guided Inference Search}
Empirical studies reveal scaling laws at inference: increasing test-time compute (\eg, via repeated sampling or deeper search) yields predictable gains akin to pretraining scaling but in a more efficient manner \cite{brown2024large, snell2025scaling}. 
A verifier, \ie, reward model, that scores the quality of intermediate or final outputs is typically used to scale inference by re-ranking or steering candidates.
Early frameworks show how verifiers can guide reasoning in LLMs \cite{cobbe2021training, creswell2023selectioninference}.
Process reward models (PRMs) enable finer-grained control by scoring intermediate steps \cite{lightman2024lets, uesato2022solving, setlur2025rewarding, zhang2025generative}, inspiring PRM-guided greedy, beam, BoN, and tree search methods \cite{zhang2024restmcts,jiang2024rewardTreeSearch, gandhi2024stream, jinnai2025regularizedBoN, yang2025multi}. More recent work explores efficiency-oriented strategies such as speculative rejection \cite{sun2024fast} and list-wise re-ranking \cite{gangi2024first}.
Together, these advances demonstrate how inference-time scaling with verifiers can allow smaller models to outperform LLMs with significantly larger parameter sizes \cite{liu2025can}. 

While these ideas originated in LLM reasoning, similar inference scaling has been studied for diffusion \cite{sohl2015deep, ddpm, song2019generative} and flow models \cite{lipman2023flow, rect_flow, flow3}. Prior efforts optimize denoising trajectories through search \cite{diffusion_tts, evo_search, classical} or steer generation with reward-guided objectives \cite{fk_steering}, improving sample quality, diversity, and alignment at test time without retraining. However, most methods build the candidate pool only through shallow and local perturbations, leaving the search vulnerable to local optima. In contrast, \abb{} adaptively deepens exploration into higher-noise manifolds when stagnation is detected, enabling targeted search where needed and avoiding unnecessary computation.
Most existing inference-scaling studies characterize scaling laws under extremely large search budgets. Due to resource constraints, we limit our evaluation to moderate budgets, and leave broader large-scale scaling analyses as an important direction for future work.

\section{Experiment Settings} \label{sec:setting}
\subsection{Datasets}
We evaluate different sampling methods on three text-to-image diffusion benchmarks: \textbf{Pick-a-Pic} \cite{pick}, \textbf{DrawBench} \cite{imagen}, and \textbf{T2I-CompBench} \cite{huang2023t2icompbench}. These benchmarks are designed to assess the performance of diffusion models in generating images from text prompts, each with distinct characteristics to test various aspects of model capabilities.
\\
\textbf{Pick-a-Pic} includes 500 diverse, real-user prompts accompanied by human preference annotations. This benchmark emphasizes the diversity of prompts, reflecting a wide range of user-generated inputs and preferences, making it ideal for evaluating how well models handle varied, real-world scenarios.
\\
\textbf{DrawBench} comprises 200 curated prompts designed to target compositional and semantic challenges. These prompts focus on complex scenarios, such as combining multiple objects or attributes, testing the models' ability to understand and accurately render intricate textual descriptions.
\\
\textbf{T2I-CompBench} \cite{huang2023t2icompbench} features 6,000 compositional text prompts, categorized into attribute binding (\eg, color, shape, texture), object relationships (\eg, spatial and non-spatial), and complex compositions. This benchmark is tailored for evaluating open-world compositional text-to-image generation, pushing models to generalize across nuanced combinations of elements. It includes specialized evaluation metrics and explores the use of multimodal large language models (MLLMs) for assessment, providing a comprehensive framework for testing diffusion models' compositional accuracy.

\subsection{Metrics}

We evaluate our method using four human-aligned metrics commonly adopted for assessing text-to-image diffusion models. 
\\
\textbf{HPSv2}~\cite{wu2023human} and \textbf{PickScore}~\cite{pick} are preference prediction models built on the CLIP~\cite{clip} backbone and fine-tuned on large-scale human preference data. HPSv2 is trained on structured same-prompt image pairs with binary preference labels, enabling robust evaluation of prompt-aligned generation quality. In contrast, PickScore leverages the more diverse and open-ended Pick-a-Pic dataset, which reflects user-generated prompts and preferences collected at scale, thereby improving generalization across prompt styles. Both models outperform the standard CLIPScore~\cite{clipscore} in aligning with human judgments.
\\
\textbf{ImageReward (IR)}~\cite{ir} is a learned reward model trained on expert-annotated image–text pairs, designed to produce a scalar score that reflects alignment with human preferences. Unlike HPSv2 and PickScore, which operate on image pairs, ImageReward evaluates each image–prompt pair independently, assigning a continuous reward typically in the range of –2 to +2. This formulation enables both fine-grained assessment and potential use in reinforcement-style optimization.
\\
\textbf{Aesthetic Score (AES)}~\cite{aes_laion} estimates the perceptual appeal of images based on large-scale aesthetic annotations. While it provides a lightweight and prompt-agnostic measure of visual quality, AES is not conditioned on the input prompt and thus may fail to reflect semantic alignment or faithfulness to the generation condition.

\subsection{Implementation}
We adopt the base version of Stable Diffusion 2.1 \cite{StructureDiffusion} for U-Net-based and Hunyuan-DiT \cite{hunyuandit} for Transformer-based diffusion models, for their efficiency and effectiveness in text-to-image generation. We set the resolution to $512$ for Stable Diffusion and to $1024$ for Hunyuan-DiT, the number of sampling steps to $T = 50$,  All experiments use classifier-free guidance with a scale factor of $5.5$ to ensure stable generation quality.
\rev{For $\mathtt{Ctrl}$-$\mathtt{Z}^\dagger$ and $\mathtt{Ctrl}$-$\mathtt{Z}^\ddagger$, the exploration window is set to $\lambda = 30, 40$, respectively. The maximum number of candidate samples per step is $N=2,4$, and the max search depth is set to ($d_{\max} = 2,3$) to balance quality and efficiency for compute-performance tradeoff under different inference budgets.}
We use ImageReward~\cite{ir} as the reward model across all methods, as it has demonstrated strong alignment with human preference in prior evaluations~\cite{diffusion_tts}. The accept threshold is set to $\delta = 0$. 
Experiments were conducted on two Linux machines (Ubuntu 22.04). The first used an Intel i7-10700 CPU with an NVIDIA RTX 3090 GPU; the second used an AMD Ryzen 9 7950X3D CPU with an NVIDIA RTX 4090 GPU. Both setups ran with CUDA 12.1. Experiments were implemented in Python using PyTorch (v2.4.0) and Diffusers (v0.32.2).

\subsection{Baselines}
We compare \abb{} with several existing sampling strategies, including the standard \textbf{DDIM} sampling~\cite{ddim}, \textbf{Resampling}~\cite{resample}, \textbf{Z Sampling}~\cite{zigzag}, \textbf{Search over Paths (SOP)}~\cite{diffusion_tts}, as introduced in the main paper (\Cref{fig::main}). The details on these methods are as follows:
\\
\textbf{Resampling}~\cite{resample} perturbs the current intermediate representation by reintroducing noise, hence stepping backward in the diffusion process to reach a higher-noise state. From this noisier point, the model continues denoising again, which allows it to better incorporate conditioning information and potentially generate outputs that are more aligned with the conditional input.
\\
\textbf{Z Sampling}~\cite{zigzag} perturbs the current latent by stepping backward to a higher-noise state. At each step, it estimates the forward generation trajectory under weak or no classifier-free guidance, which typically leads to visually plausible but less aligned outputs. It then injects noise in the opposite direction of this estimated denoising path, effectively inverting the model’s default tendency toward local optima. This targeted reversal helps escape alignment traps and improves consistency with the conditional input during subsequent denoising.
\\
\textbf{Search Over Path (SOP)}~\cite{kim2025test} reverses the current latent to a higher-noise state, where it samples multiple perturbed candidates by injecting different noise patterns at each step. Each candidate is then denoised forward, and a reward model evaluates their alignment with the conditional input. The best-performing path is selected to continue generation. This strategy enables SOP to improve sampling by exploring multiple nearby directions and explicitly selecting the most promising one based on reward. The number of search candidates in SOP per step is set to 4 to ensure an overall comparable NFEs to the default settings of \abb{}.

Various other sampling techniques such as \cite{zhao2023unipc, ays} are not included for comparison as they are considered orthogonal and may be combined with the inversion-driven exploration methods discussed here. In \Cref{sec::combine}, we provide studies on how combining other orthogonal sampling techniques with \abb{} will affect performance.

\section{Additional Experiments and Results} \label{sec::more_abla}

\subsection{Further Analysis on Quantitative Results} \label{sec::further}
While the quantitative evaluation in \Cref{tab:sampling-results} primarily demonstrates the effectiveness of \abb{} in guided exploration, several additional observations are noteworthy. 
With Stable Diffusion 2.1, \abb{} guided by ImageReward yields substantial improvements not only in alignment metrics but also in AES. This can be attributed to the weaker baseline quality of SD2.1, where enhancements in human preference alignment are often accompanied by perceptual gains. In contrast, Hunyuan-DiT, with its larger capacity and higher-resolution outputs, already delivers strong aesthetic quality. In this setting, ImageReward, which is trained to capture human preference, primarily emphasizes improvements in alignment rather than aesthetics when comparing candidate states. As a result, applying \abb{} to Hunyuan-DiT consistently improves IR, the guiding metric, but may lead to a slight reduction in AES.

It is also worth noting that, under identical search parameters, \abb{} with Hunyuan-DiT requires about one additional forward call per sampling step on average (8.79 NFEs vs. 7.72 for Stable Diffusion 2.1). A plausible explanation is that Hunyuan-DiT, being a larger and more capable model, produces higher-quality denoising outputs at each step, making it harder for additional exploration to exceed the reward threshold (\ie, $R([c], \hat{x}_0^{\,t-1}) \geq r_{\text{prev}} + \delta$). At first glance, this may seem counterintuitive: a stronger model might be expected to achieve larger reward improvements and thus initiate exploration less often. However, the observed gains in ImageReward for Hunyuan-DiT are smaller than those for SD2.1, indicating that its denoising outputs are already closer to optimal. Consequently, finding further improvements through exploration is more challenging, which results in slightly stronger adaptive explorations with larger inversion step size and also more computation. While \abb{} still outperforms SOP in this setting, the variability in sampling cost introduces a potential weakness compared to SOP, whose cost is fixed. Nonetheless, this issue can be mitigated through strategies such as adaptive pruning or early termination of deeper searches, helping to stabilize inference cost across models of different capacities.

An additional observation from \Cref{tab:compbench-results} is that \abb{} outperforms SOP at most cases except for numeracy prompts on Hunyuan-DiT, where SOP outperforms \abb{}, likely because its fixed local perturbations are sufficient to recover enumerations when small object instances are concentrated in high-resolution outputs. Nonetheless, the proposed dynamic adaptive exploration scheme also incorporates local optimizations, yielding numeracy results comparable to SOP while substantially improving performance on other subsets.
Conversely, SOP performs poorly on spatial prompts. Interestingly, Resampling, which also injects noise more randomly but without reward-based guidance, demonstrates notable improvements in spatial-related prompts. 
This suggests a potential limitation of SOP: its reliance on reward signals at every step may contribute to early reward saturation, where strong reward gains from object presence encourages the model to commit prematurely to locally optimal states. 
Although these states align well with the reward model, they may fall short in capturing the broader spatial relationships intended by the prompt. 
In comparison, the proposed \abb{} method applies reward-guided exploration adaptively with less frequency and increasing strength. The selective exploration reduces early saturation, while the ability to increase exploration strength helps escape locally optimal states and recovers more coherent spatial structures.

\subsection{Max Inversion Steps and Accept Threshold} 

\begin{figure*}[t!]
    \centering

    \begin{subfigure}[t]{0.24\textwidth}
        \includegraphics[width=\linewidth]{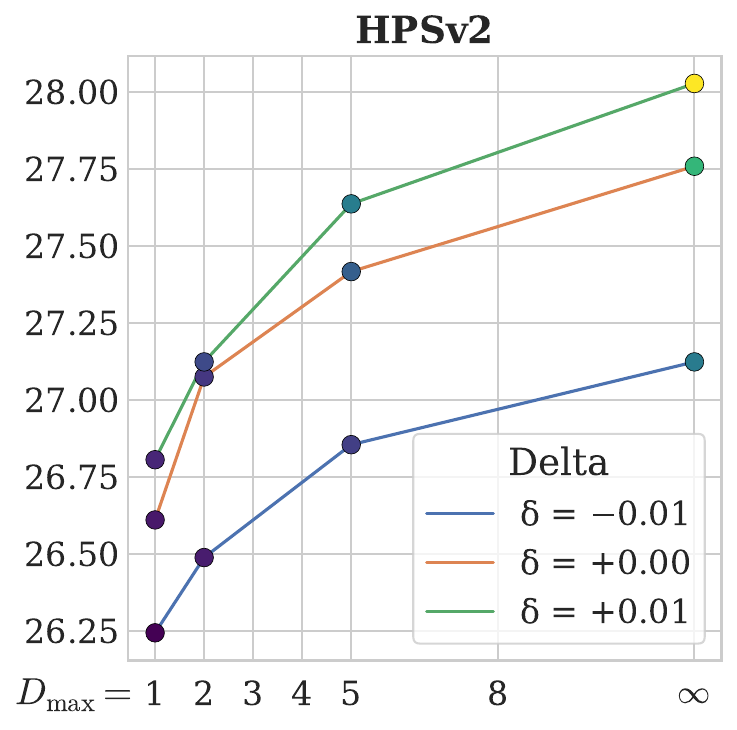}
    \end{subfigure}
    \hfill
    \begin{subfigure}[t]{0.24\textwidth}
        \includegraphics[width=\linewidth]{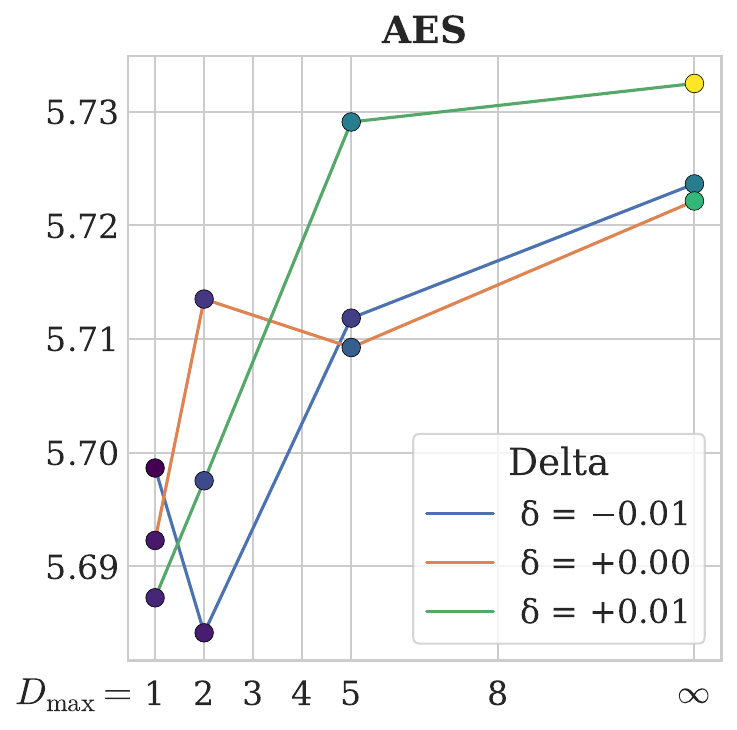}
    \end{subfigure}
    \hfill
    \begin{subfigure}[t]{0.24\textwidth}
        \includegraphics[width=\linewidth]{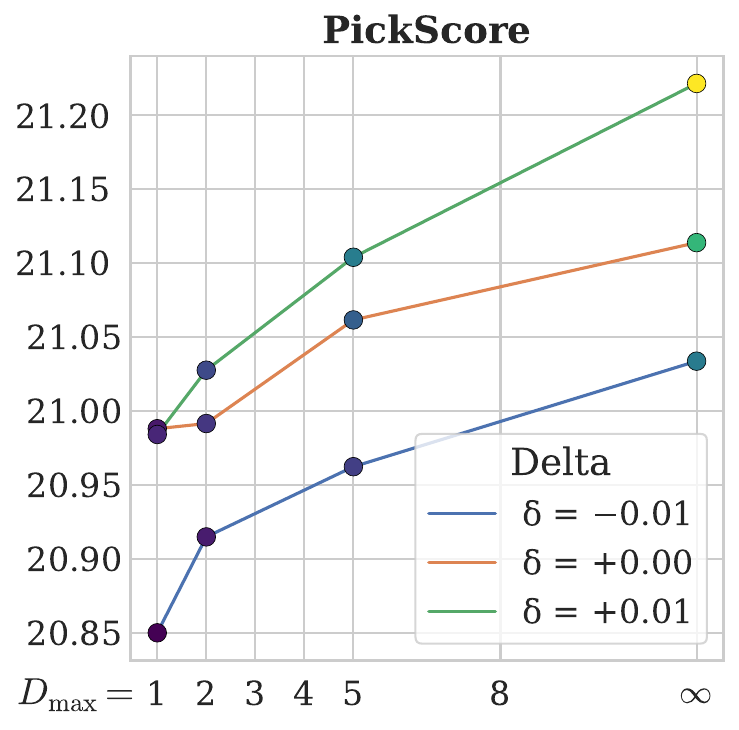}
    \end{subfigure}
    \hfill
    \begin{subfigure}[t]{0.24\textwidth}
        \includegraphics[width=\linewidth]{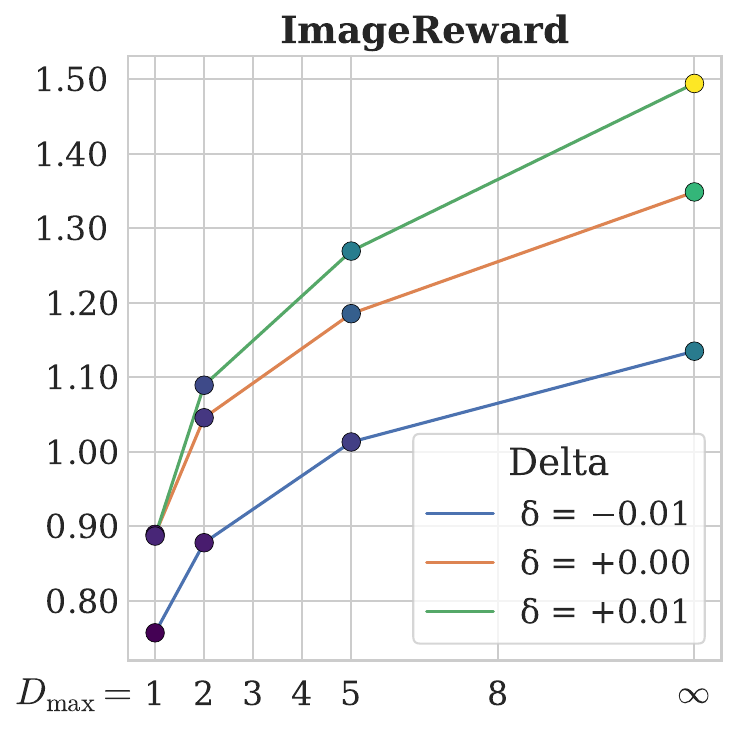}
    \end{subfigure}

    \caption{Evaluation metrics across different accept threshold $\delta = -1, 0, 1$ for different max search depth $d_{max} \in \{1, 2, 5, \infty\}$.}
    \label{fig:depth_delta}
\end{figure*}

This section examines the impact of the acceptance threshold ($\delta$) and maximum inversion depth ($d_{\text{max}}$), with results shown in \Cref{fig:depth_delta}. In our design, $\delta$ controls both when to initiate exploration and whether to accept each inversion step, requiring the reward to exceed that of the previous step by at least $\delta$ (which may be negative to allow slight decreases).

We find that increasing either parameter generally improves generation quality, supporting the effectiveness and scalability of deeper, more permissive zigzag exploration. These improvements align with our adaptive control design for flexible inference-time scaling. Aesthetic Score (AES), however, shows less consistent gains, likely due to its emphasis on visual appeal over conditional alignment. Notably, when $\delta = 0.01$, which enforces stricter acceptance, AES improves more reliably with increased $d_{\text{max}}$. This suggests that stricter thresholds filter out superficial reward increases (such as those resulting from overfitting to prompt-specific details) and instead favor candidate steps that introduce more substantive changes to the generation trajectory. These structural improvements may contribute to better global coherence and visual quality, which are more aligned with what AES measures.

Overall, $\delta$ provides a flexible mechanism for balancing quality and cost. However, setting a fixed stepwise reward improvement as acceptance threshold (as in \Cref{eq::reward}) can introduce instability, especially when early steps yield high rewards, making future acceptance more difficult and potentially causing unnecessary explorations with negligible benefit. Future work could explore global reward scheduling to adaptively stabilize this process.

\subsection{Exploration Initiation Criteria}

\begin{table*}[t!]
\caption{Ablation results on sampling with different exploration initiation criteria under similar number of function evaluations (NFEs) settings.
The Reward-based strategy that initiates exploration on detected plateau demonstrates comparable performance to the ones that `Always' initiate exploration at each step, with overall fewer NFEs. 
}
\label{tab:abla-init-always}
\centering
\small
\begin{tabular}{ccccc|cccc|c}
\toprule
\textbf{Parameters} & \multicolumn{4}{c}{Pick-a-Pic $\uparrow$} & \multicolumn{4}{c}{DrawBench $\uparrow$} & \multirow{2}{*}{NFEs $\downarrow$} \\
\cmidrule(lr){2-5} \cmidrule(lr){6-9}
($d_{max}$, $N$, $\lambda$) & HPS v2& AES & PickScore  & IR  & HPS v2  & AES & PickScore & IR & \\
\midrule
\multicolumn{9}{l}{\textbf{Always Explore}} & \\

(3, 2, 40) & 27.44	 &5.691	 &\textbf{21.09}	 &1.189  &
\textbf{26.98}	 &5.436 &	21.85 &	\textbf{1.076} &
10.00\\
(2, 3, 40) & \textbf{27.54} &	5.683	 &21.07 &	\textbf{1.199} &
27.10	&5.459	&\textbf{21.93}	&1.049 &9.46\\
(1, 4, 40) &  27.05	 &5.689	 &21.00 &	1.057  &
26.59 &	5.484	 &21.78 &	0.813  &7.24\\
(3, 3, 30)& 27.34	&5.704	&21.04&	1.132&
26.98&	5.469&	21.86&	0.949 & 9.90\\
\midrule
\multicolumn{9}{l}{\textbf{Reward-based} (Ours)} & \\
(3, 4, 40) & {27.34} & \textbf{5.705} & {21.02} & {1.138} & {26.73} & \textbf{5.501} & {21.84} & {1.025} & 7.72 \\
\bottomrule
\end{tabular}

\end{table*}

\begin{table*}[t!]

\caption{
Ablation results for randomly initiating the exploration process with varying probabilities. $p = 0$ corresponds to standard DDIM sampling, while $p = 1$ is equivalent to the `{Always}' exploration strategy. Increasing the probability consistently improves generation quality, accompanied by a corresponding increase in NFEs.
}
\label{tab:abla-init-prob}
\centering
\small
\begin{tabular}{lcccc|cccc|c}
\toprule
{\textbf{Exploration}}  & \multicolumn{4}{c}{Pick-a-Pic $\uparrow$} & \multicolumn{4}{c}{DrawBench $\uparrow$} & \multirow{2}{*}{NFEs $\downarrow$} \\
\cmidrule(lr){2-5} \cmidrule(lr){6-9}
{\textbf{Probability}} & HPS v2  & AES  & PickScore & IR  & HPS v2 & AES  & PickScore  & IR & \\
\midrule
$p=0$&  25.34 & 5.649 & 20.67 & 0.194 & 24.90 & 5.410 & 21.39 & 0.046 & 1.00
\\
$p=0.1$ &25.83	&5.677&	20.78	&0.474&
25.29	&5.445	&21.52	&0.347& 2.12
\\
$p=0.3$ &26.54&	5.705	&20.89	&0.801 &
26.37&	5.470	&21.70&	0.758 & 4.68
\\
$p=0.5$ &
27.00&	{5.713}&	{21.03}	&1.062&
26.71	&{5.469}	&21.83&	0.906 &  7.81
\\
$p=0.7$ &27.37	&5.695	&21.04	&1.193&
26.90	&5.471&	21.83	&1.038& 11.11
\\
$p=1$  & \textbf{27.86}& 	\textbf{5.737}	& \textbf{21.10}& 	\textbf{1.317} &
\textbf{27.17}&	\textbf{5.482}	&\textbf{21.91}	&\textbf{1.174} & 16.72
  \\

\bottomrule
\end{tabular}
\end{table*}

\subsubsection{Always Initiate Exploration.}
To more thoroughly evaluate the effectiveness of the proposed reward-based exploration initiation criterion, we conduct a set of comparative experiments using an `Always' explore strategy, where exploration is triggered at every diffusion step. To ensure a fair comparison, we reduce the associated exploration parameters. Specifically, we reduce the maximum exploration depth $d_{\text{max}}$, the number of candidate paths $N$, or exploration window $\lambda$, to achieve a similar average number of function evaluations (NFEs) as those under the proposed settings with SD2.1. The results are presented in \Cref{tab:abla-init-always}.

Despite the reduced exploration budget, most configurations under the `Always' strategy still result in higher NFEs on average. While they offer slightly improved prompt alignment, our proposed method consistently achieves better aesthetic scores. This suggests that always triggering exploration may lead to excessive optimization of reward values, potentially overfit to prompt-specific or reward-specific features and degrading visual quality. In contrast, the reward-based initiation strategy selectively activates exploration, striking a better balance between conditional alignment and perceptual quality. This indicates its robustness in avoiding over-optimization and improving aesthetic appeal.

\subsubsection{Randomly Initiate Exploration.}
Moreover, we also study the effect of randomly initiate the exploration process with different probabilities, the results are demonstrated in \Cref{tab:abla-init-prob}. As the probability increases, generation quality improves across all metrics, including aesthetic score (AES), with $p = 1.0$ achieving the best performance. However, this also leads to substantially higher NFEs.

In contrast to our budget-controlled comparison with the `Always' strategy, the random initiation setting allows NFEs to scale freely. As such, the observed gains at higher $p$ are largely attributable to increased computational budget rather than the effectiveness of the initiation strategy itself. While less efficient than our reward-based method, random initiation can serve as a flexible alternative when computational resources are not a primary constraint.

\subsubsection{Effect of Exploration Window $\lambda$.}

\begin{table*}[t!]
\caption{Quantitative results under different $\lambda$ values for ImageReward-guided controlled exploration. NFE indicates the average number of function evaluations. Overall, generation quality improves as the $\lambda$ value increases.}
\label{tab:lambda-step}
\small
\centering
\begin{tabular}{lcccc|cccc|c}
\toprule
 & \multicolumn{4}{c}{Pick-a-Pic$\uparrow$} & \multicolumn{4}{c}{DrawBench$\uparrow$} & \multirow{2}{*}{NFEs $\downarrow$} \\
\cmidrule(lr){2-5} \cmidrule(lr){6-9}
\textbf{$\lambda$} & HPS v2 & AES  & PickScore & IR& HPS v2 & AES  & PickScore  & IR  & \\
\midrule
0 / 50 & 25.34 & 5.649 & 20.67 & 0.194 & 24.90 & 5.410 & 21.39 & 0.046 & 1.00 \\
10 / 50 & 26.05 & 5.669 & 20.81 & 0.478 & 25.70 & 5.415 & 21.52 & 0.374 & 2.21 \\
20 / 50 & 26.64 & 5.693 & 20.93 & 0.748 & 26.20 & 5.442 & 21.67 & 0.669 & 3.83 \\
30 / 50 & 27.08 & 5.700 & 20.99 & 0.964 & 26.60 & 5.435 & 21.78 & 0.874 & 5.67 \\
\rowcolor{gray!15} 40 / 50 & 27.34 & 5.705 & 21.02 & 1.138 & 26.73 & 5.501 & 21.84  &1.025 & 7.72 \\
50 / 50 & 27.29 & 5.714 & 21.03 & 1.286 & 26.80 & 5.468 & 21.80 & 1.170 & 10.19 \\
\bottomrule
\end{tabular}
\end{table*}

Table~\ref{tab:lambda-step} reports the effect of varying the exploration window $\lambda$. As shown, increasing $\lambda$ generally improves performance, particularly in terms of the reward model metric (IR), which directly guides exploration. However, the performance gains gradually diminish as $\lambda$ becomes large, with some metrics even exhibiting slight degradation. Notably, the number of additional NFEs also increases substantially under larger $\lambda$ values.

We attribute this behavior to the tendency of late-stage explorations to `overfit' to the reward model. When the latent representation has already committed to local optima, \abb{} exhaustively searches for marginal reward improvements, which leads to increased computational cost and potential performance drop on metrics not directly optimized by the reward. This resembles classical overfitting, where excessive model flexibility results in reduced generalization and efficiency. 
On the other hand, limiting exploration to the early stages, with smaller $\lambda$ values, can be viewed as initiating exploration before the trajectory descends too far into a particular basin of the loss landscape. At this point, the generation remains relatively flexible and is less constrained by local optima. As a result, \abb{} is more likely to discover alternative directions that lead toward better optima, enabling more efficient and effective convergence.
Based on this trade-off, we adopt $\lambda = 40$ as a default, which achieves decent aesthetic quality besides condition alignment, as well as a favorable balance between performance and computational overhead.

\subsection{Additional Baseline Comparisons}

\begin{table}[t]
\centering
\caption{
Additional baseline comparisons on Pick-a-Pic with Stable Diffusion 2.1.
$\dagger$ and $\ddagger$ denote lower- and higher-compute configurations, respectively.
NFEs denote the average number of denoiser forward evaluations per sampling step.
}
\label{tab:additional_baselines}
\vspace{-2mm}
\small
\setlength{\tabcolsep}{4.2pt}
\renewcommand{\arraystretch}{1.05}
\begin{tabular}{@{}lccccc@{}}
\toprule
Method
& HPS v2$\uparrow$
& AES$\uparrow$
& PickScore$\uparrow$
& IR$\uparrow$
& NFEs$\downarrow$ \\
\midrule
DDIM
& 25.34 & 5.649 & 20.67 & 0.194 & 1.00 \\
\midrule
BoN-3
& 26.13 & 5.688 & 20.83 & 0.666 & 3.00 \\
BoN-8
& 26.54 & 5.712 & 21.00 & 1.096 & 8.00 \\
DTS$^\dagger$
& 26.41 & 5.693 & 20.82 & 0.608 & 2.94 \\
DTS$^\ddagger$
& 26.64 & 5.694 & 20.97 & 0.659 & 8.40 \\
DIS$^\dagger$
& 26.48 & 5.678 & 20.89 & 0.734 & 3.00 \\
DIS$^\ddagger$
& 27.33 & 5.718 & 21.07 & 1.186 & 8.00 \\
DAS$^\dagger$
& 26.42 & 5.672 & 20.88 & 0.726 & 3.00 \\
DAS$^\ddagger$
& 27.30 & 5.710 & 21.05 & 1.134 & 8.00 \\
\rowcolor{gray!15} $\mathtt{Ctrl}$-$\mathtt{Z}^{\dagger}$
& 26.44 & 5.686 & 20.88 & 0.720 & 2.77 \\
\rowcolor{gray!15}  $\mathtt{Ctrl}$-$\mathtt{Z}^{\ddagger}$
& 27.34 & 5.705 & 21.02 & 1.138 & 7.72 \\
\bottomrule
\end{tabular}
\vspace{-2mm}
\end{table}

To further position \ab{} among recent scalable inference-time methods for diffusion models, we compare it with Best-of-$N$ (BoN), Diffusion Tree Sampling (DTS)~\cite{dts}, classical-search-based diffusion inference scaling (DIS)~\cite{classical}, and Diffusion Alignment as Sampling (DAS)~\cite{kim2025test}. These methods allocate additional inference compute through different search mechanisms, complementing the trajectory-search baselines considered in the main paper. Table~\ref{tab:additional_baselines} reports results on Pick-a-Pic with Stable Diffusion 2.1 under matched low- and high-compute regimes.

BoN scales inference by sampling independent initial noise latents and reranking final outputs. DTS instead constructs a tree of trajectory continuations, but is less competitive in our setting and shows limited gains when the compute budget increases. DIS and DAS both achieve strong higher-budget results; however, like BoN, they maintain or resample a population of particles initialized from multiple noise seeds. Their gains therefore combine trajectory-level search with inter-latent population scaling. In contrast, \ab{} fixes a single initial latent and allocates computation to adaptive intra-trajectory refinement, progressively rolling back and re-denoising only when reward progress stalls. Thus, \ab{} represents a complementary scaling axis to BoN, DIS, and DAS. At comparable budgets, \ab{} achieves competitive performance against the particle-based variants on diffusion inference scaling.

\subsection{Combining \ab{} with Additional Optimization Methods} \label{sec::combine}
\abb{} enhances diffusion sampling by inverting to earlier timestep with random perturbations and selecting promising candidates using reward-based metrics. It is complementary to sampling methods that optimize the forward process or scheduling, and can be seamlessly combined with them. In this section, we study its integration with two representative approaches: AYS sampling \cite{ays}, which improves scheduling, and CFG++ \cite{chung2024cfg++}, which refines classifier-free guidance with manifold constraints. We do not examine forward-process variations here, as experimenting with different diffusion architectures already constitutes a form of forward improvement.

\subsubsection{\abb{} + DPM/AYS Sampling.}
DPM-Solver \cite{lu2022dpm} accelerates diffusion sampling by exploiting the semi-linear structure of the probability flow ODE. Instead of discretizing both linear and nonlinear terms as in generic solvers, it solves the linear component analytically and applies exponential integrator techniques to approximate the nonlinear term. The first-order variant (similar to DDIM) uses a single noise prediction per step, while the practical second-order solver introduces an intermediate state to refine the estimate and substantially reduce discretization error.

\begin{table*}[t!]
\caption{Quantitative evaluation results when combining  \abb{} with AYS sampling on Pick-a-Pic and DrawBench.}
\label{tab:ays-main}
\small
\centering
\begin{tabular}{lcccc|cccc}
\toprule
 & \multicolumn{4}{c}{Pick-a-Pic$\uparrow$} & \multicolumn{4}{c}{DrawBench$\uparrow$} \\
\cmidrule(lr){2-5} \cmidrule(lr){6-9}
\textbf{Method} & HPS v2 & AES  & PickScore & IR & HPS v2 & AES  & PickScore  & IR \\
\midrule
DPM & 23.08 & 5.478 & 20.20 & -0.203 & 22.92 & 5.190 & 20.888 & -0.312 \\
AYS & 23.46 & 5.460 & 20.26 & -0.145 & 23.34 & 5.190 & 20.942 & -0.191 \\
\rowcolor{gray!15}  $\mathtt{Ctrl}$-$\mathtt{Z}$ + DPM  & 24.23 & \textbf{5.494} & 20.39 & 0.158 & \textbf{24.17} & \textbf{5.240} & \textbf{21.104} & 0.088 \\
\rowcolor{gray!15}  $\mathtt{Ctrl}$-$\mathtt{Z}$ + AYS  & \textbf{24.94} & 5.450 & \textbf{20.42} & \textbf{0.433} & 24.06 & 5.170 & 20.995 & \textbf{0.146} \\
\bottomrule
\end{tabular}
\end{table*}

\begin{table}[t!]
\centering
\caption{Quantitative evaluation results when combining  \abb{} with AYS sampling on T2I-CompBench.
}
\label{tab:ays-compbench-results}
\setlength{\tabcolsep}{6pt}
\begin{tabular}{lccccc}
\toprule
\textbf{Method} & Color $\uparrow$ & Shape $\uparrow$ & Texture $\uparrow$ & Spatial $\uparrow$ & Numeracy $\uparrow$ \\
\midrule
DPM & 35.98 & 36.53 & 42.41 & 9.26 & 43.50 \\
AYS & 35.96 & 36.71 & 42.66 & 8.87 & 42.98 \\
\rowcolor{gray!15}  $\mathtt{Ctrl}$-$\mathtt{Z}$ + DPM & 37.78 & 37.87 & 43.14 & \textbf{10.70} & 44.47 \\
\rowcolor{gray!15}  $\mathtt{Ctrl}$-$\mathtt{Z}$ + AYS & \textbf{39.02} & \textbf{39.73} & \textbf{46.68} & 10.24 & \textbf{45.47} \\
\bottomrule
\end{tabular}

\end{table}

AYS (Align Your Steps) \cite{ays} is a method for optimizing diffusion sampling schedules. Instead of relying on hand-crafted heuristics like cosine or polynomial timesteps, it formulates schedule selection as an optimization problem using stochastic calculus, yielding solver-specific schedules that significantly improve output quality, especially in few-step regimes.

\paragraph{Implementation.}

We adopt the AYS schedule based on the 10-step second-order DPM solver \cite{lu2022dpm}, as it is the default scheduler for AYS and the only one publicly available. At inference, AYS is implemented by replacing the default diffusion timesteps with the optimized schedule. Since the AYS schedule is model-specific, we use Stable Diffusion 1.5 in this section to align with the released parameters. For \abb{}, we set the inversion strength to match the schedules in DPM or AYS and apply identical formulations for $x_0$ prediction and other steps as described in the main text.

With only 10 sampling steps, each step yields substantial reward gains toward the final image, making it difficult to design an effective exploration threshold $\delta$. Under the default criterion ($\delta = 0$), exploration would rarely be triggered, as subsequent steps almost always improve reward. Therefore, in these experiments we initiate exploration at every step.

\paragraph{Results and Discussion.}
Quantitative results on Pick-a-Pic, DrawBench, and CompBench for combining \abb{} with DPM and AYS are reported in \Cref{tab:ays-main,tab:ays-compbench-results}. By replacing the default scheduler with learned timesteps, AYS yields substantial improvements over DPM on Pick-a-Pic and DrawBench. However, its advantage is less evident on CompBench, where DPM achieves stronger performance on the Spatial and Numeracy subsets. 
These trends remain largely consistent when either scheduler is combined with \abb{} and showing improved performance, supporting our expectation that \abb{} enhances diffusion sampling by escaping local optima and remains orthogonal to the choice of scheduling strategy.

While \abb{} improves generation quality on both DPM and AYS, the gains are smaller than with standard 50-step DDIM sampling, particularly on CompBench. This is largely due to the coarse step size in the DPM solver: each step removes substantial noise and already yields strong reward gains. As a result, even when exploration is initiated, deeper adaptive exploration is seldom triggered, since higher-reward alternatives are often accessible with only one backward step. In contrast, with more sampling steps, as in DDIM, some denoising updates differ only marginally in direction, making exploration more effective by forcing the model to search harder for alternative states and thereby escape local optima. On CompBench, where capturing structure-relevant low-frequency information is critical, this explains why \abb{} contributes less when paired with large-step solvers. Although raising the acceptance threshold $\delta$ could partially alleviate the issue, tuning it under coarse schedules remains challenging. Overall, these factors account for the diminished gains when combining \abb{} with 10-step AYS and highlight the need for more flexible scheduling of exploration initiation and acceptance criteria.

\subsubsection{\abb{} + CFG++.}

\begin{table*}[t!]
\caption{Quantitative evaluation results when combining  \abb{} with CFG++ on Pick-a-Pic and DrawBench. $\mathtt{Ctrl}$-$\mathtt{Z} \uparrow$ indicates the $\mathtt{Ctrl}$-$\mathtt{Z}$ experimented with larger exploration parameters.}
\label{tab:cfg-main}
\small
\centering
\begin{tabular}{lcccc|cccc}
\toprule
 & \multicolumn{4}{c}{Pick-a-Pic$\uparrow$} & \multicolumn{4}{c}{DrawBench$\uparrow$} \\
\cmidrule(lr){2-5} \cmidrule(lr){6-9}
\textbf{Method (DDIM)} & HPS v2  & AES & PickScore & IR  & HPS v2  & AES  & PickScore  & IR \\
\midrule
\midrule
+ CFG & 25.34 & 5.649 & 20.67 & 0.194 & 24.90 & 5.410 & 21.39 & 0.046 \\
+ CFG++ & 25.78&	5.675&	20.72&	0.276 &  25.39&	5.452&	21.46	&0.189 \\
\rowcolor{gray!15} + $\mathtt{Ctrl}$-$\mathtt{Z}$ $+$ CFG & {27.34} & \textbf{5.705} & {21.02} & {1.138} & {26.73} & \textbf{5.501} & \textbf{21.84} & {1.025} \\
\rowcolor{gray!15} + $\mathtt{Ctrl}$-$\mathtt{Z}$ $+_{a}$ CFG++  & 26.68&	5.675&	20.88&	0.968 & 26.00 & 5.470 & 21.63 & 0.753  \\
\rowcolor{gray!15} + $\mathtt{Ctrl}$-$\mathtt{Z}$   $+_{p}$ CFG++ &\textbf{27.44}	&5.696	&\textbf{21.02}&	\textbf{1.217}& \textbf{26.86}&	5.434&	21.81&	\textbf{1.101} \\ 
\bottomrule
\end{tabular}
\end{table*}

\begin{table}[t!]
\centering
\caption{Quantitative evaluation results when combining  \abb{} with CFG++ on T2I-CompBench.
}
\label{tab:cfg-compbench-results}
\setlength{\tabcolsep}{6pt}
\begin{tabular}{lccccc}
\toprule
\textbf{Method (DDIM)} & Color $\uparrow$ & Shape $\uparrow$ & Texture $\uparrow$ & Spatial $\uparrow$ & Numeracy $\uparrow$ \\
\midrule
+CFG  & 46.27 & 41.01 & 46.06 & 13.80 & 46.44 \\
+CFG++ & 49.25&	42.89&	46.65&	14.00&	45.75 \\
\rowcolor{gray!15}  + $\mathtt{Ctrl}$-$\mathtt{Z}$ $+$ CFG & {61.26} & {53.97} & {62.24}  & \textbf{19.29} & \textbf{53.73}   \\
\rowcolor{gray!15}  + $\mathtt{Ctrl}$-$\mathtt{Z}$ $+_{a}$ CFG++ & 58.89&	47.85&	57.46&	16.93&	51.03 \\
\rowcolor{gray!15} + $\mathtt{Ctrl}$-$\mathtt{Z} $  $+_{p}$ CFG++ &\textbf{62.76}&	\textbf{54.04}&	\textbf{62.46}&	19.25&	53.65 \\
\bottomrule
\end{tabular}

\end{table}

A known limitation of standard classifier-free guidance (CFG) is that the guided prediction $\epsilon_\theta^t(x_t, c)$ extrapolates away from the unconditional branch. More explicitly, the two outputs can be combined through a guidance weight $\omega$ as:
\begin{equation}
\epsilon_\theta^{t,\omega}(x_t) = \epsilon_\theta^t(x_t,\varnothing) + \omega\big(\epsilon_\theta^t(x_t,c)-\epsilon_\theta^t(x_t,\varnothing)\big),
\end{equation}
and for $\omega > 1$ this extrapolation tends to push the trajectory off the diffusion manifold, resulting in instability and poor inversion. While DDIM with CFG applies the guided noise as follows,
\begin{equation}
\hat x_0 = \frac{x_t - \sqrt{1-\bar\alpha_t}\,\epsilon_\theta^t(x_t, c)}{\sqrt{\bar\alpha_t}}, 
\quad
x_{t-1}^{\text{CFG}} = \sqrt{\bar\alpha_{t-1}}\,\hat x_0 + \sqrt{1-\bar\alpha_{t-1}}\,\epsilon_\theta^t(x_t, c),
\end{equation}
CFG++ \cite{chung2024cfg++} makes a simple but effective change: the clean estimate $\hat x_0$ still leverages the guided noise, but the re-noising step reverts to the unconditional prediction:
\begin{equation}
x_{t-1}^{\text{CFG++}} = \sqrt{\bar\alpha_{t-1}}\,\hat x_0 + \sqrt{1-\bar\alpha_{t-1}}\,\epsilon_\theta^t(x_t,\varnothing).
\end{equation}
Intuitively, this replaces \emph{extrapolation} with \emph{interpolation} between the conditional and unconditional paths, keeping the trajectory closer to the data manifold. The result is more stable sampling and easier inversion, making CFG++ a natural companion to additional guidance mechanisms.

\paragraph{Implementation.}
We evaluate two variants. In $\mathtt{Ctrl}$-$\mathtt{Z}$ $+_{a(ll)}$ CFG++, we replace CFG with CFG++ in all forward computations of \abb{} (\ie, Lines 6 and 19 in \Cref{algo:main}). In $\mathtt{Ctrl}$-$\mathtt{Z}$ $+_{p(artial)}$ CFG++, we apply CFG++ only to the standard denoising steps (Line 6), while retaining traditional CFG for the exploration steps.
We use $\omega = 0.6$ as in the official CFG++ implementation. Experiments are conducted on all benchmarks with all other parameters identical to the main text. 

\paragraph{Results and Discussion.}

The quantitative results are presented in \Cref{tab:cfg-main,tab:cfg-compbench-results}. While CFG++ consistently outperforms traditional CFG across benchmarks, combining \abb{} with CFG++ on \emph{all} denoising steps yields smaller gains than when paired with CFG. This outcome can be explained by two factors. First, CFG++ smooths the denoising trajectory and mitigates off-manifold deviations, which reduces the number of sharp local optima that \abb{} is designed to exploit through inversion and adaptive exploration (effectively constraining exploration by `chaining' the diffusion process to the unconditional denoising direction). Second, CFG++ has its strongest influence in the early denoising stages where \abb{} is also most effective, as structural errors at this stage propagate through the entire trajectory. By stabilizing these early steps, CFG++ preemptively limits the opportunities for \abb{} to discover alternative high-reward trajectories. Together, these factors account for the diminished marginal benefit observed when the two methods are combined.

These deductions are further supported by experiments where \abb{} is combined with CFG++ only \emph{partially}, leaving the exploration denoising steps unmodified. In this setting, the aforementioned limitations are alleviated, as \abb{} can still explore off-manifold regions to identify promising alternative trajectories. This hybrid strategy leverages the strengths of both methods: in-manifold updates from CFG++ are used when reward signals steadily improve, while \abb{} activates off-manifold explorations to escape local optima when progress stalls. Consequently, $\mathtt{Ctrl}$-$\mathtt{Z}$ $+_{p}$ CFG++ achieves stronger overall performance than its counterparts.

\subsection{Distribution of Exploration Steps}
\begin{figure}[t]
    \centering
    \includegraphics[width=\linewidth]{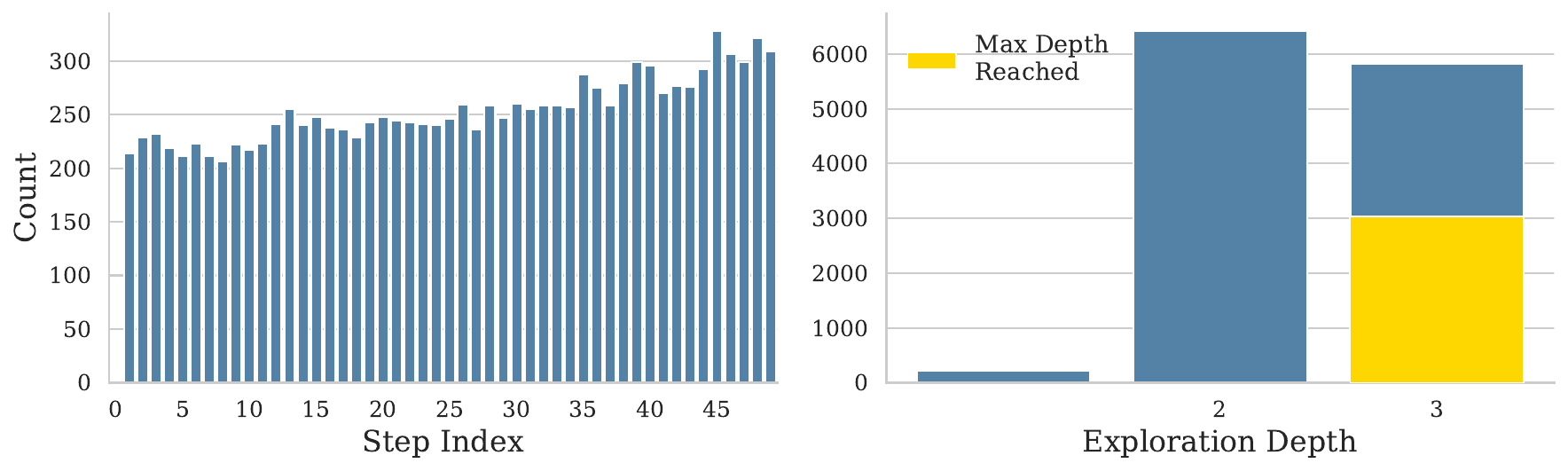}
    \caption{
    \textit{Left}: Count of explorations initiated at each diffusion step. \textit{Right}: Count of final exploration depths when adaptive exploration terminates; the \textit{yellow bar} indicates cases where termination occurs due to reaching the maximum depth limit.
    }
    \label{fig:explore_distribution}
\end{figure}

To analyze the behavior of \abb{}, we record the frequency of exploration initiations across sampling steps and the final inversion depth reached when adaptive exploration terminates. Results are accumulated over 500 prompts with $\lambda = 50$, $N=4$, and $d_{\max}=3$, as shown in \Cref{fig:explore_distribution}. The left plot shows that explorations, triggered by the criterion $R(c, \hat{x}_0^{\,t-1}) < r_{\text{prev}} + \delta$ (\Cref{eq::reward}), are distributed throughout the trajectory but occur more frequently at later steps. We attribute this to reward saturation: as $\hat{x}_0^{\,t-1}$ changes only marginally with larger timesteps, rewards plateau and explorations are triggered more often. This observation highlights the benefit of adopting a smaller $\lambda$, which improves efficiency by avoiding unnecessary late-stage explorations with limited effect. For detailed analyses and visualizations of how explorations impact generation quality at different steps, see \Cref{sec::intermediate}.

From the right plot, we observe that with four exploration candidates, the process rarely terminates after a single inversion step; more often, \abb{} identifies better candidate states with an inversion depth of two. A substantial number of explorations also terminate at the maximum depth of three, often because no more satisfying state is found, yet the search is upper-bounded at the specific depth and terminated (as indicated by the yellow bar). Since computational cost grows linearly with depth, we cap inversion at three steps. Overall, the uneven distribution of exploration depths highlights the importance of adaptive depth, which allocates compute and exploration strength flexibly according to the difficulty of each case.

\section{Limitations and Future Works}
\subsection{Global Scheduling of Exploration Criteria}
A key limitation of the current framework lies in the use of a fixed per-step acceptance threshold $\delta$ to control both the initiation and progression of zigzag exploration. This design assumes that a meaningful improvement in reward is always possible at each step, but in practice, diffusion trajectories often pass through intermediate states that are already near-optimal under the given reward model. In such cases, requiring a fixed threshold improvement may be overly strict.
As a result, exploration may overfit to intermediate states that align well with the specific reward model but fail to yield tangible improvements in overall generation quality.
This limitation can be further exacerbated when a large reward improvement is required too early in the generation process. In such cases, the model may pass through intermediate states that already satisfy prominent reward features, such as object presence or visual plausibility, making further improvements in global structure or semantic alignment difficult to achieve.

Future work could study globally adaptive exploration strategies that adjust explore initiations and depth. For example, the system could track reward gradients, or normalized improvements to determine whether stricter or more permissive criteria should be applied at a given step. This would help the model recognize when it has entered a plateau or near-optimal region, and respond appropriately, for example by relaxing acceptance conditions or terminating exploration more effectively. 
In addition, we may consider reward scheduling mechanisms that explicitly guide the expected reward progression over time. These could enforce soft constraints, such as targeting specific reward levels at certain diffusion steps.
Ideally, these mechanisms would remain general enough to apply across different reward models and diffusion pipelines.

However, in this work, we focus on effective exploration strategies and intentionally adopt a simple scheduling formulation to demonstrate that even basic adaptive criteria already yield substantial benefits. This provides a solid foundation for future extensions incorporating more sophisticated scheduling for diffusion sampling with controlled explorations.

\subsection{Experimental Validations on Inference Scaling}
While we demonstrate the effectiveness of adaptive exploration under moderate computational budgets, current evaluation does not fully capture performance at larger scales due to resource constraints. Future work could investigate broader configurations, including deeper exploration depths, increased candidate pool sizes, and stricter acceptance thresholds, to better understand how the proposed method scales with inference cost. A more comprehensive evaluation may reveal whether the quality-efficiency trade-offs follow predictable trends, as suggested by recent work on test-time scaling in diffusion models. These insights would guide the design of inference-time strategies adaptable to different budget constraints.

\section{LLM Usage Statement}
We used the GPT series exclusively during the writing phase of this paper, for the purpose of polishing the manuscript. Specifically, we provided prompts such as: ``Revise the grammar and academic wording of this paragraph, and list aspects to be improved, including suggestions on how to improve them.'' All outputs were subsequently reviewed and manually revised to ensure that technical accuracy was preserved. No part of the research ideation or experimental analysis relied on LLM assistance.

\section{Qualitative Analysis}\label{sec:sup_qualitative}

\subsection{More Qualitative Samples}
\begin{figure*}[th!]
  \centering
  \scriptsize

  \begin{minipage}[t]{0.07\textwidth}
      \rotatebox{90}{\parbox{2.2cm}{\centering  A photo of a tree with
\textbf{eggs growing} on it}}
    \end{minipage}%
\begin{subfigure}[t]{0.183\textwidth}
  \centering
  \includegraphics[width=\linewidth, trim=0pt 0pt 0pt 0pt, clip]{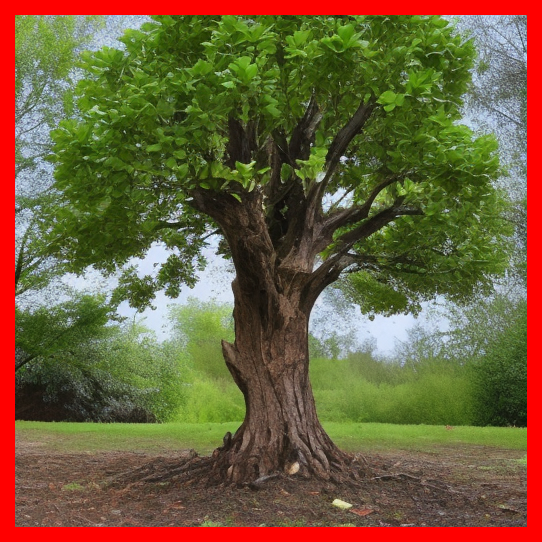}
\end{subfigure}\hspace{1pt}%
\begin{subfigure}[t]{0.183\textwidth}
  \centering
  \includegraphics[width=\linewidth, trim=0pt 0pt 0pt 0pt, clip]{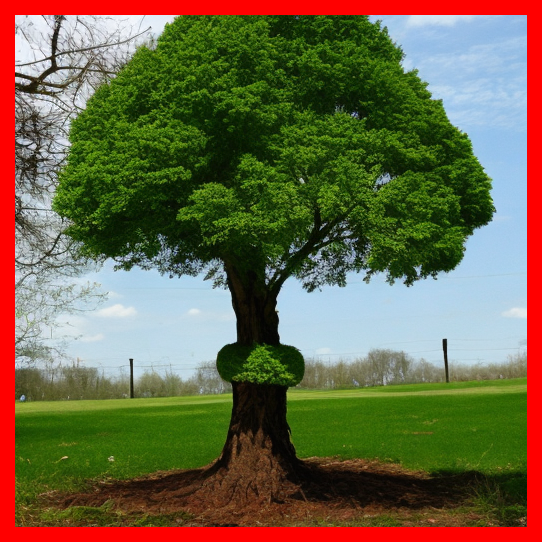}
\end{subfigure}\hspace{1pt}%
\begin{subfigure}[t]{0.183\textwidth}
  \centering
  \includegraphics[width=\linewidth, trim=0pt 0pt 0pt 0pt, clip]{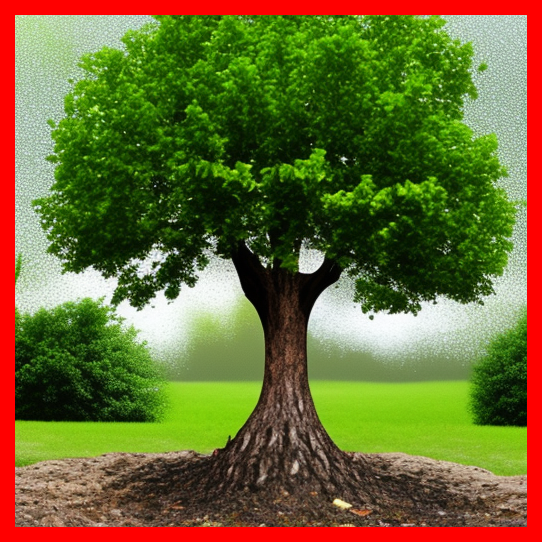}
\end{subfigure}\hspace{1pt}%
\begin{subfigure}[t]{0.183\textwidth}
  \centering
  \includegraphics[width=\linewidth, trim=0pt 0pt 0pt 0pt, clip]{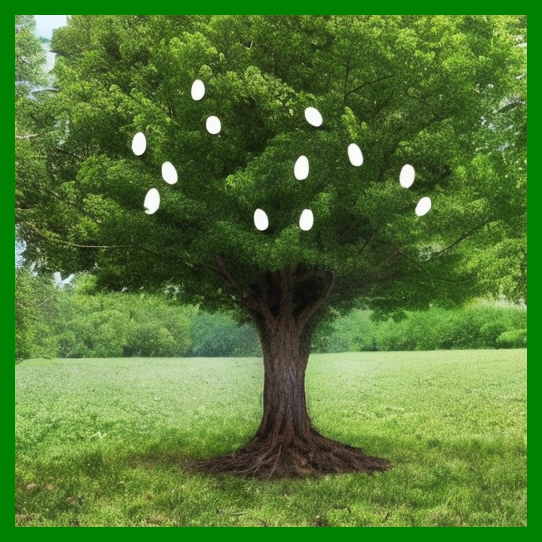}
\end{subfigure}\hspace{1pt}%
\begin{subfigure}[t]{0.183\textwidth}
  \centering
  \includegraphics[width=\linewidth, trim=0pt 0pt 0pt 0pt, clip]{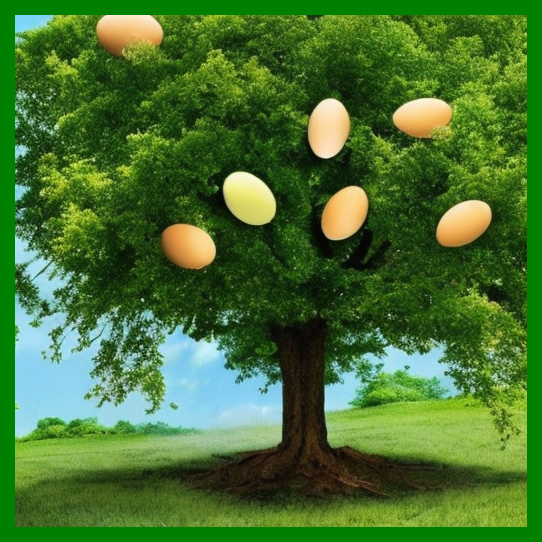}
\end{subfigure}%

  \begin{minipage}[t]{0.07\textwidth}
      \rotatebox{90}{\parbox{2.1cm}{\centering 3D \textbf{Pac Man} \\ in real life}}
    \end{minipage}%
\begin{subfigure}[t]{0.183\textwidth}
    \centering
    \includegraphics[width=\linewidth]{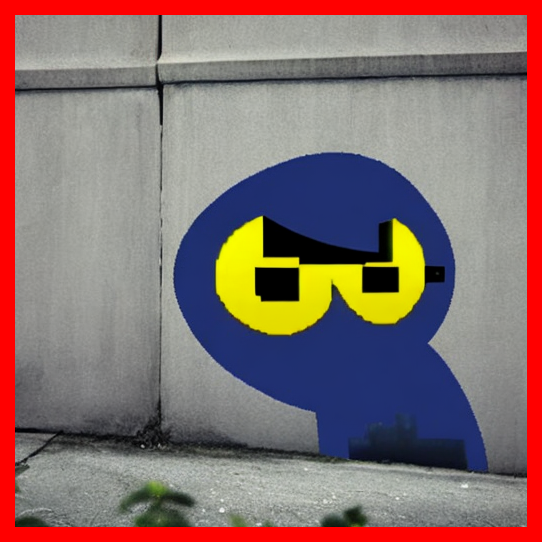}
  \end{subfigure}\hspace{1pt}%
  \begin{subfigure}[t]{0.183\textwidth}
    \centering
    \includegraphics[width=\linewidth]{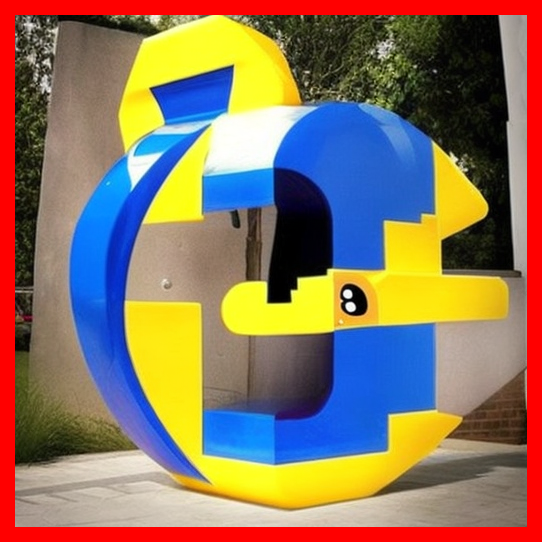}
  \end{subfigure}\hspace{1pt}%
  \begin{subfigure}[t]{0.183\textwidth}
    \centering
    \includegraphics[width=\linewidth]{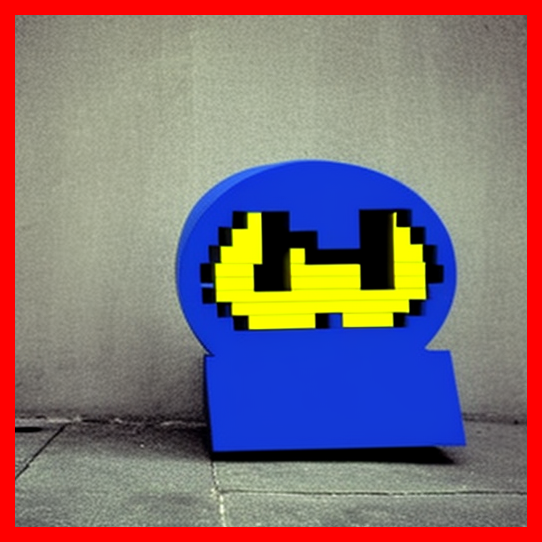}
  \end{subfigure}\hspace{1pt}%
  \begin{subfigure}[t]{0.183\textwidth}
    \centering
    \includegraphics[width=\linewidth]{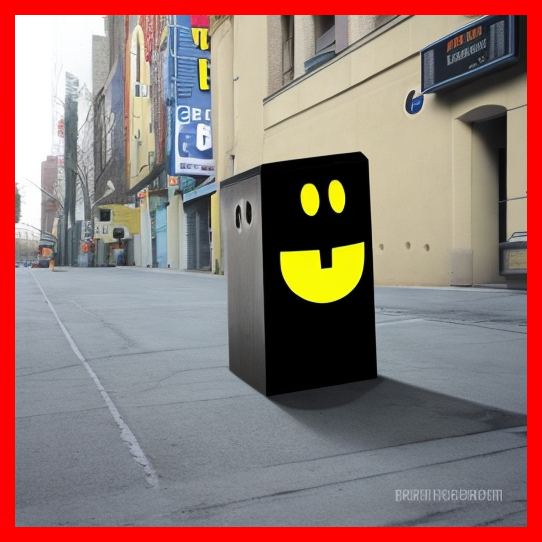}
  \end{subfigure}\hspace{1pt}%
  \begin{subfigure}[t]{0.183\textwidth}
    \centering
    \includegraphics[width=\linewidth]{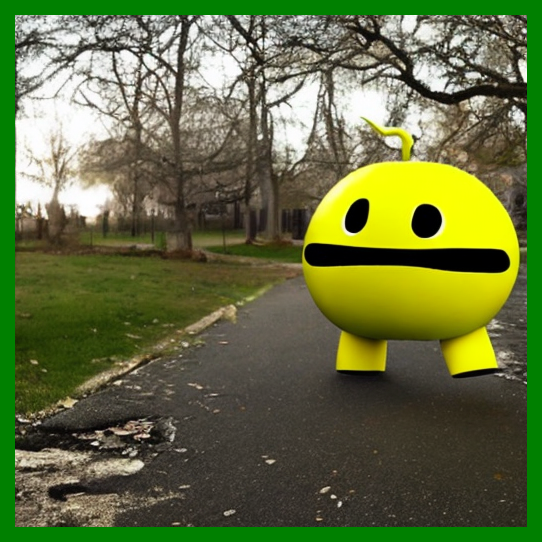}
  \end{subfigure}

  \begin{minipage}[t]{0.07\textwidth}
      \rotatebox{90}{\parbox{2.2cm}{\centering \textbf{Two cats watering} roses in a greenhouse}}
    \end{minipage}%
\begin{subfigure}[t]{0.183\textwidth}
    \centering
    \includegraphics[width=\linewidth]{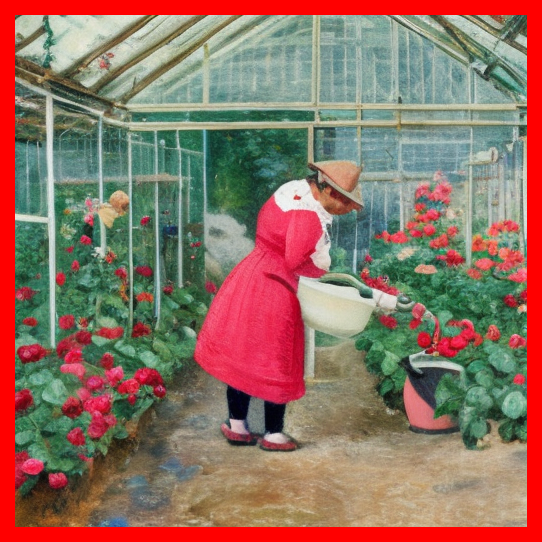}
  \end{subfigure}\hspace{1pt}%
  \begin{subfigure}[t]{0.183\textwidth}
    \centering
    \includegraphics[width=\linewidth]{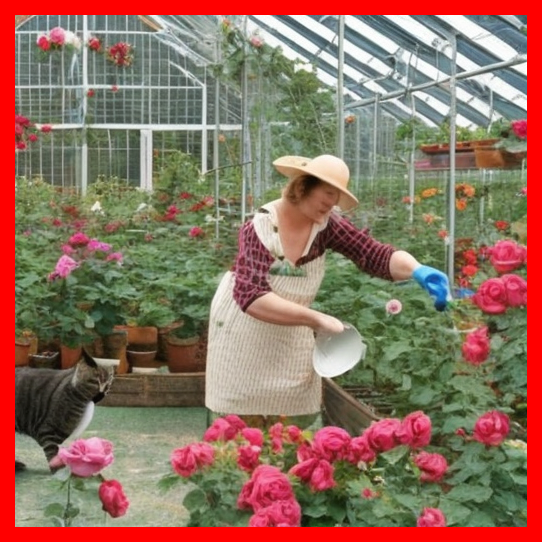}
  \end{subfigure}\hspace{1pt}%
  \begin{subfigure}[t]{0.183\textwidth}
    \centering
    \includegraphics[width=\linewidth]{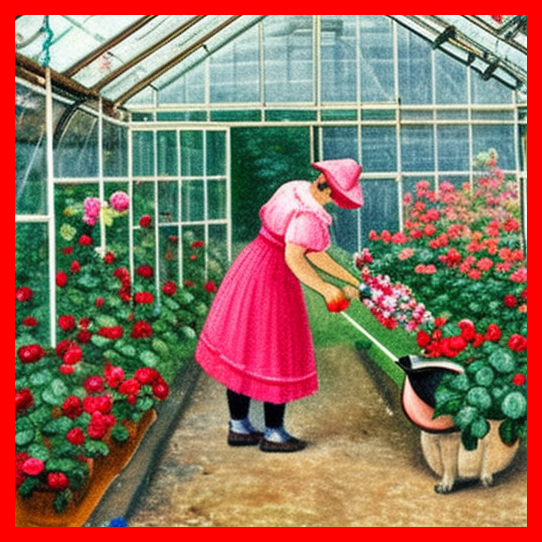}
  \end{subfigure}\hspace{1pt}%
  \begin{subfigure}[t]{0.183\textwidth}
    \centering
    \includegraphics[width=\linewidth]{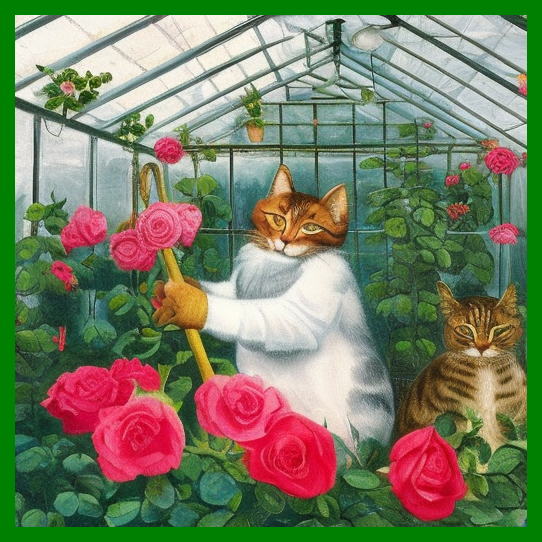}
  \end{subfigure}\hspace{1pt}%
  \begin{subfigure}[t]{0.183\textwidth}
    \centering
    \includegraphics[width=\linewidth]{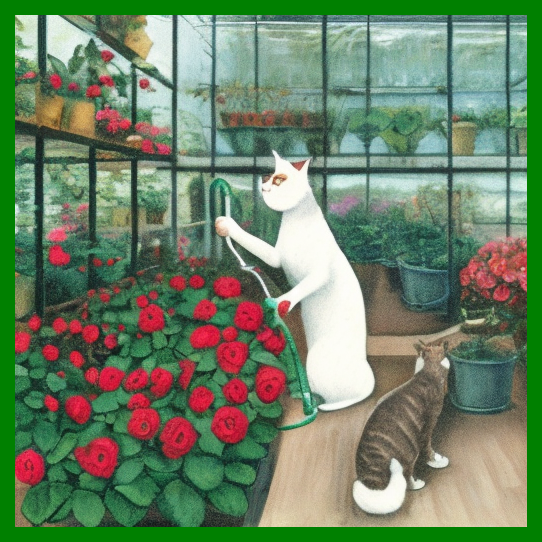}
  \end{subfigure}

  \begin{minipage}[t]{0.07\textwidth}
      \rotatebox{90}{\parbox{2.1cm}{\centering ~a \textbf{blue cake} and   a red suitcase}}
    \end{minipage}%
\begin{subfigure}[t]{0.183\textwidth}
    \centering
    \includegraphics[width=\linewidth]{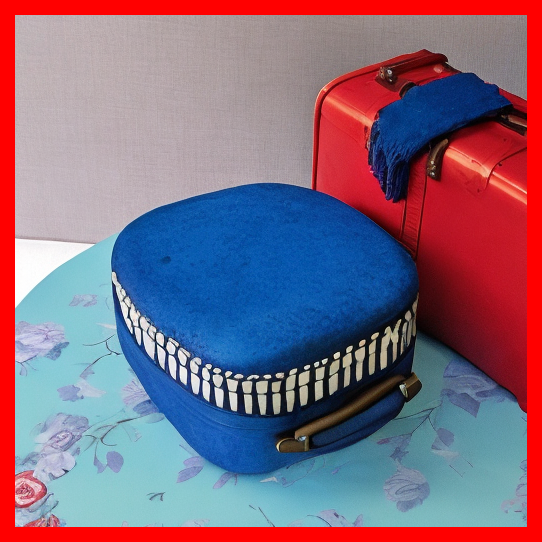}
  \end{subfigure}\hspace{1pt}%
  \begin{subfigure}[t]{0.183\textwidth}
    \centering
    \includegraphics[width=\linewidth]{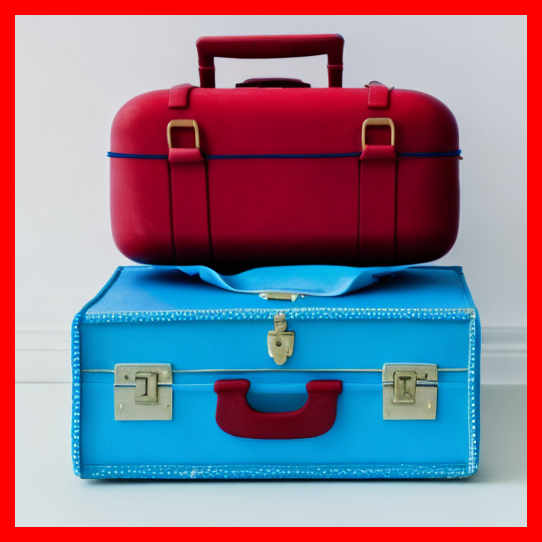}
  \end{subfigure}\hspace{1pt}%
  \begin{subfigure}[t]{0.183\textwidth}
    \centering
    \includegraphics[width=\linewidth]{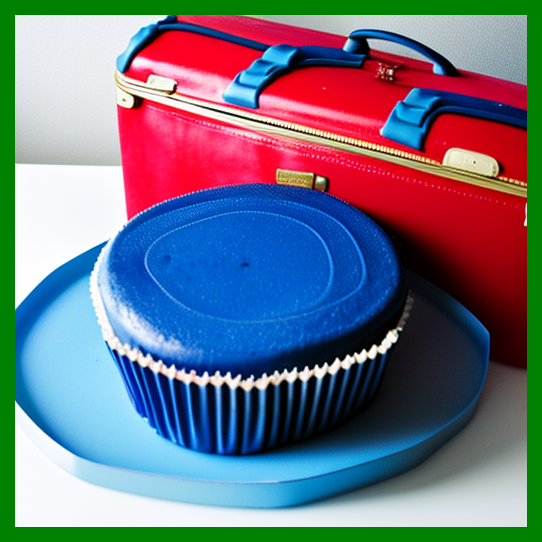}
  \end{subfigure}\hspace{1pt}%
  \begin{subfigure}[t]{0.183\textwidth}
    \centering
    \includegraphics[width=\linewidth]{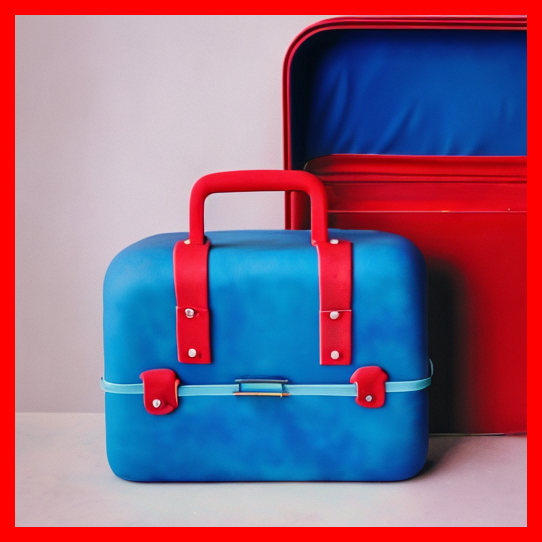}
  \end{subfigure}\hspace{1pt}%
  \begin{subfigure}[t]{0.183\textwidth}
    \centering
    \includegraphics[width=\linewidth]{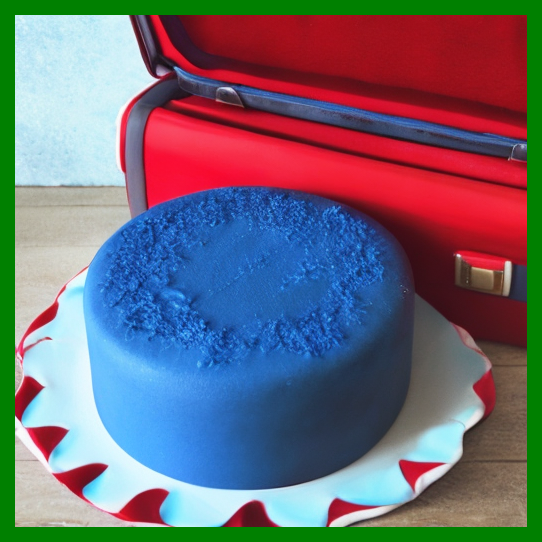}
  \end{subfigure}

\begin{minipage}[t]{0.07\textwidth}
      \rotatebox{90}{\parbox{2.2cm}{\centering a circular clock \& a \textbf{triangular shelf}}}
    \end{minipage}%
\begin{subfigure}[t]{0.183\textwidth}
    \centering
    \includegraphics[width=\linewidth]{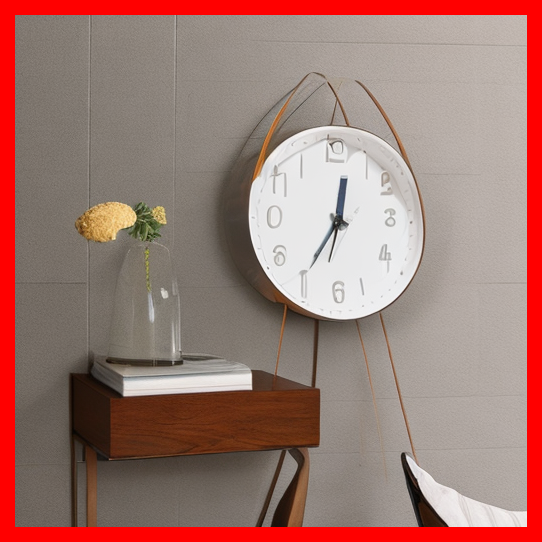}
  \end{subfigure}\hspace{1pt}%
  \begin{subfigure}[t]{0.183\textwidth}
    \centering
    \includegraphics[width=\linewidth]{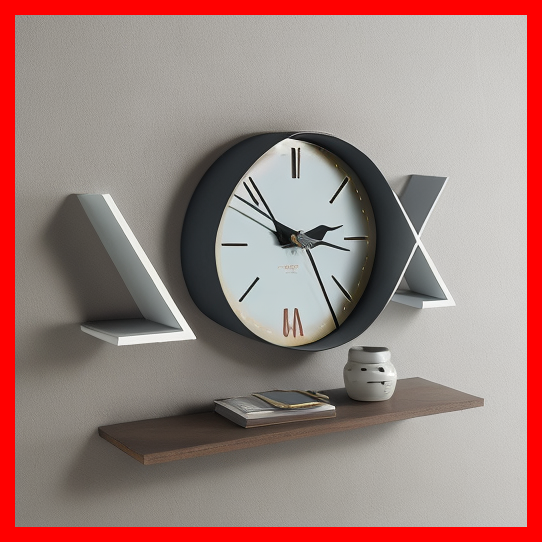}
  \end{subfigure}\hspace{1pt}%
  \begin{subfigure}[t]{0.183\textwidth}
    \centering
    \includegraphics[width=\linewidth]{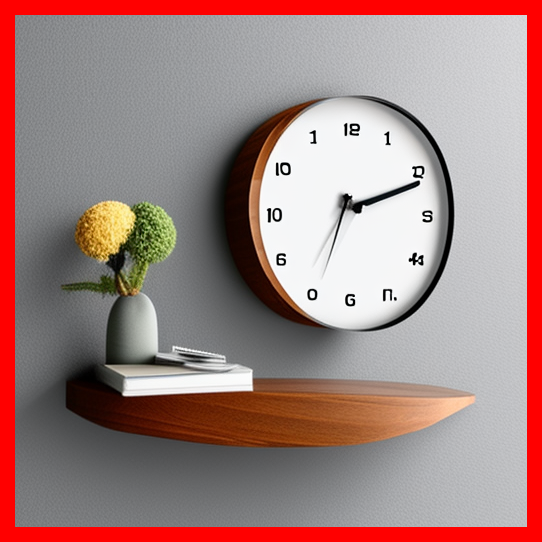 }
  \end{subfigure}\hspace{1pt}%
  \begin{subfigure}[t]{0.183\textwidth}
    \centering
    \includegraphics[width=\linewidth]{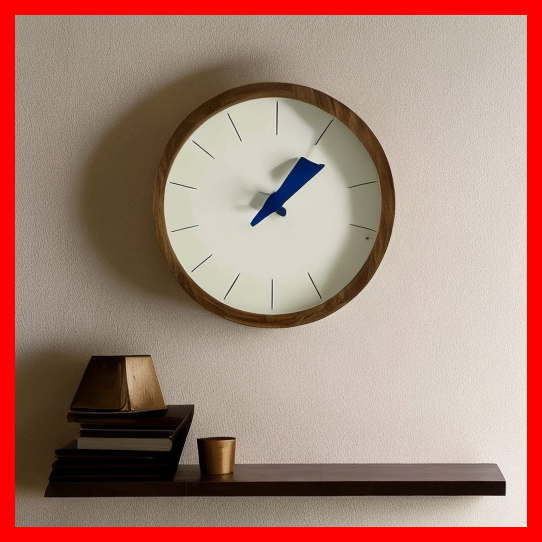}
  \end{subfigure}\hspace{1pt}%
  \begin{subfigure}[t]{0.183\textwidth}
    \centering
    \includegraphics[width=\linewidth]{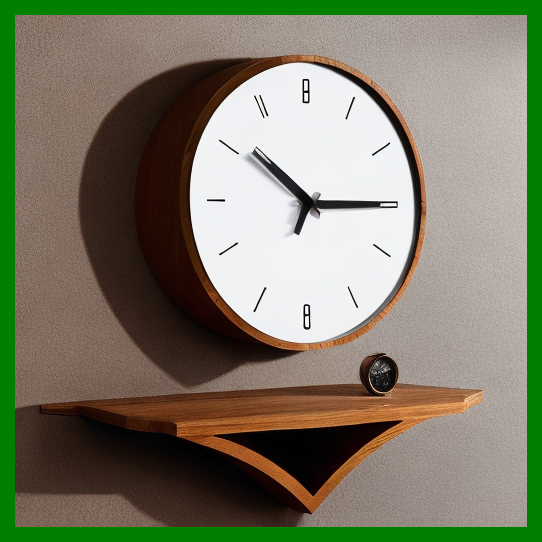}
  \end{subfigure}

\begin{minipage}[t]{0.07\textwidth}
      \rotatebox{90}{\scriptsize\parbox{2.1cm}{\centering a \textbf{plastic container} and \\  a leather chair}}
    \end{minipage}%
\begin{subfigure}[t]{0.183\textwidth}
    \centering
    \includegraphics[width=\linewidth]{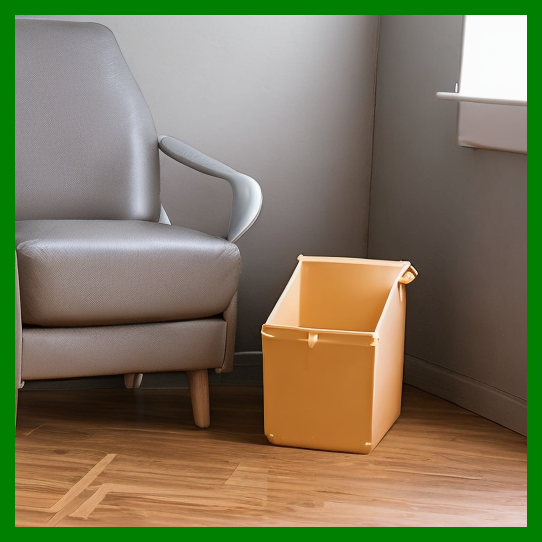}
    \textbf{\normalsize DDIM}
  \end{subfigure}\hspace{1pt}%
  \begin{subfigure}[t]{0.183\textwidth}
    \centering
    \includegraphics[width=\linewidth]{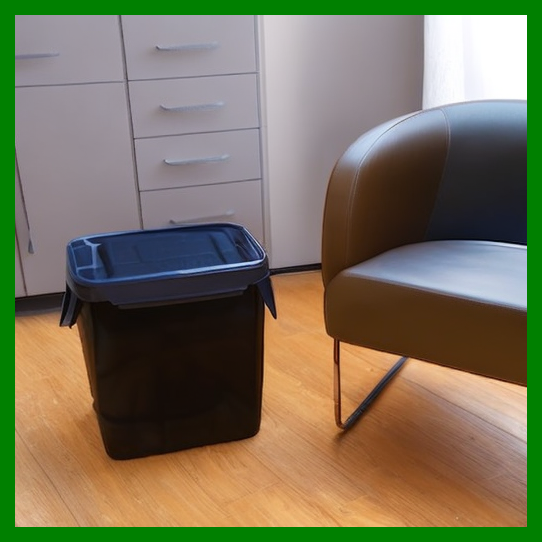}
    \textbf{\normalsize Resampling}
  \end{subfigure}\hspace{1pt}%
  \begin{subfigure}[t]{0.183\textwidth}
    \centering
    \includegraphics[width=\linewidth]{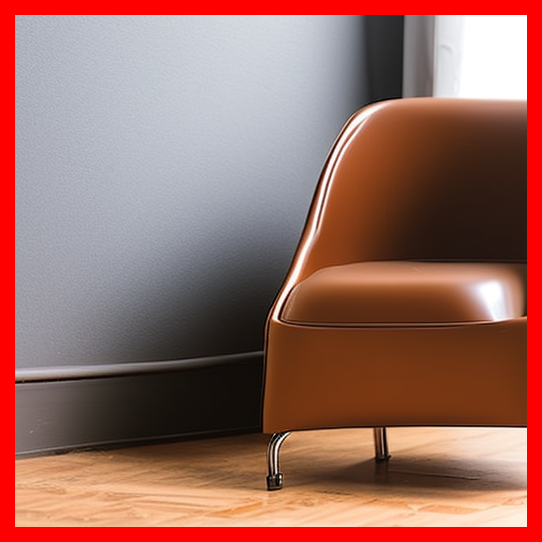}
    \textbf{\normalsize Z-Sampling}
  \end{subfigure}\hspace{1pt}%
  \begin{subfigure}[t]{0.183\textwidth}
    \centering
    \includegraphics[width=\linewidth]{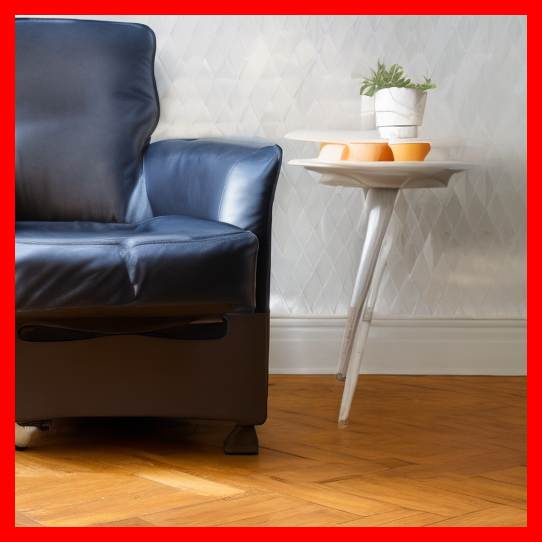}
    \textbf{\normalsize SOP}
  \end{subfigure}\hspace{1pt}%
  \begin{subfigure}[t]{0.183\textwidth}
    \centering
    \includegraphics[width=\linewidth]{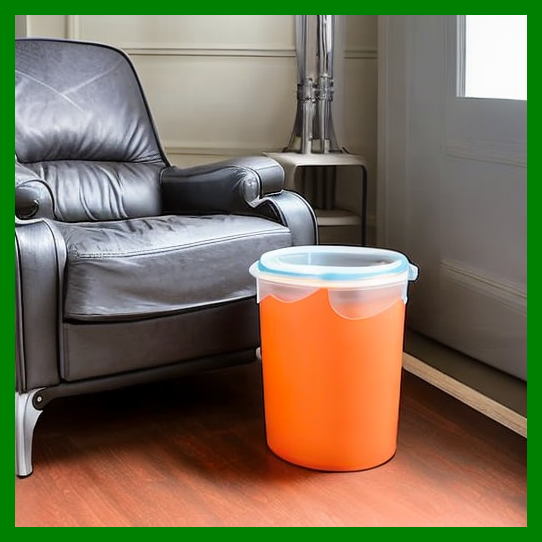}
    \textbf{\normalsize $\mathtt{Ctrl}$-$\mathtt{Z}$ (Ours)}
  \end{subfigure}

  \caption{Qualitative comparison of different sampling methods. Generated images that align with the input condition (shown on the left) are highlighted with \textit{green bounding boxes}, while erroneous or suboptimal images are marked in \textit{red}. 
  Inconsistently generated concept are also \textbf{bolded} in the prompt. `\&' stands for `and'. Our proposed \abb{} generates more coherent and condition-aligned outputs.} 
  \label{fig:more_qualitative}
\end{figure*}

Figure~\ref{fig:more_qualitative} presents a set of additional qualitative results generated using different sampling methods. As shown, the proposed \abb{} consistently produces the most visually coherent and prompt-aligned outputs across a variety of test cases. In contrast, the baseline methods often fail to maintain global consistency, producing outputs that may appear locally plausible but deviate from the intended semantics of the prompt. These results highlight the effectiveness of \abb{} in generating globally aligned images, which we attribute to its adaptive mechanism for escaping local optima. This mechanism enables the method to avoid premature convergence to suboptimal visual representations, thereby mitigating the common failure mode of generating outputs that are visually appealing but misaligned with the conditional input.

\subsection{Samples on Intermediate Exploration Steps}\label{sec::intermediate}

\begin{figure}
    \centering
    \hspace*{10mm}
    \includegraphics[width=0.8\linewidth]{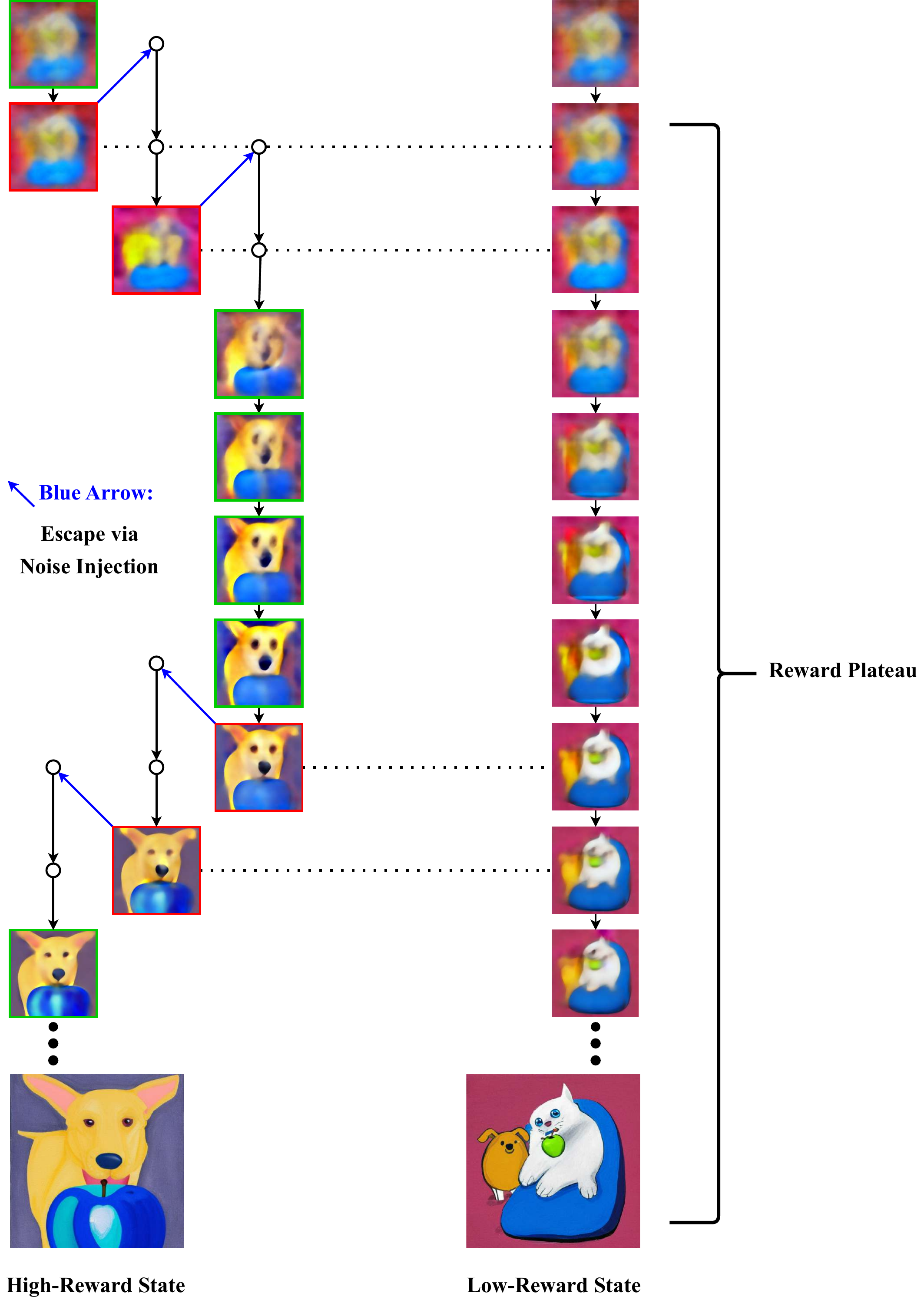}
    \caption{\textit{Sampling trajectories} over the first 10 steps, comparing DDIM (\textit{right}) and \abb{} (\textit{left}). Prompt: ``a yellow dog and a blue apple''. \textit{Red boxes} mark steps where exploration is triggered, and \textit{green boxes} mark regular steps. \rev{DDIM greedily steps towards higher probability regions and becomes trapped in a local optima state (\textit{white dog}). In contrast, $\mathtt{Ctrl}$-$\mathtt{Z}$ Sampling explores via inversion to escape this local optima, converging to a higher-reward state (\textit{yellow dog}).}
    }
    \label{fig:treepath}
    \vspace{-5mm}
\end{figure}

We first present in \Cref{fig:treepath}, a tree-structured visualization of the sampling path during the first 10 steps, comparing our \abb{} with standard DDIM. While DDIM follows a single \rev{greedy optimization path determined by the diffusion prior}, \abb{} performs controlled zigzag explorations that branch out when the reward fails to improve. \rev{ This enables the model to escape local optima states and converge to a high-reward state that satisfies the conditional alignment.} In all visualizations, steps where the stepwise reward shows insufficient improvement compared to the previous step and exploration is initialized (\ie, $R([c], \hat{x}_0^{\,t-1}) < r_{\text{prev}} + \delta$ with $\delta = 0$) are highlighted in red, while steps without explorations are shown in green.

\begin{figure}[t]
    \centering
    \begin{subfigure}{0.75\linewidth}
        \centering
        \includegraphics[width=\linewidth]{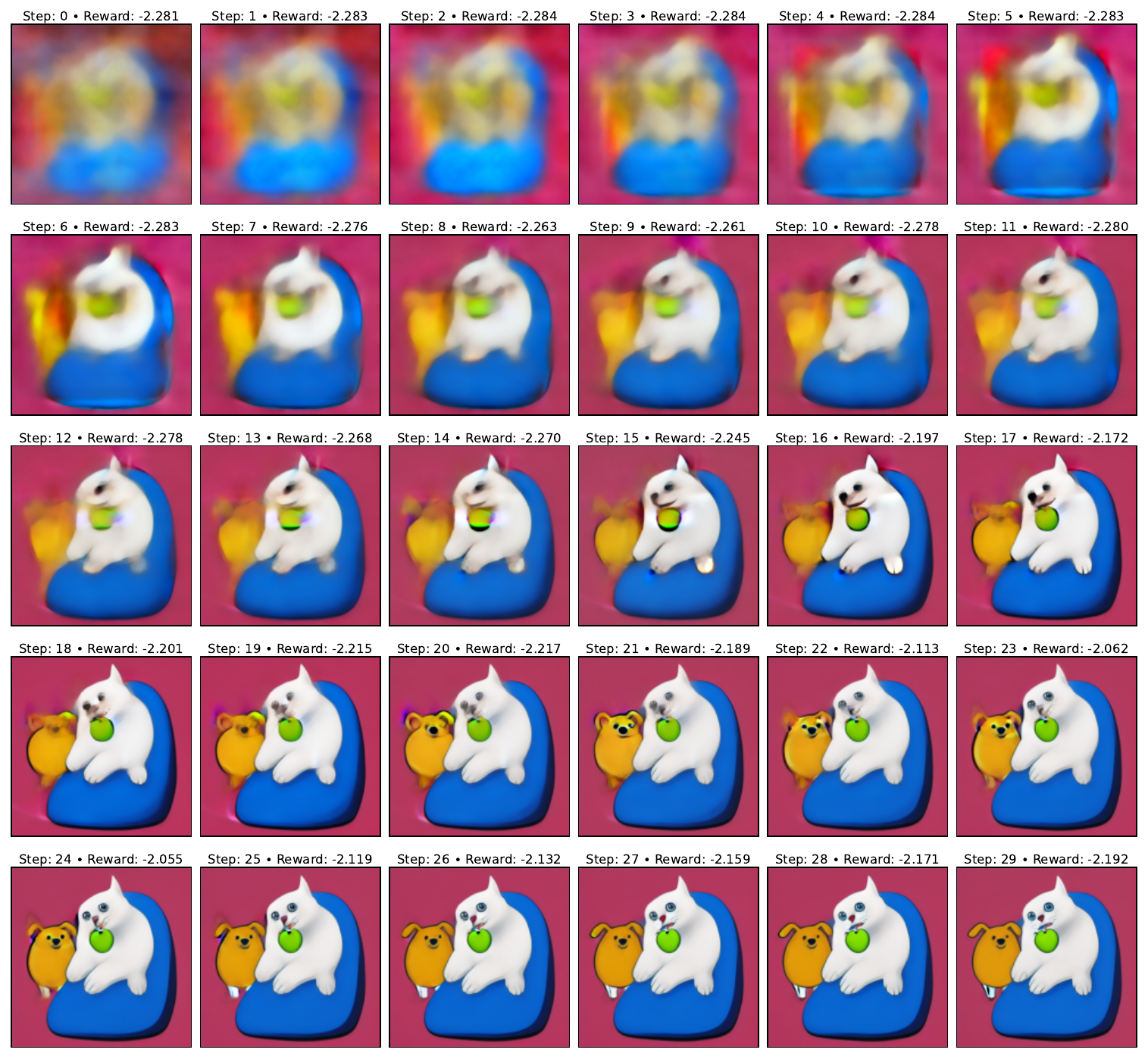}
        \vspace{-5mm}
        \caption{DDIM. Prompt: ``a yellow dog and a blue apple''}
    \end{subfigure}
    
    
    
    
    \begin{subfigure}{0.75\linewidth}
        \centering
        \includegraphics[width=\linewidth]{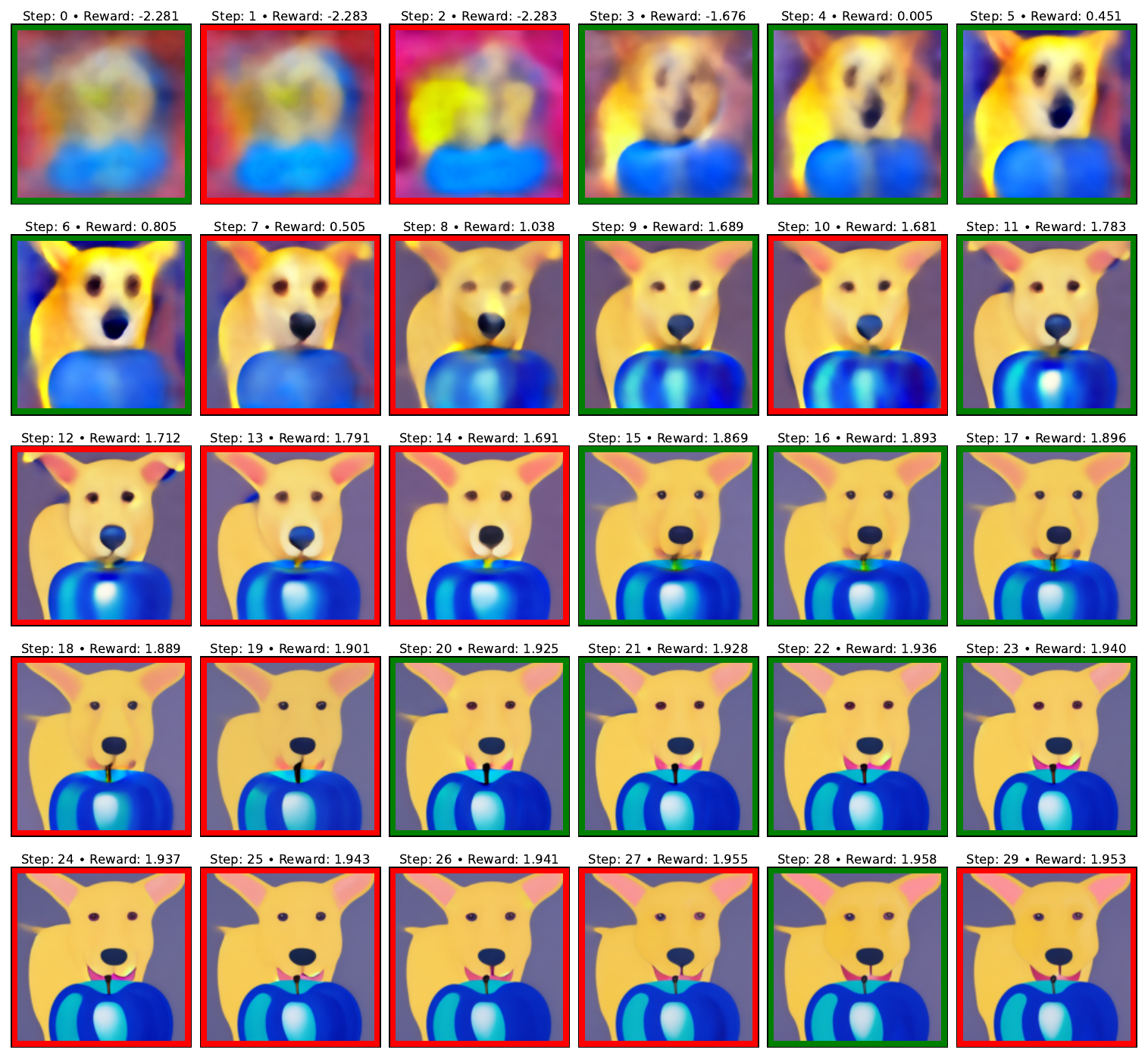}
        \vspace{-5mm}
        
        \caption{\abb{}. Prompt: ``a yellow dog and a blue apple''}
    \end{subfigure}
    
    \vspace{-1mm}
    \caption{Qualitative comparison of the decoded $X_0^{t-1}$ across the first 30 over 50 generation steps.
    Steps with exploration initiated are highlighted in \textit{red}, while others are shown in \textit{green}.}
    \label{fig:step1}
\end{figure}

\begin{figure}[t]
    \centering
    \begin{subfigure}{0.75\linewidth}
        \centering
        \includegraphics[width=\linewidth]{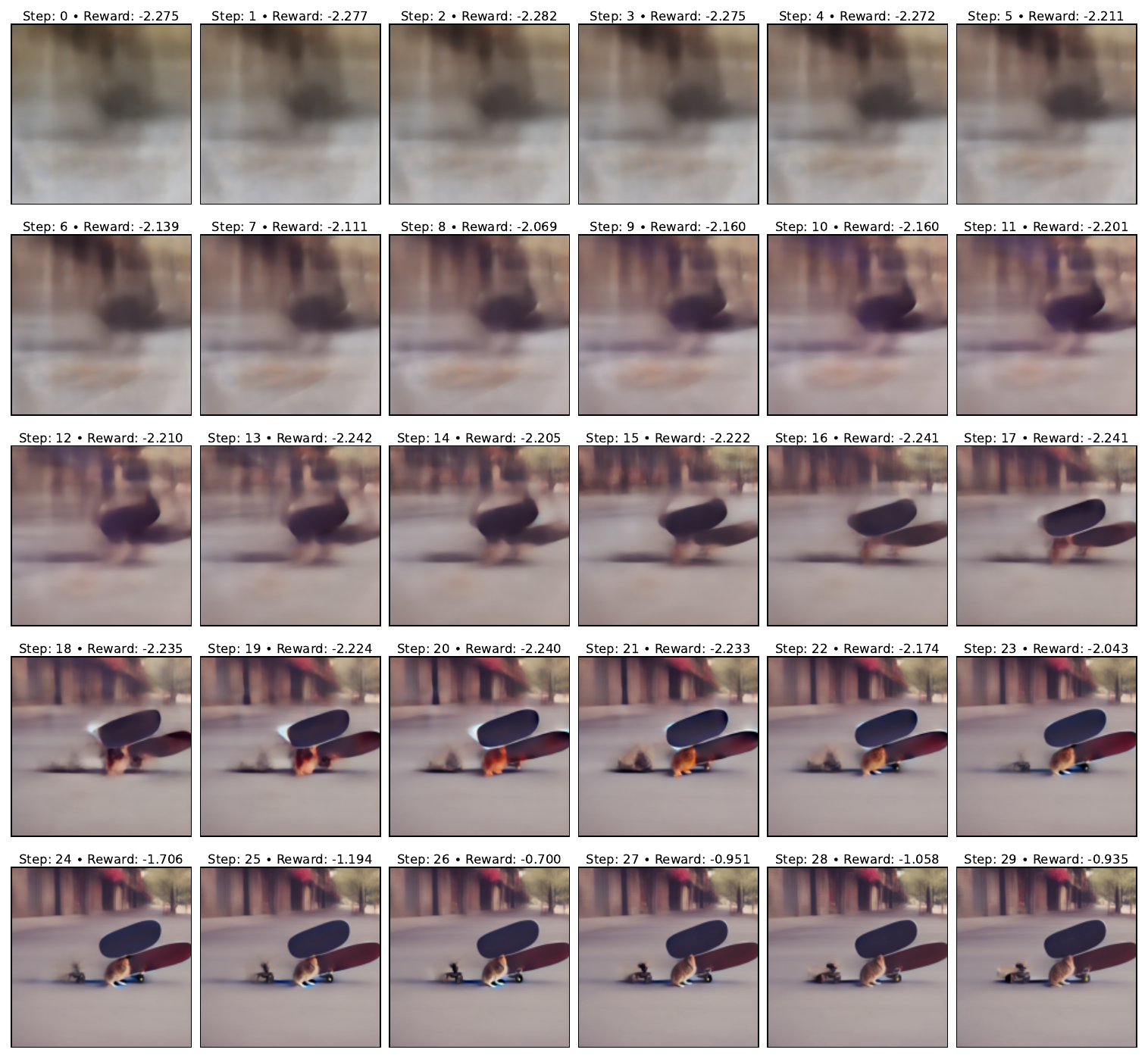}
        \vspace{-5mm}
        \caption{DDIM. Prompt: ``A cat sitting on a skateboard in the middle of the street.''}
    \end{subfigure}
    
    
    
    
    \begin{subfigure}{0.75\linewidth}
        \centering
        \includegraphics[width=\linewidth]{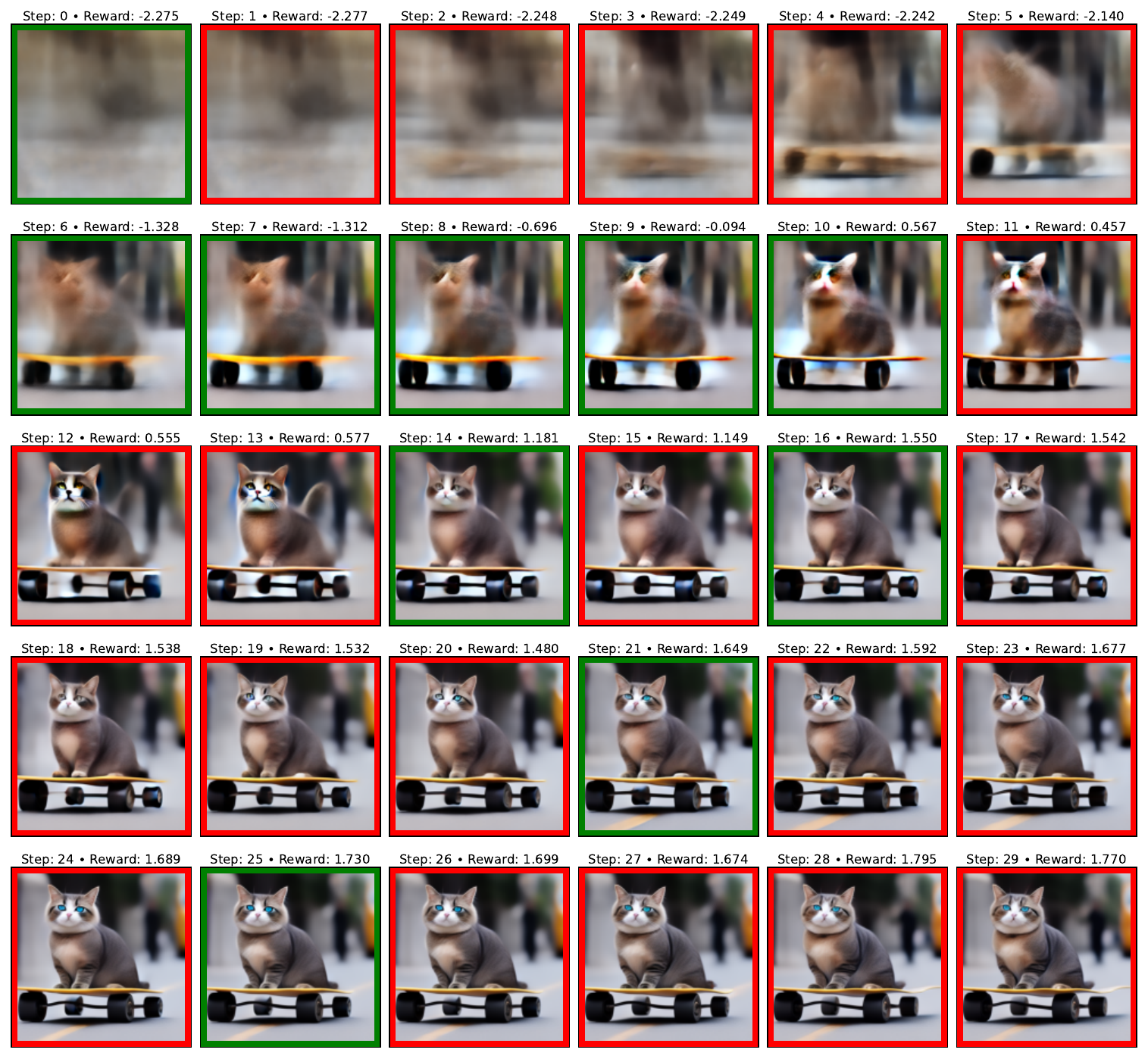}
        \vspace{-5mm}
        
        \caption{\abb{}. Prompt: ``A cat sitting on a skateboard in the middle of the street.''}
    \end{subfigure}
    
    \vspace{-1mm}
    \caption{\scriptsize Qualitative comparison of the decoded $X_0^{t-1}$ across the first 30 over 50 generation steps.
    Steps with exploration initiated are highlighted in \textit{red}, while others are shown in \textit{green}.}
    \label{fig:step2}
\end{figure}

\begin{figure}[t]
    \centering
    \begin{subfigure}{0.75\linewidth}
        \centering
        \includegraphics[width=\linewidth]{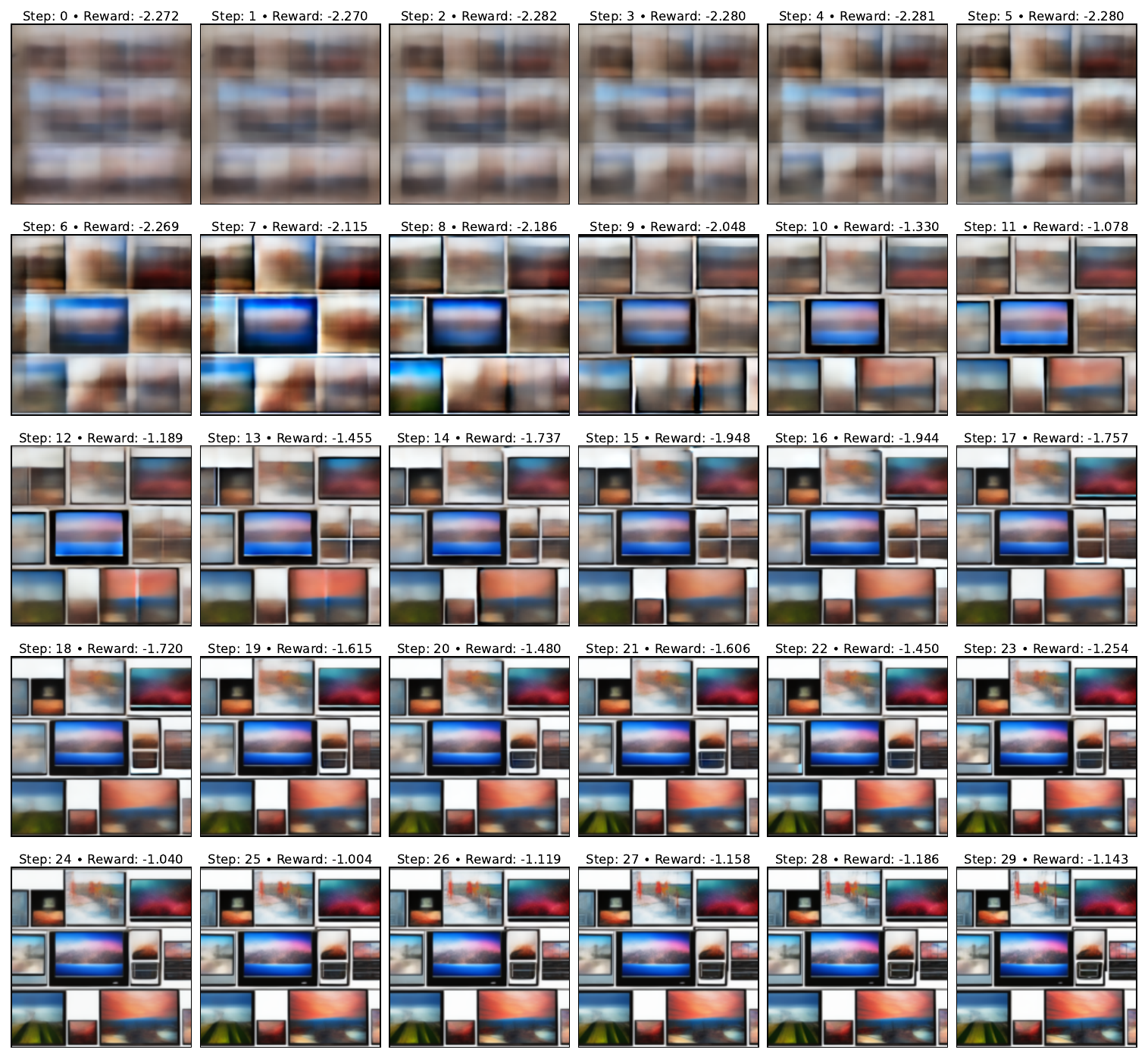}
        \vspace{-5mm}
        \caption{DDIM. Prompt: ``six televisions''}
    \end{subfigure}
    
    \begin{subfigure}{0.75\linewidth}
        \centering
        \includegraphics[width=\linewidth]{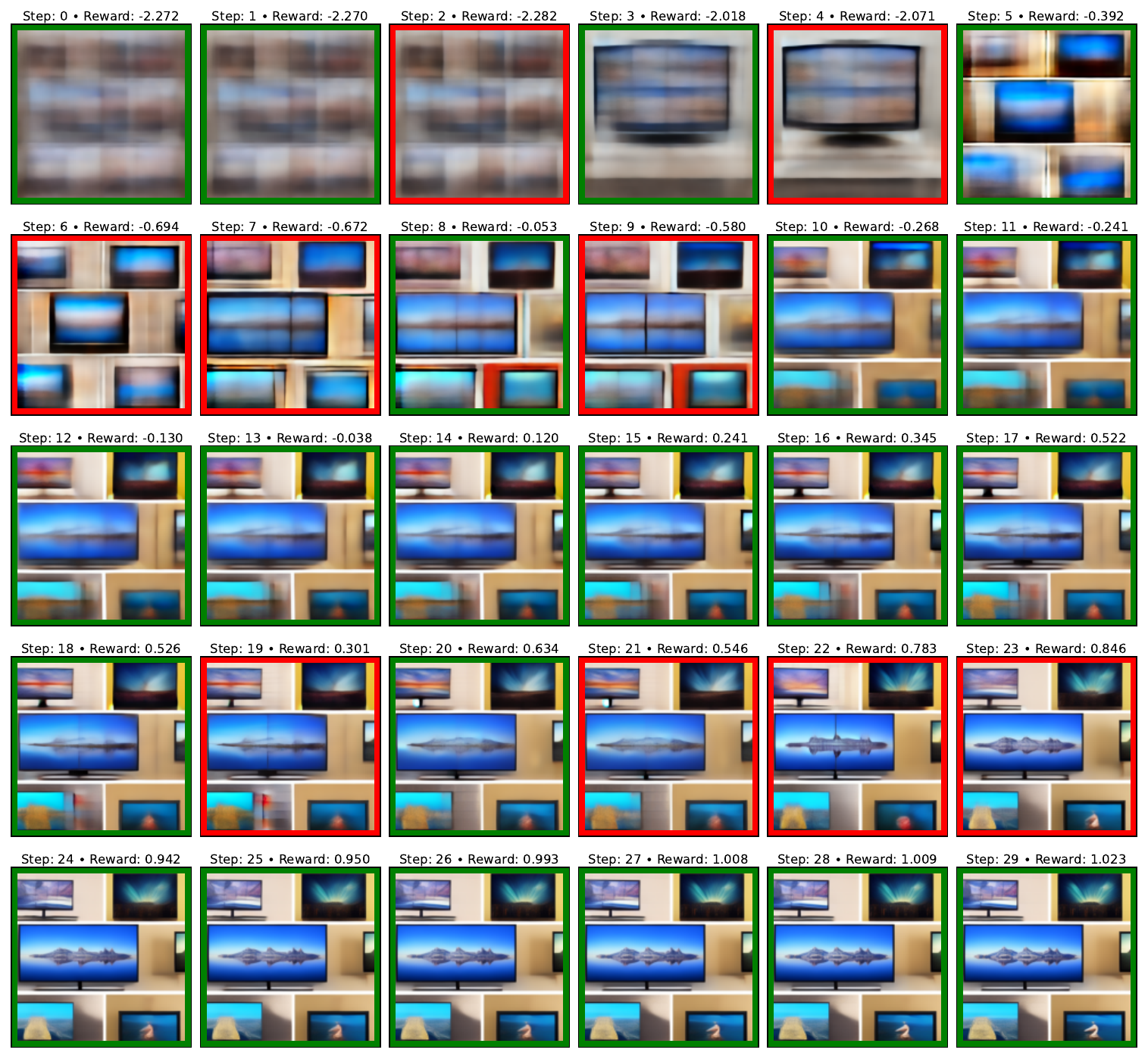}
        \vspace{-5mm}
        
        \caption{\abb{}. Prompt: ``six televisions''}
    \end{subfigure}
    
    \vspace{-1mm}
    \caption{Qualitative comparison of the decoded $X_0^{t-1}$ across the first 30 over 50 generation steps.
    Steps with exploration initiated are highlighted in \textit{red}, while others are shown in \textit{green}.}
    \label{fig:step3}
\end{figure}

\begin{figure}[t]
    \centering
    \scriptsize
    \begin{subfigure}{0.75\linewidth}
        \centering
        \includegraphics[width=\linewidth]{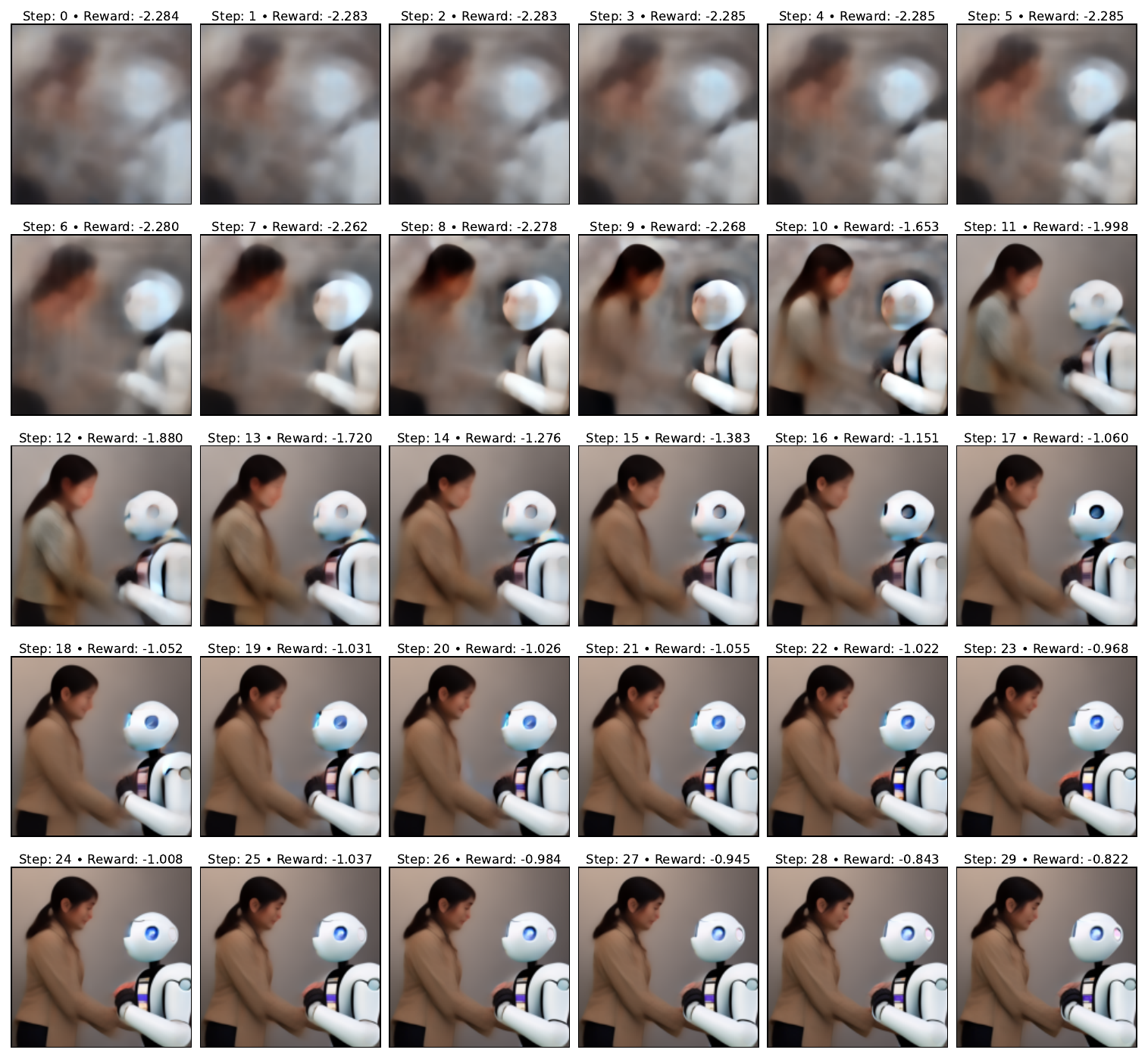}
        \vspace{-5mm}
        \caption{DDIM. Prompt: ``A robot comforting a researcher who is sad about being replaced.''}
    \end{subfigure}
    
    \begin{subfigure}{0.75\linewidth}
        \centering
        \includegraphics[width=\linewidth]{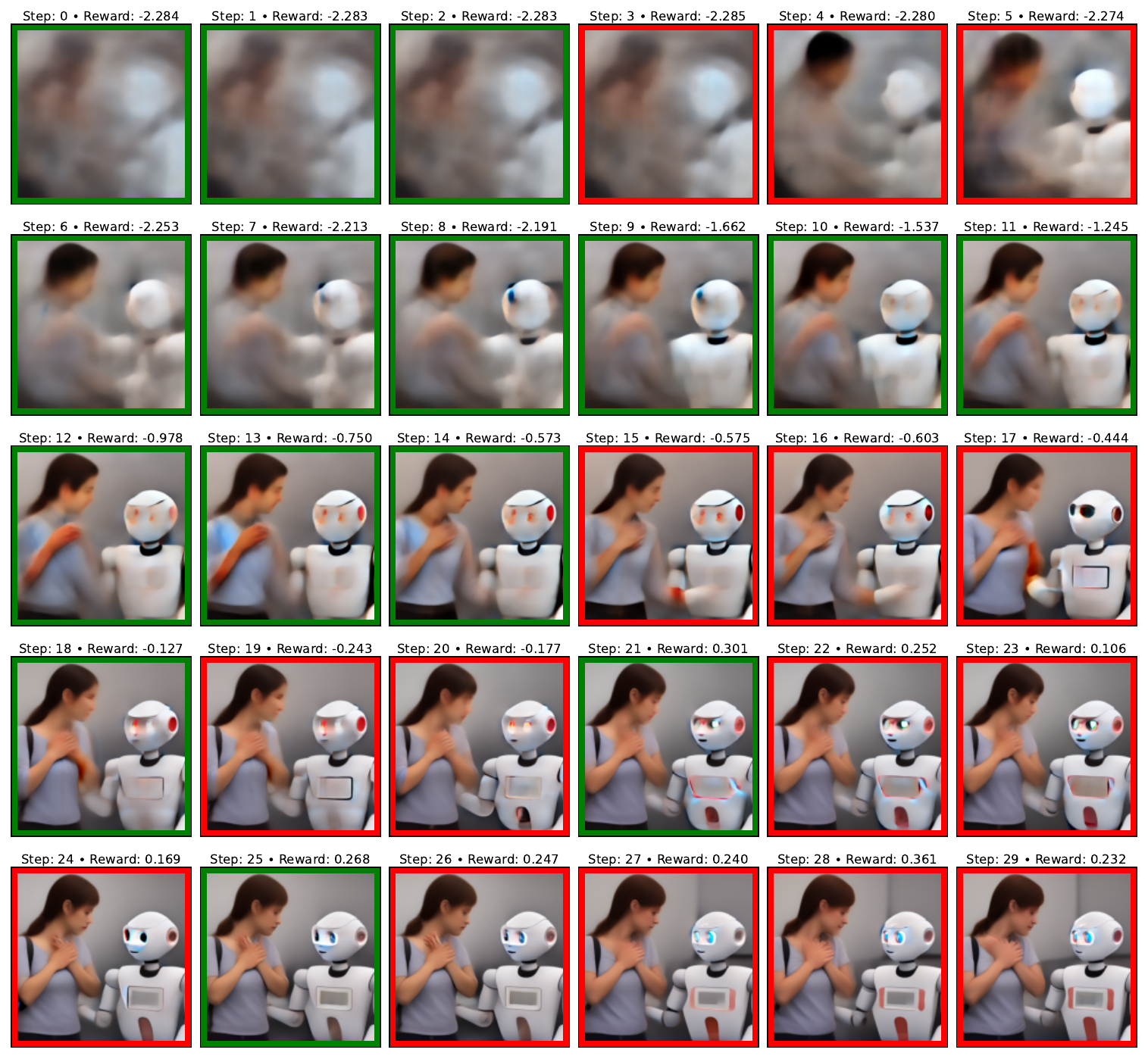}
        \vspace{-5mm}
        
        \caption{\abb{}. Prompt:  ``A robot comforting a researcher who is sad about being replaced.''}
    \end{subfigure}
    
    \vspace{-1mm}
    \caption{\scriptsize Qualitative comparison of the decoded $X_0^{t-1}$ across the first 30 over 50 generation steps.
    Steps with exploration initiated are highlighted in \textit{red}, while others are shown in \textit{green}.}
    \label{fig:step4}
\end{figure}

We then provide more detailed visualizations of the predicted $\hat{x}_0^{\,t-1}$ and their associated rewards in \Cref{fig:step1,fig:step2,fig:step3,fig:step4}. For each step, we show the decoded intermediate prediction together with its step index and reward. Only the first 30 out of 50 total sampling steps are displayed, as the diffusion model already converges to reasonable generations in the middle steps, whereas later steps show only minor improvements that are barely perceivable. For red-highlighted steps, the updated $X_0^{t-1}$ after exploration is not visualized, and the subsequent $X_0^{t-2}$ shown in the image may still fail the reward check relative to the omitted updated state, thereby also being highlighted in red (these $X_0^{t-1}$ states are omitted in all figures for clarity).

\abb{} improves sampling in two main ways. (1) Early-stage exploration. By injecting noise at the initial steps, \abb{} perturbs low-frequency patterns and alters the denoising trajectory, effectively steering the model away from locally optimal states with limited future reward. For instance, in \Cref{fig:step1}, explorations at steps 1–2 reduce the excessive blueness in the top-right corner of the dog and enhance the yellow tones in the center, yielding a more aligned trajectory. (2) Mid-stage refinement. Once low-frequency structures are largely fixed, \abb{} continues to refine local details. As shown in \Cref{fig:step4}, from step 15 onward it repeatedly improves alignment around the robot’s eyes and hand pose, producing structurally consistent yet more plausible results. 

It can also be observed that explorations near the 30th step often yield imperceptible changes, highlighting the limited effectiveness of late-stage explorations and motivating the use of an exploration window that restricts them to the first $\lambda$ steps. Nonetheless, some inefficiency remains. The reward-based adaptive initiation strategy is intended to suppress unnecessary explorations, yet when the image already reaches a satisfactory state early in the diffusion process, rewards tend to saturate, making further improvements difficult. This often triggers redundant explorations with trivial gains, as seen in \Cref{fig:step2}, where many explorations were initiated despite negligible improvements in visual quality or alignment after the first 15 steps. These findings suggest that more sophisticated scheduling of exploration initiation could be a promising direction for future work.

